\documentclass{article}

\PassOptionsToPackage{numbers, compress}{natbib}

\usepackage[main, final]{neurips_2026}

\usepackage[utf8]{inputenc} % allow utf-8 input
\usepackage[T1]{fontenc}    % use 8-bit T1 fonts
\usepackage{hyperref}       % hyperlinks
\usepackage[font=scriptsize,labelfont=bf]{caption}
\usepackage{url}            % simple URL typesetting
\usepackage{booktabs}       % professional-quality tables
\usepackage{amsfonts}       % blackboard math symbols
\usepackage{nicefrac}       % compact symbols for 1/2, etc.
\usepackage{microtype}      % microtypography
\usepackage{xcolor}         % colors

\usepackage{algorithm}
\usepackage[noend]{algpseudocode}

\usepackage{enumitem} 

\usepackage{pgfplots}
\usepackage{pgfplotstable}
\pgfplotsset{compat=1.18}

\usepackage{microtype}
\usepackage{graphicx}
\usepackage{subfigure}
\usepackage{booktabs} 
\usepackage{hyperref}

\usepackage{amsmath}
\usepackage{amssymb}
\usepackage{mathtools}
\usepackage{amsthm}

\usepackage{cleveref}

\usepackage{url}            % simple URL typesetting
\usepackage{booktabs}       % professional-quality tables
\usepackage{amsfonts}       % blackboard math symbols
\usepackage{nicefrac}       % compact symbols for 1/2, etc.
\usepackage{microtype}      % microtypography
\usepackage{xcolor}         % colors

\theoremstyle{plain}
\newtheorem{theorem}{Theorem}[section]

\newtheorem{lemma}[theorem]{Lemma}
\newtheorem{corollary}[theorem]{Corollary}
\theoremstyle{definition}
\newtheorem{definition}[theorem]{Definition}

\newtheorem{remark}[theorem]{Remark}

\usepackage[textsize=tiny]{todonotes}

\usepackage{dsfont}
\usepackage{braket}
\usepackage{pdflscape}
\usepackage{enumitem}
\usepackage{url}            % simple URL typesetting
\usepackage{amsfonts}       % blackboard math symbols
\usepackage{nicefrac}       % compact symbols for 1/2, etc.
\usepackage{microtype}      % microtypography
\usepackage{xcolor}         % colors

\usepackage{array}
\usepackage{booktabs} 
\usepackage{bm}

\usepackage{soul}

\usepackage{xcolor}
\hypersetup{
    colorlinks,
    linkcolor={red!25!black},
    citecolor={blue!50!black},
    urlcolor={blue!30!black}
}

\definecolor{firooz}{HTML}{1da8ad}
\definecolor{firooz2}{HTML}{106a6e}

\definecolor{javad}{HTML}{1da8ad}   % slate blue
\definecolor{javad2}{HTML}{106a6e}  % dark purple-blue

\title{When Graph Structure Provably Helps Classification: Non-Asymptotic Recovery Guarantees}

\author{%
  Firooz Shahriari-Mehr\textsuperscript{1}\thanks{Equal contribution. Corresponding authors:
    \texttt{\{firooz, javada\}@chalmers.se}}\hspace{1.2em}%
  Javad Aliakbari\textsuperscript{1}\footnotemark[1]\hspace{1.2em}%
  Alexandre Graell i Amat\textsuperscript{1}\hspace{0.9em}% 
  Ashkan Panahi\textsuperscript{1,2}\\[2ex]
  \textsuperscript{1}Chalmers University of Technology, Gothenburg, Sweden \\
  \textsuperscript{2}% 
  University of Gothenburg, Gothenburg, Sweden
}

\newcommand{\bGamma}{\bm{\Gamma}}

\newcommand{\bdelta}{\bm{\delta}}
\newcommand{\bomega}{\bm{\omega}}

\newcommand{\btheta}{\bm{\theta}}

\newcommand{\btau}{\bm{\tau}}

\newcommand{\bA}{\boldsymbol{A}}

\newcommand{\bC}{\boldsymbol{C}}

\newcommand{\bE}{\boldsymbol{E}}

\newcommand{\bH}{\mathbf{H}}
\newcommand{\bI}{\mathbf{I}}

\newcommand{\bL}{\boldsymbol{L}}
\newcommand{\bM}{\mathbf{M}}
\newcommand{\bN}{\mathbf{N}}
\newcommand{\bP}{\mathbf{P}}
\newcommand{\bQ}{\mathbf{Q}}

\newcommand{\bS}{\mathbf{S}}
\newcommand{\bT}{\mathbf{T}}
\newcommand{\bU}{\boldsymbol{U}}

\newcommand{\bY}{\mathbf{Y}}
\newcommand{\bZ}{\mathbf{Z}}

\newcommand{\ba}{\boldsymbol{a}}

\newcommand{\bc}{\mathbf{c}}

\newcommand{\be}{\boldsymbol{e}}

\newcommand{\bg}{\mathbf{g}}

\newcommand{\bq}{\mathbf{q}}

\newcommand{\bs}{\mathbf{s}}
\newcommand{\bu}{\mathbf{u}}

\newcommand{\bx}{\boldsymbol{x}}

\newcommand{\bz}{\boldsymbol{z}}

\newcommand{\inner}[2]{\left\langle #1,\, #2 \right\rangle}

\newcommand{\bPi}{\bm{\Pi}}

\newcommand{\bbR}{\mathbb{R}}

\newcommand{\bbS}{\mathbb{S}}

\newcommand{\calA}{\mathcal{A}}
\newcommand{\calB}{\mathcal{B}}
\newcommand{\calC}{\mathcal{C}}

\newcommand{\calE}{\mathcal{E}}

\newcommand{\calG}{\mathcal{G}}

\newcommand{\calV}{\mathcal{V}}

\newcommand{\calY}{\mathcal{Y}}

\newcommand{\bzero}{\mathbf{0}}
\newcommand{\bone}{\mathbf{1}}

\begin{document}
\maketitle
\setcounter{footnote}{0} 

\begin{abstract}
The joint use of graph structure and node-specific information is central to transductive node classification, yet a fundamental theoretical question remains: when does graph structure actually help? We provide the first deterministic, non-asymptotic, distribution-free answer to this question. We build on nuclear-norm graph clustering and generalized convex clustering. Our formulation of the latter incorporates node features, partial labels, and general convex loss functions, making it suitable for transductive node classification. We analyze each problem separately to characterize when graph structure alone, or node information alone, is sufficient for perfect recovery. We then formulate a unified convex optimization problem through a shared atomic norm representation and prove that combining the two sources expands the range of conditions under which perfect recovery is guaranteed beyond what either source alone can certify. This shows when node information can improve clustering and graph structure can improve classification.
\end{abstract}

\section{Introduction}
\label{sec:intro}
Transductive node classification addresses a core problem in graph-based machine learning: given a graph with node features where only a subset of nodes are labeled, predict the labels of the remaining nodes. This setting underlies many  real-world applications including social networks, fraud detection, and citation networks~\citep{bhagat2011node, zhu2003semi, ying2018graph, hu2020open}. A common premise  is \emph{homophily}, the tendency of connected or structurally similar nodes to share labels~\citep{mcpherson2001birds,sen2008collective,namata2012query,mcauley2015image}. When graph structure is aligned with labels, it can provide useful information beyond node-specific information alone, a principle underlying the empirical success of many graph-based learning methods~\citep{zhu2003semi,kipf:2016semi,hamilton2017inductive,velicovic:2018gat}.

Despite this empirical success, a fundamental theoretical question remains: \emph{when does graph structure actually help?} That is, under what conditions can we rigorously guarantee that incorporating graph information improves node classification beyond what node information alone achieves?

This paper provides the first deterministic, non-asymptotic, distribution-free answer to this question. We develop a convex optimization framework that precisely characterizes when graph structure improves classification beyond node information alone. Our framework accommodates arbitrary deterministic graphs and general convex node-specific loss functions, going beyond existing asymptotic, model-dependent results for graph clustering~\citep{braun2022iterative, dreveton2023exact, yang2023weighted, braun2024vec}, which address the dual problem of when features help clustering rather than when graph structure helps classification.

Our approach builds on two well-established convex relaxations. Convex relaxations for graph clustering based on semidefinite formulations~\citep{korlakai2014graph} use the nuclear norm, a tight convex surrogate for matrix rank, to recover community structure. Convex clustering~\citep{hocking2011clusterpath,lindsten2011clustering,panahi2017clustering} is the standard convex relaxation of $k$-means and hierarchical clustering, replacing combinatorial objectives with a sum-of-norms penalty.
These problems have largely been studied separately and primarily in the context of clustering. Our contributions are as follows:
\vspace{-1ex}
\begin{list}{\labelitemi}{\leftmargin=1.5em}
\addtolength{\itemsep}{-0.2\baselineskip}
    \item \textbf{Deterministic guarantees for graph clustering.}
    When only graph structure is available, the problem reduces to graph clustering. We analyze nuclear-norm graph clustering without distributional assumptions on the graph, unlike prior work based on stochastic block models. We provide a sufficient condition for perfect recovery and show this condition is tight (Thms.~\ref{thm:graph_sufficient} and \ref{thm:graph_tight_upper}).

    \item \textbf{Recovery guarantees for generalized convex clustering.}
    When only node information is available, we build on convex clustering to address node classification, accommodating partial labels, constraints, and regularization. The resulting problem is the complete-graph network lasso, for which the only known recovery guarantees are vacuous: they require non-uniform edge weights that already distinguish within- from between-class pairs (Remark~\ref{rem:network_lasso}). We derive sufficient conditions for perfect recovery with a matching tightness result (Cor.~\ref{cor:node_only_nonempty} and Thm.~\ref{thm:node_only_tight}). 

    \item \textbf{A unified convex problem for the joint regime.}
    To analyze recovery from graph structure and node information jointly, we formulate a single convex optimization problem that couples the two problems. A naive coupling leads to a non-convex problem; we exploit the structure of the nuclear norm to maintain convexity.  To our knowledge, this is the first convex formulation that jointly denoises the graph and classifies its nodes from structure, features and partial labels.

    \item \textbf{Precise characterization of when graph structure helps.}
    Our main result (Thm.~\ref{thm:combined_recovery}) establishes non-asymptotic, distribution-free guarantees showing that using graph structure and node information jointly  certifies perfect recovery under strictly weaker conditions than using either alone. This reveals, for a given graph, the bidirectional synergy known asymptotically under generative models: node information improves clustering, while graph structure improves classification.
\end{list}
\vspace{-1ex}

Our results shed light on a fundamental principle underlying graph-based learning: graph structure can provably improve prediction beyond node-specific information alone. By providing the first non-asymptotic, distribution-free characterization of when  this improvement occurs, our work offers rigorous theoretical foundations for understanding the benefit of graph structure in node classification.

\vspace{-2ex}
\paragraph{Paper roadmap.} 
Sec.~\ref{sec:node_classification_problem} sets up the problem, defines perfect recovery, reviews graph clustering, and introduces  the generalization of convex clustering. Sec.~\ref{sec: perfect_recovery} establishes recovery guarantees for the graph-only and node-information-only regimes. Sec.~\ref{sec:combined} formulates a unified convex optimization problem for analyzing graph structure and node information jointly, and  shows that this achieves perfect recovery under strictly weaker conditions than using either source alone.

\vspace{-0.5ex}
\section{Related work}
\label{sec: Related_work}

\vspace{-0.5ex}
\paragraph{Graph clustering.} 
Graph clustering is classically studied under the stochastic block model (SBM)~\citep{holland1983stochastic, schaeffer2007graph, abbe2018community}, with exact-recovery thresholds known in several settings~\citep{abbe2015exact, wu2015achieving, chen2014clustering, korlakai2014graph}. Nuclear-norm relaxations provide a convex approach to this problem, with recovery guarantees that match information-theoretic limits in standard SBM settings~\citep{korlakai2014graph, hajek2016achieving}. However, these results are model-based and asymptotic. We provide deterministic, non-asymptotic guarantees for nuclear-norm graph clustering under arbitrary graphs, without assuming a generative model.

\vspace{-2ex}
\paragraph{Convex clustering and network lasso.} 
The network lasso \citep{hallac2015network} extends convex clustering \citep{hocking2011clusterpath,lindsten2011clustering,panahi2017clustering} to arbitrary graphs and node losses; its complete-graph instance is the generalized convex clustering problem we analyze. Recovery guarantees are available for specific losses, mainly the quadratic \citep{zhu2014convex,sun2021convex,chi2019recovering}, or, for general losses, under strict convexity and smoothness \citep{yagishita2024pursuit}. We provide guarantees for general convex losses, as needed to incorporate labels, together with a converse.

\vspace{-2ex}
\paragraph{Covariate-assisted graph clustering.}
The contextual stochastic block model (CSBM)~\citep{deshpande2018contextual} studies how graph structure and node features combine, with asymptotic guarantees under SBM graphs with parametric features  \citep{braun2022iterative, dreveton2023exact, yang2023weighted, braun2024vec}. Convex estimators combining a graph with covariates also exist \citep{yan2021covariate}, with error bounds under an SBM and a sub-Gaussian mixture. These results are probabilistic and address clustering. We instead give deterministic guarantees for classification with partially observed labels within one convex formulation whose limits recover the graph-only and node-information-only regimes, which lets us quantify the gain from each source.

\vspace{-2ex}
\paragraph{Node classification.}
Graph neural networks \citep{kipf:2016semi, hamilton2017inductive, velicovic:2018gat} and graph-based semi-supervised learning \citep{zhu2003semi, zhou2003learning} combine structure, features and labels empirically, but take the edge set as given. CSBM-based analyses characterize when graph convolution improves classification \citep{baranwal21a} and how this depends on homophily \citep{luan2023graph}, asymptotically and under generative models. We instead denoise the graph and classify within one convex problem formulation, with deterministic guarantees for the given instance.

\vspace{-0.5ex}
\section{Problem Setup: Graph clustering and generalized convex clustering}
\label{sec:node_classification_problem}
\vspace{-0.5ex}

Throughout the paper, we adopt \emph{perfect recovery} as the deterministic metric for comparing three regimes: graph structure alone, node information alone, and the two combined. Sec.~\ref{sec:problem_setup} introduces the problem, fixes notation, and defines perfect recovery.  Sec.~\ref{sec:atomic_graph_clustering_new} reviews  nuclear-norm graph clustering  (graph-only regime) and Sec.~\ref{sec:feature_only_clustering} introduces the generalized convex clustering problem (node-information-only regime, or node-only for short) for node classification. Sec.~\ref{sec:combined} couples them into a single convex optimization problem for analyzing graph structure and node information jointly.

\subsection{Problem setup and perfect recovery}
\label{sec:problem_setup}

Consider a graph $\calG=(\calV,\calE)$ with $n$ nodes, where $\calV$ is the set of nodes and $\calE$ is the set of edges. Let $\bA \in \{0,1\}^{n \times n}$ denote its adjacency matrix, where $A_{uv}=1$ if $(u,v)\in\calE$ and $A_{uv}=0$ otherwise. Each node $v\in\calV$ has an associated feature vector $\bx_v\in\mathbb{R}^{m_d}$ and a true label \(y_v\in\calY,\) where \(\calY\) is the finite set of class labels. A subset \(\calV_{\rm train}\subset\calV\) has observed labels \(\tilde y_v\) (which may be noisy), and the goal is to predict the labels of the remaining nodes \(\calV_{\rm test}:=\calV\setminus\calV_{\rm train}\). Denote by $\bz_v$  the information available at node $v$: $\bz_v=\bx_v$ for $v\in\calV_{\mathrm{test}}$ and $\bz_v=(\bx_v,\tilde y_v)$ for $v\in\calV_{\mathrm{train}}$. Each node is further associated with a parameter $\btheta_v\in\mathbb{R}^d$ and a proper closed convex function
\(
f_v:\mathbb{R}^d\to(-\infty,+\infty],
\)
(we omit the dependence of $f_v$ on $\bz_v$ for simplicity). Predicted labels are obtained through a decoding map $\phi:\mathbb{R}^d\to\calY$, so that a solution $\{\btheta^\star_v\}_{v\in\calV}$ (see Sec.~\ref{sec:feature_only_clustering}) induces $\hat y_v :=\phi(\btheta^\star_v)$ for $v\in\calV_{\rm test}$.  We further assume that the population $\calV$ is partitioned into $K$ latent classes $\calC_1,\ldots,\calC_K$ of sizes $n_1,\ldots,n_K$, and that the true label of each node \(y_v\) is determined by the latent class containing \(v\). 

\begin{remark}[Balanced clusters] For clarity, we present the analysis in the balanced case $n_i = n/K$, so that $K = n/n_i$. The results extend directly to imbalanced clusters by replacing $n_i$ with $\min_i n_i$, yielding worst-case guarantees controlled by the smallest class.
\end{remark}

\begin{definition}[Perfect recovery]
A convex problem achieves perfect recovery on an instance  \((\calG,\{\bz_v\}_{v\in\mathcal V})\) if it admits an optimal solution whose structure exactly encodes the latent partition \(\mathcal C_1,\ldots,\mathcal C_K\). The \emph{perfect recovery solution} takes different forms for recovery from graph structure alone and from node information alone, as specified in Secs.~\ref{sec:atomic_graph_clustering_new} and~\ref{sec:feature_only_clustering}.  For node classification, perfect recovery requires class parameters $\btheta_1^\star,\ldots,\btheta_K^\star$ such that
\(
\btheta^\star_v=\btheta_i^\star,
\; \forall v\in\calC_i,
\)
with correct decoding
\(
\phi(\btheta_i^\star)=y_v,
\; \forall v\in\calC_i.
\)
This guarantees
\(
\hat y_v = y_v,
\; \forall v\in\calV_{\rm test}.
\)
\end{definition}

\subsection{Graph-only regime: nuclear-norm clustering}
\label{sec:atomic_graph_clustering_new}
When only the adjacency matrix $\bA$ is available, recovering the latent partition reduces to a graph clustering problem: grouping nodes into classes based solely on the connectivity structure. We use the convex relaxation of~\cite{korlakai2014graph}, which seeks a low-rank, positive semi-definite approximation of $\bA$,
\begin{equation}\label{eq:LR_approx}
\min_{\bL \in \calB}
\ \|\bL-\bA\|_1 + \tau n\|\bL\|_*,
\quad
\calB :=
\left\{
\bL \in \mathbb{R}^{n\times n} :
\bL=\bL^\top \succeq 0,\;
0 \le L_{uv} \le 1,\ \forall u,v \in \calV
\right\}.
\end{equation}

\vspace{-1ex}
Here, $\|\cdot\|_1$ is the elementwise $\ell_1$ norm, and $\|\cdot\|_*$ is the nuclear norm,
$
\|\bL\|_* \;=\; \sum_{k=1}^n \sigma_k(\bL),
$
where \(\sigma_k(\bL)\) denotes the \(k\)-th singular value of \(\bL\). The parameter $\tau \ge 0$ controls the rank of $\bL$, as illustrated in Fig.~\ref{fig:gamma_tau_role}.

\vspace{-1.2ex}
\paragraph{Perfect recovery solution.}
The perfect recovery solution $\bL^\star$~\citep{jalali2011clustering, chen2014clustering} is a matrix that reveals which nodes belong to the same class, so that the latent partition can be read directly from its entries,
\begin{align}
\label{eq:optimal_L}
L^\star_{uv} =
\begin{cases}
1, & u, v \in \calC_i \text{ for some } i, \\
0, & \text{otherwise}.
\end{cases}
\end{align}
Equivalently, $\bL^\star$ is the adjacency matrix of the disjoint union of cliques induced by $\calC_1,\ldots,\calC_K$, with unit diagonal.

\subsection{Node-information-only regime: generalized convex clustering}
\label{sec:feature_only_clustering}

\begin{figure}
    \centering
    \includegraphics[width = 0.99\linewidth]{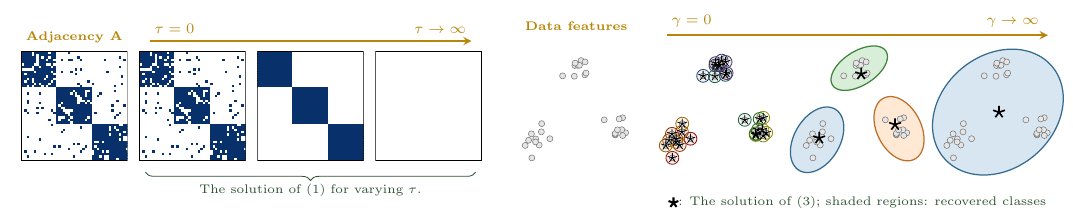}
    \vspace{-1.5ex}
    \caption{The effect of the rank parameter $\tau$ (left) and the fusion parameter $\gamma$ (right) in problems \eqref{eq:LR_approx} and \eqref{eq:classical_convex_clustering}. As $\tau$ grows, the optimal $\bL^\star$ collapses from the adjacency $\bA$ to a block-diagonal cluster indicator and finally to zero. As $\gamma$ grows, the optimal centroids collapse from one-per-point to the true clusters and finally to a single point.}
    \vspace{-1.8ex}       
    \label{fig:gamma_tau_role}
\end{figure}

When only node-specific information $\{\bz_v\}_{v\in\calV}$ is available, we must recover the latent partition from features and (potentially) labels alone. Convex clustering~\citep{hocking2011clusterpath,lindsten2011clustering,panahi2017clustering} associates each node $v$ with a parameter $\btheta_v\in\mathbb R^d$ and solves
\begin{equation}\label{eq:classical_convex_clustering}
\min_{\{\btheta_v\in \mathbb{R}^d\}_{v\in\calV}}
\ \sum_{v\in\calV} \tfrac{1}{2}\|\bx_v-\btheta_v\|_2^2
\;+\;
\frac{\gamma}{n}\sum_{u<v}\|\btheta_u-\btheta_v\|_2.
\end{equation}
The first term is the $k$-means data-fit loss. The second term is the sum-of-norms (SON) \emph{fusion} penalty, which couples the per-node parameters and promotes shared values across nodes. As $\gamma$ grows from $0$, the optimal parameters $\{\btheta_v^\star\}$ move from $\{\bx_v\}$ to progressively fewer shared values, inducing a partition of the nodes, as illustrated in Fig.~\ref{fig:gamma_tau_role}.

Problem~\eqref{eq:classical_convex_clustering} clusters features but cannot incorporate observed labels, parameter constraints, or node-wise regularization, all needed for transductive classification.   Inspired by the network lasso~\citep{hallac2015network, yagishita2024pursuit}, these are accommodated by replacing the quadratic loss with a proper closed convex function $f_v:\mathbb R^d\to(-\infty,+\infty]$, yielding the \emph{generalized convex clustering} problem
\begin{equation}\label{eq:node_info_only}
\min_{\{\btheta_v\in \mathbb{R}^d\}_{v\in\calV}}
\ \sum_{v \in \calV} f_v(\btheta_v)
\;+\;
\frac{\gamma}{n}\sum_{u<v}\|\btheta_u-\btheta_v\|_2.
\end{equation}
For instance, with data-fit loss $\ell_v(\bz_v;\btheta)$, the choice
$f_v=\ell_v+I_\Theta$ 
enforces $\btheta\in\Theta$ as $I_\Theta$ denotes the indicator function of $\Theta$, while
$f_v=\ell_v+\tfrac{\alpha}{2}\|\btheta\|_2^2$ 
adds ridge regularization. Setting
$f_v(\btheta)=\tfrac{1}{2}\|\bx_v-\btheta\|_2^2$
recovers~\eqref{eq:classical_convex_clustering}.
Observed labels enter through the node losses $f_v$, which is what separates classification from clustering. 
Sec.~\ref{sec: experiments} gives a concrete choice of $f_v$ and $\phi$.

\vspace{-1.2ex}
\paragraph{Perfect recovery solution.} The perfect recovery solution of~\eqref{eq:node_info_only} assigns a common parameter to all nodes in the same class,
$\btheta_v^\star=\btheta_i^\star\;\;\forall v\in\calC_i$, 
so that the latent partition can be read from the optimal parameters. 
Restricting~\eqref{eq:node_info_only} to such assignments gives the \emph{characteristic optimization problem}:
\begin{equation}\label{eq:characteristic_opt_features}
\{\btheta^\star_i\}_{i=1}^K
\;\in\;
\arg\min_{\{\btheta_i\}_{i=1}^K}
\ \sum_{i=1}^K n_i F_i(\btheta_i)
\;+\;\frac{\gamma}{n}\sum_{i<j} n_i n_j\,\|\btheta_i-\btheta_j\|_2,
\quad
F_i(\btheta):=\tfrac{1}{n_i}\sum_{v\in\calC_i} f_v(\btheta).
\end{equation}

The solutions $\{\btheta_i^\star\}$ are the \emph{biased centroids}: without the fusion term, each $\btheta_i^\star$ minimizes $F_i$, while the coupling term pulls the class parameters toward each other, introducing a $\gamma$-dependent bias.  Perfect recovery holds when the minimizer of~\eqref{eq:node_info_only} coincides with these class-level parameters.

\section{Recovery guarantees for the graph-only and node-information-only regimes}
\label{sec: perfect_recovery}
We establish perfect-recovery guarantees for the two problems introduced in Sec.~\ref{sec:node_classification_problem}. The guarantees are deterministic and hold for arbitrary graphs and node information, with no distributional assumptions beyond convexity of the node losses. These results characterize what each source of information can achieve in isolation and provide the baselines against which recovery from graph structure and node information jointly (Sec.~\ref{sec:combined}) is compared. All proofs are deferred to Apps.~\ref{appendix:graph_proofs} and~\ref{appendix:node_proofs}.

%=====================================================================
\subsection{Recovery from graph structure alone}
\label{sec:graph_only}
%=====================================================================

\begin{figure}
    \centering
    \includegraphics[width = 0.99\linewidth]{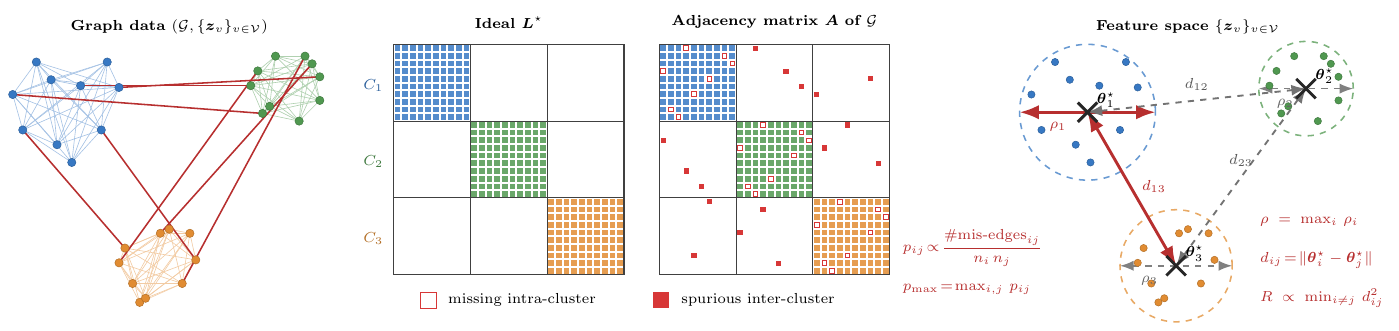}
    \vspace{-0.25ex}
    \caption{
    Geometric intuition for the recovery parameters.
\textit{Left:} ideal partition matrix $\bL^\star$ vs.\ observed adjacency $\bA$ with missing (hollow red) and spurious (filled red) entries that drive $p_{\max}$ (Eq.~\eqref{eq:p_def}). 
\textit{Right:} biased centroids $\btheta_i^\star\,(\boldsymbol\times)$ illustrating $\rho$ and $R$ specialized to the $k$-means loss: $\rho_i$ is the cluster diameter (Remark \ref{cor:kmeans_rho}), and the centroid gaps $d_{ij}=\|\btheta_i^\star-\btheta_j^\star\|$ govern $R$-separability via $R\propto\min_{i\neq j}d_{ij}^{2}$  (Remark \ref{cor:strong_convexity_R_separability}).
    }
    \vspace{-1.8ex}
    \label{fig:params_}
\end{figure}

We analyze the graph clustering problem~\eqref{eq:LR_approx}. The recovery conditions depend on graph parameters quantifying how far $\bA$ departs from ideal block structure $\bL^\star$ defined in~\eqref{eq:optimal_L}, as depicted in Fig.~\ref{fig:params_}.

\vspace{-1.2ex}
\paragraph{Graph parameters.}

For clusters $\calC_1,\ldots,\calC_K$ and a node $v\in\calC_i$, let
\(
n_{v,j} \coloneqq \sum_{u\in\calC_j} A_{uv}
\)
denote the number of edges from $v$ to cluster $\calC_j$.  Under perfect cluster separation, every within-class pair is connected and every between-class pair is not. We quantify deviations from this ideal through the \emph{misconnection count} \(\bar n_{v,j}\coloneqq n_i-n_{v,j}\) if \(j=i\), and
\(\bar n_{v,j}\coloneqq n_{v,j}\) if \(j\neq i\), 
which count missing intra-cluster edges (when $j=i$) and  spurious inter-cluster edges (when $j\neq i$). Aggregating over $\calC_i$, define
\(\bar N_{i,j}\coloneqq \sum_{v\in\calC_i}\bar n_{v,j}\) 
and
\(\bar\rho_{i,j}\coloneqq \bar N_{i,j}/(n_i n_j)\). 
The block-level rate
\(\bar\rho_{i,j}\) 
measures the fraction of misconnected edges between \(\calC_i\) and \(\calC_j\), and vanishes under perfect cluster separation. 
We bound local deviations from these block averages by
\(
\delta_{ij}:=\max_{v\in\calC_i}\left|\bar\rho_{i,j}-\bar n_{v,j}/n_j\right|.
\)
Combining $\bar\rho_{i,j}$ with the deviation $\delta_{ij}$, we define the \emph{graph misconnection rate}
\begin{equation}\label{eq:p_def}
p_{ij} \;\coloneqq\; \bar\rho_{i,j} + 2\delta_{ij},
\qquad
p_{\max} \;\coloneqq\; \max_{i,j} \ p_{ij}.
\end{equation}
The scalar $p_{\max}$ summarizes how well the graph structure aligns with the latent partition: it is small when most within-class pairs are connected and most between-class pairs are not, and equals zero for a perfectly separated graph.

\paragraph{Recovery guarantee.} Thm.~\ref{thm:graph_sufficient} gives a sufficient condition for perfect recovery in terms of $p_{\max}$, and Thm.~\ref{thm:graph_tight_upper} shows that the upper bound on $\tau$ in this condition is tight.

\begin{theorem}\label{thm:graph_sufficient}
Perfect recovery in the graph-only regime is achievable whenever
\[
p_{\max}\le p_K^\star
\coloneqq\frac{(K+4)-\sqrt{(K+4)^2-16}}{8}.
\]
In this case, the ideal solution $\bL^\star$ defined in Sec.~\ref{sec:atomic_graph_clustering_new} is a global minimizer of~\eqref{eq:LR_approx} for every
\begin{equation}\label{eq:graph_sufficient_cond}
0\le\tau\le\frac{1-2p_{\max}}{K}.
\end{equation}
For large $K$, $p_K^\star\sim1/(K+4)$.
\end{theorem}

% \end{corollary}

We next show that the bound of the interval~\eqref{eq:graph_sufficient_cond} is tight: there exists a graph consistent with $p_{\max}$ on which perfect recovery fails above the interval.

\begin{theorem}\label{thm:graph_tight_upper}
Fix a class $\calC_i$ and  $p_{\max}\in[0,1/2)$. There exists a graph configuration with $p_{ii}=p_{\max}$ such that, if
\(
\tau > \frac{1}{K}\,(1-2p_{\max}),
\)
then $\bL^\star$ is not a minimizer of~\eqref{eq:LR_approx}, and perfect recovery fails.
\end{theorem}

\begin{remark}\label{rem:sbm}
Problem~\eqref{eq:LR_approx} is, up to a reparameterization of the regularization weight, the problem analyzed by~\cite{korlakai2014graph}. For balanced clusters and full observation, our upper limit $\nicefrac{(1-2p_{\max})}{K}$ on $\tau$ coincides with the upper limit of their admissible window, see App.~\ref{appendix:remark4.4}. The lower bounds differ, the price of a guarantee for every graph rather than with high probability under a generative model.
\end{remark}

%=====================================================================
\subsection{Recovery from node information alone}
\label{sec:node_only}
%=====================================================================
We now analyze the generalized convex clustering
problem~\eqref{eq:node_info_only}, which uses only node information
$\{\bz_v\}_{v\in\calV}$. Recovery depends on two quantities:
within-class variability, measuring how similar nodes in the same class
are, and inter-class separation, measuring how distinguishable different
classes are.

\paragraph{Within-class variability.}  Let $\bomega_i$ denote the subgradient of the fusion term in~\eqref{eq:characteristic_opt_features} with respect to $\btheta_i$, scaled by $1/n_i$ and evaluated at the biased centroids:
\[
\bomega_i 
\;:=\;
\frac{1}{n_i}\frac{\partial}{\partial \btheta_i}
\left(
\frac{\gamma}{n}\sum_{l<j} n_l n_j\,\|\btheta_l-\btheta_j\|_2
\right)\bigg|_{\btheta=\btheta^\star}
\;=\;
\frac{\gamma}{n}
\sum_{j\neq i}
n_j\,
\frac{\btheta_i^\star-\btheta_j^\star}{\|\btheta_i^\star-\btheta_j^\star\|_2}.
\]
The first-order optimality of~\eqref{eq:characteristic_opt_features} then implies $-\bomega_i \in \partial F_i(\btheta_i^\star)$.

Define
\[
\rho_i
:=
\inf_{\substack{
\bs_v\in\partial f_v(\btheta_i^\star),\ v\in\calC_i\\
\frac{1}{n_i}\sum_{v\in\calC_i}\bs_v=-\bomega_i
}}
\;
\sup_{u,v\in\calC_i}
\|\bs_v-\bs_u\|_2\,.
\]
If each $f_v$ is differentiable, this reduces to
\(
\rho_i
=
\sup_{u,v\in\calC_i}
\|
\nabla f_v(\btheta_i^\star)
-
\nabla f_u(\btheta_i^\star)
\|_2, 
\; 
\rho \coloneqq \max_i \ \rho_i.
\)

Smaller $\rho$ means nodes within the same class are more similar to each other, while larger $\rho$ reflects greater within-class heterogeneity.

\begin{remark}[Within-class variability for quadratic loss function]
\label{cor:kmeans_rho}
For the squared loss
\(f_v(\btheta)=\tfrac12\|\bx_v-\btheta\|_2^2\),
the variability parameter reduces to the cluster  diameter, as illustrated in Fig.~\ref{fig:params_}:
\[
\rho_i
=
\sup_{u,v\in\calC_i}\|\bx_u-\bx_v\|_2,
\qquad
\rho=\max_i \rho_i.
\]
\end{remark}

\paragraph{Node-wise separability.}

For each $v\in\calC_i$, let $\bs_v\in\partial f_v(\btheta_i^\star)$ be a node-level subgradient. The node-wise Bregman divergence is defined as
\begin{equation*}
D_{f_v}^{\bs_v}(\btheta,\btheta_i^\star)
:=
f_v(\btheta)
-
f_v(\btheta_i^\star)
-
\left\langle
\bs_v,\btheta-\btheta_i^\star
\right\rangle .
% \label{eq:node_bregman_def}
\end{equation*}
Since $\bs_v\in\partial f_v(\btheta_i^\star)$, we have
\(
\quad
D_{f_v}^{\bs_v}(\btheta,\btheta_i^\star)\ge0,
\qquad
\forall \btheta\in\operatorname{dom}(f_v).
\)

We say that the centroids $\{\btheta_i^\star\}_{i=1}^K$ are \emph{node-wise $R$-separable} if, for every $\btheta$, in the common effective domain
\(
\Theta_f:=\bigcap_{v\in\calV}\operatorname{dom}(f_v)
\),
there exists at most one class $i_0\in[K]$ such that 
\begin{equation}
D_{f_v}^{\bs_v}(\btheta,\btheta_j^\star)>R,
\qquad \forall v\in\calC_j,\ \forall j\neq i_0,
\label{eq:nodewise_separability}
\end{equation}
that is, at most one class can contain a node whose Bregman gap to its own centroid is at most $R$.

\begin{remark}
\label{cor:strong_convexity_R_separability}

Assume that, for every \(v\in\mathcal C_i\), the loss \(f_v\) is
\(m\)-strongly convex on \(\operatorname{dom}(f_v)\). Let
\(
d_{\min}
:=
\min_{i\neq j}
\|\btheta_i^\star-\btheta_j^\star\|_2.
\)
Then the biased centroids \(\{\btheta_i^\star\}_{i=1}^K\) are node-wise \(R\)-separable for every 
$R<m d_{\min}^2/8$.
\end{remark}

\paragraph{Recovery guarantee.} We  give a sufficient condition for perfect recovery: a lower bound on \(\gamma\), which ensures within-class fusion, and an upper bound on $\gamma$, preventing different class representatives from collapsing.

\begin{theorem}
\label{thm:node_only_sufficient}
Consider the generalized convex clustering problem~\eqref{eq:node_info_only}. For each class
\(\calC_i\), 
let
\(
\btheta_i^{(0)}\in\arg\min_{\btheta} F_i(\btheta)
\)
be the unregularized class representative, and define
\begin{align}
\label{eq:dmin}
d_{\min}^{(0)}
:=
\min_{i\neq j}
\|\btheta_i^{(0)}-\btheta_j^{(0)}\|_2.
\end{align}
Assume that each \(F_i\) is \(m\)-strongly convex, with \(m>0\). If
\begin{equation}
\label{eq:node_sufficient_cond}
K\rho
\le
\gamma
<
{m\,d_{\min}^{(0)}},
\end{equation}
then problem~\eqref{eq:node_info_only} admits a class-wise
constant global minimizer,
\(
\btheta_v^\star=\btheta_i^\star,
\forall v\in\calC_i,
\)
with distinct class representatives \(\btheta_i^\star\neq\btheta_j^\star\), for all \(i\neq j\).
\end{theorem}
\begin{corollary}
\label{cor:node_only_nonempty}
Let 
\(
\mathrm{SVR}:=d_{\min}^{(0)}/\rho
\)
be the separation-to-variability ratio. 
For $m$-strongly convex functions, perfect recovery in the node-information-only regime is achievable whenever the interval~\eqref{eq:node_sufficient_cond} is nonempty, namely whenever
\begin{align}
\label{eq:node_svr}
\mathrm{SVR}
>
\frac{K}{m}\,,
\end{align}
i.e., the class separation dominates the within-class variability.
\end{corollary}

\begin{remark}
The lower bound \(K\rho\) in~\eqref{eq:node_sufficient_cond} controls within-class fusion and applies to general convex  losses. The upper bound uses \(m\)-strong convexity of the class losses \(F_i\) to prevent collapse of distinct class representatives.
\end{remark}

The next result shows that the bounds of the interval in~\eqref{eq:node_sufficient_cond} are essentially tight.

\begin{theorem}
\label{thm:node_only_tight}
Assume balanced classes. Suppose the node-loss family $f_v$ is subgradient-rich at the biased centroids \(\btheta_i^\star\),\footnote{The precise meaning of subgradient-richness is given in App.~\ref{appendix:node_only_tight}. Crucially, this assumption is mild for common losses.} If $K\rho_i>\gamma$ for some class \(\calC_i\), then there exists a data configuration with within-class variability \(\rho_i\) for which perfect recovery fails.

Moreover, if the loss family attains its modulus $m$ and $\Theta_f$ contains a regular simplex with $K$ vertices, there exist data with $\rho=0$ and $d_{\min}^{(0)}>0$ for which all classes collapse whenever $\gamma\ge m d_{\min}^{(0)}$. Thus the strict upper bound $\gamma<m d_{\min}^{(0)}$ is tight as a uniform sufficient condition.
\end{theorem}

Together, Thms.~\ref{thm:node_only_sufficient} and~\ref{thm:node_only_tight} reveal a fundamental tension in the node-information-only regime: $\gamma$ must be large enough to fuse within classes, yet small enough to keep classes separated.

\begin{remark}\label{rem:network_lasso}
Problem~\eqref{eq:node_info_only} is the network lasso on the complete graph with uniform weights. There, the condition of \citet{yagishita2024pursuit} requires $\alpha_k > 2L(K-1)\,n_k$, where $\sum_{v\in\calC_k} f_v$ is $\alpha_k$-strongly convex and $L$-smooth, which is impossible for $K \ge 2$ since $\alpha_k \le n_k L$. Thm.~\ref{thm:node_only_sufficient} requires no such condition and holds for general proper closed convex losses.
\end{remark}

\section{Recovery from graph structure and node information jointly}
\label{sec:combined}
%=====================================================================

To analyze recovery from graph structure and node information jointly, we formulate a single convex optimization problem that couples the graph clustering problem~\eqref{eq:LR_approx} and the generalized convex clustering problem~\eqref{eq:node_info_only}.  All proofs are deferred to App.~\ref{sec:appendix:perfect}.

\subsection{Unified convex optimization problem}
\label{sec:unified_estimator}

A natural way to couple the graph clustering problem~\eqref{eq:LR_approx} and the generalized convex clustering problem~\eqref{eq:node_info_only} is to weight the fusion term by graph similarity, e.g.,
\(
\;\sum_{u<v} L_{uv}\|\btheta_u-\btheta_v\|_2.
\)
However, this leads to a non-convex problem in the joint variables $(\bL,\{\btheta_v\})$. To obtain a convex formulation, we instead exploit the atomic decomposition underlying the nuclear norm in~\eqref{eq:LR_approx}: any $\bL\succeq 0$ can be written as a nonnegative combination of rank-one atoms $\be_i\be_i^\top$, and we use the same coefficients and atoms to parameterize the node variables, coupling the two problems.

\paragraph{Atomic representation.}
Define the atomic set
\(
\mathcal A
:=
\left\{
(\be\be^\top,\btheta)
:\;
\be\in\mathbb S^{n-1},\ \btheta\in \Theta_f
\right\}.
\)
Any matrix $\bL\succeq 0$ can be decomposed as
\begin{equation}
\label{eq:atomic_L}    
\bL=\sum_i\lambda_i\be_i\be_i^\top,
\qquad
(\be_i\be_i^\top,\btheta_i)\in\mathcal A,
\qquad
\lambda_i\ge 0.
\end{equation}
This decomposition expresses $\bL$ as a conic combination of rank-one atoms.
We use the same coefficients $\{\lambda_i\}$ to parameterize the node variables. Specifically, we require each node parameter $\btheta_v$ to be a convex combination of atom-level parameters,
\begin{equation}
\label{eq:convex_comb}
\btheta_v
=
\sum_i \lambda_i\,e_{i,v}^2\,\btheta_i,
\ \forall v\in\calV,
\qquad
\sum_i \lambda_i\,e_{i,v}^2 = 1.
\end{equation}
The normalization constraint ensures that the coefficients form a valid convex combination and is equivalent to enforcing $L_{vv}=1$ for all $v$.
Under this constraint,
\(
\|\bL\|_* = \mathrm{tr}(\bL) = \sum_i \lambda_i = n,
\)
so the nuclear-norm penalty in~\eqref{eq:LR_approx} becomes a constant and can be omitted.

\paragraph{Unified problem.}
Combining the graph-only objective~\eqref{eq:LR_approx} (without its constant nuclear-norm term) with an atom-wise lift of the node-only loss~\eqref{eq:node_info_only}, and enforcing the shared atomic structure~\eqref{eq:atomic_L}--\eqref{eq:convex_comb}, yields the joint convex optimization problem
\begin{gather}
\min_{\substack{
\bL\in\calB',\ \{\btheta_v\}_{v\in\calV},\\
\{\lambda_i\ge0,\ (\be_i\be_i^\top,\btheta_i)\in\mathcal A\}_i
}}
\quad
\|\bL-\bA\|_1
+
\mu n\!\left(
\sum_{v\in\calV}
\sum_i
\lambda_i\,e_{i,v}^2
f_v(\btheta_i)
+
\frac{\gamma}{n}\sum_{u<v}\|\btheta_u-\btheta_v\|_2
\right)
\nonumber 
\\[0.5ex]
\text{s.t.}\quad
\bL = \sum_i \lambda_i\,\be_i\be_i^\top,
\qquad
\btheta_v = \sum_i \lambda_i\,e_{i,v}^2\,\btheta_i,
\quad \forall v\in\calV,
\label{eq:main_model}
\end{gather}
where $\calB' := \calB \cap \{\bL : L_{vv}=1\ \forall v\}$.
The parameter $\mu\ge0$ controls the relative weight of node-specific information and graph structure: as $\mu\to 0$, the problem reduces to the graph-only problem~\eqref{eq:LR_approx}, 
while for $\mu\to\infty$ the node-information term dominates and the formulation approaches a lifted convex surrogate of the node-information-only problem~\eqref{eq:node_info_only}.

\subsection{Perfect recovery solution}
The perfect recovery solution uses \(K\) atoms, one per class,
\begin{equation}
\lambda_i^\star=n_i,\quad
e_{i,v}^\star=\tfrac{1}{\sqrt{n_i}}\mathbf{1}\{v\in\mathcal C_i\},\quad
\be_i^\star=(e_{i,v}^\star)_{v\in\mathcal V}.
\label{eq:gen_ideal_atoms}
\end{equation}
This gives the partition matrix defined in \eqref{eq:optimal_L}, and the class representatives \(\{\btheta_i^\star\}_{i=1}^K\) are the solutions of the characteristic problem in \eqref{eq:characteristic_opt_features}.

\subsection{Recovery guarantee}

We now characterize when using graph structure and node information jointly achieves perfect recovery. The key quantity is a \emph{residual price} that captures the within-class variability not absorbed by the fusion term, and determines how much graph noise can be tolerated for recovery to succeed.

\paragraph{Residual price.}

Let
\(
r_i := \left(\rho_i - \frac{\gamma}{K}\right)_+,
r_{\max} := \max_i r_i.
\)
The quantity $r_i$ depends on the geometry of the node losses and measures how much the within-class variability $\rho_i$ exceeds the fusion strength $\gamma/K$: when $\gamma/K \ge \rho_i$ the residual vanishes, otherwise a positive remainder persists.  To convert this geometric residual into an optimization-relevant quantity, we define the \emph{residual-price coefficient}
\begin{equation}\label{eq:zeta}
\zeta \;:=\; \max_i \sup_{\boldsymbol{\theta} \in \Theta_f} \frac{\|\boldsymbol{\theta} - \boldsymbol{\theta}_i^\star\|_2^2}{4 \min_{v \in \mathcal{C}_i} D^{\mathbf{s}_v}_{f_v}(\boldsymbol{\theta}, \boldsymbol{\theta}_i^\star)},
\end{equation}
where $\bs_v\in\partial f_v(\btheta_i^\star)$.
We use the conventions $0/0:=0$ and $c/0:=+\infty$ for $c>0$. The coefficient $\zeta$ measures how strongly the Bregman gaps control squared distance from the class representatives.

The following theorem is our main result. It shows that the unified problem~\eqref{eq:main_model} certifies perfect recovery under strictly weaker conditions than either the graph-only or node-information-only regime alone, establishing that graph structure provably helps node classification, and that node information provably helps graph clustering.

\begin{theorem}\label{thm:combined_recovery}
Consider~\eqref{eq:main_model} with $0<\mu<\infty$ and $\zeta<\infty$ on a graph with misconnection rate $p:=p_{\max}<1/4$, node variability $\rho$, node-wise $R$-separable biased centroids, and balanced classes. Let 
$b:={p}/{(1-p)}$,
$T:=\bigl(\sqrt{K\mu R}-\sqrt{1-4p}\,\bigr)_+^2$, and
\[
\hat B(R,p):=1-4p-\frac{2(K-1)\,b^2}{T-(K-2)b+\sqrt{T^2-2(K-2)b\,T+K^2b^2}} .
\]
The perfect recovery solution is a global minimizer of~\eqref{eq:main_model} whenever
\begin{equation}\label{eq:combined_single}
K\mu\zeta\, r_{\max}^2\le \hat B(R,p).
\end{equation}
\end{theorem}

Thm.~\ref{thm:combined_recovery} reduces the combined certificate to a single scalar condition. The left-hand side of~\eqref{eq:combined_single} is the \textbf{residual price} paid by the node term. It vanishes when fusion alone absorbs the variability ($\rho\le\gamma/K$) and otherwise grows linearly in $\mu$. The right-hand side is the \textbf{residual budget} that the graph certificate can absorb. This budget is continuous and nondecreasing in $K\mu R$. It equals the graph-only budget
\[
B_{\rm graph}(p):=\frac{1-(K+4)p+4p^2}{1-p}
\]
whenever
$K\mu R\le 1-4p$, and approaches $1-4p$ as $K\mu R\to\infty$.
The theorem recovers both single-source regimes as limits in $\mu$, and it certifies strictly more for intermediate $\mu$.

\begin{corollary}[Graph-only regime]\label{cor:combined_graph_only}
As $\mu\to0$, condition~\eqref{eq:combined_single} holds for all sufficiently small $\mu$ if
$p_{\max}<p_K^\star:=\frac{K+4-\sqrt{(K+4)^2-16}}{8}$, 
and fails for all sufficiently small 
$\mu$ if $p_{\max}>p_K^\star$.
\end{corollary}

Once $K\mu R\le1-4p$, $\hat B=B_{\rm graph}(p)$, and the residual price vanishes as $\mu\to0$. Condition~\eqref{eq:combined_single} therefore reduces to $B_{\rm graph}(p)\ge0$, which is exactly the graph-only threshold of Cor.~\ref{cor:combined_graph_only}.

\begin{corollary}[Node-information-only regime]\label{cor:combined_node_only}
Let $R>0$. As $\mu\to\infty$, condition~\eqref{eq:combined_single} holds for all sufficiently large $\mu$ if $\rho\le\gamma/K$, and fails for all sufficiently large $\mu$ if $\rho>\gamma/K$.
\end{corollary}

As $\mu\to\infty$, the budget $\hat B$ saturates at $1-4p$, while the residual price diverges unless $r_{\max}=0$. Therefore, only the fusion condition $\rho\le\gamma/K$ of Thm.~\ref{thm:node_only_sufficient} survives.

\begin{corollary}[Combined regime]\label{cor:combined_both}
Let $0 < \mu < \infty$, and define 
$s := 1-4p-B$, $b := p/(1-p)$,
\[
p_K^\star(B) \;:=\; \frac{(K+4-B) - \sqrt{(K+4-B)^2 - 16(1-B)}}{8},
\]
\[
R_{\min} \;:=\;
\begin{cases}
0, & p \le p_K^\star(B),\\[2pt]
\bigl(\sqrt{1-4p} + \sqrt{((K-1)b-s)(s+b)/s}\bigr)^2, & p > p_K^\star(B).
\end{cases}
\]
Perfect recovery is certified whenever there exists a budget $B \ge 0$ with
$\rho \le \gamma/K + \sqrt{B/(K\mu\zeta)}$ 
and at least one of the following conditions holds:
\begin{enumerate}
\item[\textup{(a)}] $p \le p_K^\star(B)$ 
with no additional restriction on $R\ge0$;
\item[\textup{(b)}] $p_K^\star(B) < p < (1-B)/4$ \ and \ $K\mu R \ge R_{\min}$.
\end{enumerate}
\end{corollary}

In case~(a), the graph alone helps node classification. The budget $B$ trades graph tolerance for variability tolerance: $B=0$ gives
$p\le p_K^\star$ with $\rho\le\gamma/K$, 
and larger $B$ admits
$\rho>\gamma/K$ 
at the price of a smaller threshold
$p_K^\star(B)<p_K^\star$. 
In case~(b), separation helps both tasks beyond their single-source regimes. The admissible misconnection rate extends from 
$p_K^\star=O(1/K)$ up to $(1-B)/4$, 
which approaches $1/4$ as $B\to0$. 
At the same time, $\rho$ may exceed
$\gamma/K$, 
and the required separation grows as $p$ approaches
$(1-B)/4$. 
For large separation, the admissible variability approaches
$\gamma/K+\sqrt{(1-4p)/(K\mu\zeta)}$. 
The combined certificate thus covers pairs $(p_{\max},\rho)$ beyond both single-source thresholds
(Fig.~\ref{fig:all}(left)).

\subsection{Strongly convex losses: improved SVR guarantees}

For the important class of strongly convex functions, we can specialize $\zeta$ and  rewrite the recovery condition as an admissible \(\gamma\)-window and a condition on the SVR (defined in Cor.~\ref{cor:node_only_nonempty}). (Extensions to other loss families, including bounded-domain convex losses, are discussed in App.~\ref{appendix:zeta_loss_families}.) 

\begin{theorem}\label{thm:combined_strongly_convex}
Under the setup of Thm.~\ref{thm:combined_recovery}, assume each $f_v$ is $m$-strongly convex. Then $\zeta\le1/(2m)$, and 
$R < m d_{\min}^2/8$.
Perfect recovery is certified whenever there exists a budget
$0\le B<1-4p_{\max}$ such that
\begin{equation}\label{eq:comb_gamma}
K\Bigl(\rho-\sqrt{\tfrac{2mB}{K\mu}}\Bigr)_+\;\le\;\gamma\;<\;\dfrac{m}{c_K}\Bigl(d_{\min}^{(0)}-\sqrt{\tfrac{8R_{\min}}{K \mu m}}\Bigr),
\end{equation}
where 
$c_K:=\frac12+\frac{\sqrt{K^2-3K+3}}{K}$ for $p > p_K^\star(B)$ 
and 
$c_K = 1$ for $p \le p_K^\star(B)$.
\end{theorem}

\begin{corollary}\label{cor:combined_svr}
Under the assumptions of Thm.~\ref{thm:combined_strongly_convex}, perfect recovery is certified whenever the interval~\eqref{eq:comb_gamma} is nonempty, or equivalently
\begin{equation}\label{eq:comb_svr}
\mathrm{SVR}>
\alpha\,c_K\frac{K}{m}+\beta, 
\qquad
\alpha:=\Bigl(1-\frac1\rho\sqrt{\tfrac{2mB}{K\mu}}\Bigr)_+,
\quad
\beta:=\frac1\rho\sqrt{\tfrac{8R_{\min}}{K \mu m}}.
\end{equation} 
In particular, choosing $K\mu\le 2mB/\rho^2$ gives $\alpha=0$,
while $\mu\to\infty$ gives $\beta\to0$. Therefore, optimizing over $\mu$, perfect recovery is
certified whenever 
\begin{equation}\label{eq:svr_star}
\mathrm{SVR}\;>\;\mathrm{SVR}^\star,
\qquad
\mathrm{SVR}^\star
\;:=\;\frac{K}{m}\,
\min\Bigl\{1,\;\frac{2}{K}\sqrt{\frac{R_{\min}}{B}}\Bigr\}.
\end{equation}
\end{corollary}

The lower endpoint of~\eqref{eq:comb_gamma} shows how the graph helps classification: the budget $B$ replaces $\rho$ by
$(\rho-\sqrt{2mB/(K\mu)})_+$, 
so less fusion is needed than in the node-only window $K\rho\le\gamma$. 
The upper endpoint shows the price. When 
$p_{\max}\le p_K^\star(B)$, 
no separation is needed, the upper endpoint is the node-only one, and 
$\mathrm{SVR}>\alpha K/m$ 
is weaker than $\mathrm{SVR}>K/m$ for every $B>0$; 
for small enough $\mu$ the SVR condition disappears. When 
$p_{\max}>p_K^\star(B)$, 
the graph borrows a separation $R_{\min}$ from the node term, which at finite $\mu$ costs the factor $c_K$ and the term $\beta$. Tuning $\mu$ trades $\alpha$ against $\beta$, and $\mathrm{SVR}^\star$ shows that the combined condition is never stronger than the node-only one, and is strictly weaker whenever $2\sqrt{R_{\min}/B}<K$. 
For instance, $K=10$, $p_{\max}=0.15$ and $B=0.25$ give $\mathrm{SVR}>9.6/m$, below $10/m$, at a misconnection rate where the graph alone fails.

Fig.~\ref{fig:all}(left) shows the recovery regions by evaluating the closed-form conditions of Thms.~\ref{thm:graph_sufficient},~\ref{thm:node_only_sufficient},~\ref{thm:combined_recovery}, and \ref{thm:combined_strongly_convex} on a grid of $(p_{\max},\rho)$ values for a strongly-convex setup. The dashed red and blue lines mark the graph-only threshold $p^\star_{\max}{=}0.117$ and the node-information-only ceiling $\rho{\leq}\gamma/K{=}0.4$, respectively: beyond these lines, the corresponding source alone cannot guarantee recovery.  For each finite $\mu$,  the combined condition already certifies recovery for pairs $(p_{\max},\rho)$ beyond both dashed lines, where neither single-source theorem does. The shaded curves show this region for different values of $\mu$, and the green boundary is their envelope over $\mu$, delineating the achievable region when using both sources jointly. Additional   sweeps and details are provided in App.~\ref{appendix:extended_sims}.

\section{Experiments}
\label{sec: experiments}

\begin{figure}[t]
  \centering
  \includegraphics[width=0.33\textwidth]{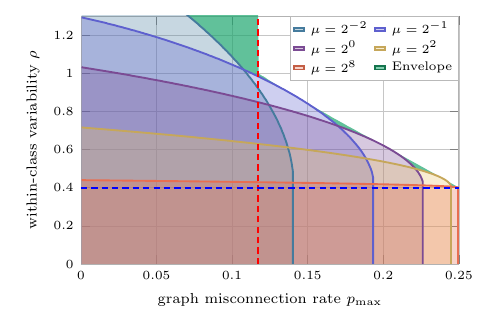}%
  \hfill
  \includegraphics[width=0.33\textwidth]{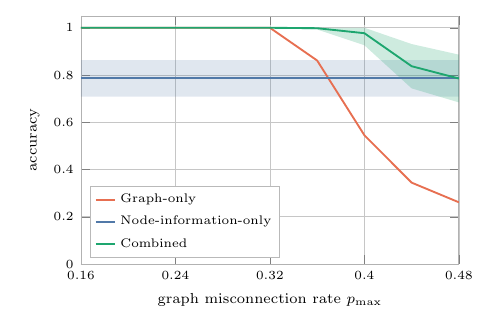}%
  \hfill
  \includegraphics[width=0.33\textwidth]{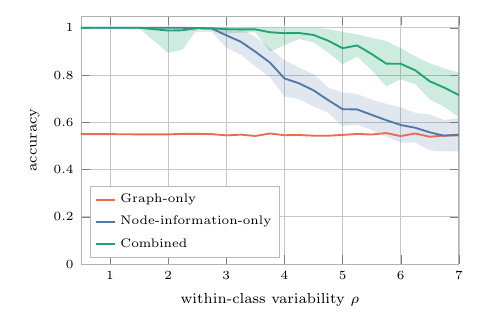}
  \vspace{-1ex}
  \caption{\textbf{Left:} Achievable recovery regions in the $(p_{\max},\rho)$ plane for
$K{=}5$, $\gamma{=}2$, $m{=}1$, $d_{\min}^{(0)}{=}4.7$.
Dashed red: graph-only threshold. Dashed blue: node-information-only
ceiling. Shaded curves: combined regime for different $\mu$. Green
boundary: envelope over all $\mu$. 
\textbf{Center:} Accuracy versus graph noise \(p_{\max}\) with fixed \(\rho\).
\textbf{Right:} Accuracy versus feature variability \(\rho\) with fixed \(p_{\max}\).
}
  \label{fig:all}
  \vspace{-1.9ex}
\end{figure}

Fig.~\ref{fig:all}(center and right) shows results for an optimization-based simulation using the CoGEnT-style atomic conditional-gradient solver described in App.~\ref{sec:algorithm}.%
\footnote{Code is available at \url{https://github.com/JavadAliakbari/Atomic-Conditional-Gradient-Solver}.}
We generate \(K=5\) clusters with \(25\) nodes per class in \(\mathbb R^2\), placing the class centroids on a circle of radius \(4\). Node features are sampled around each centroid, with the cluster diameter controlled by \(\rho\). The graph is generated with controlled graph noise, parameterized by the misconnection rate \(p_{\max}\). In the combined objective, each atom parameter  contains two blocks, $\bm{\theta}_i=(\bm{a}_i,\bm{\pi}_i)$: 
a feature centroid \(\ba_i\), which enters through the  quadratic feature-fit term, and an atom-to-class affinity vector $\bm{\pi}_i\in\Delta_{K-1}$, which uses the observed labels on the training set to encourage the assigned atom to predict the correct class. 
The experiment is thus the label-aware instance of the unified problem~\eqref{eq:main_model}, obtained by taking
\begin{equation}
\label{eq:exp-loss}
  f_v(\bm{\theta})
  \;=\; \tfrac{1}{2}\|\bm{x}_v-\bm{a}\|_2^{2}
  \;-\; \nu\,\mathbf{1}\{v\in\mathcal{V}_{\mathrm{train}}\}\,\pi_{\tilde{y}_v},
  \qquad \bm{\theta}=(\bm{a},\bm{\pi}),
\end{equation}
with $\nu>0$ weighting the label term, and decoding $\phi(\bm{\theta})=\arg\max_{c}\pi_c$.

With $5$ labeled nodes per class, we report accuracy on the 100 unlabeled nodes over 70 independent
trials (mean $\pm$ std), with $\nu=\gamma=2$ and $\mu$ chosen by 3-fold cross-validation on the labeled nodes.

We compare the node-information-only, graph-only, and the combined regimes. Fig.~\ref{fig:all}(center) fixes \(\rho=4\) and increases \(p_{\max}\): graph-only recovery degrades sharply as graph corruption grows, while the combined method remains accurate by using node information and labels to compensate for graph noise. Fig.~\ref{fig:all}(right) fixes \(p_{\max}=0.4\) and increases \(\rho\): node-only recovery deteriorates as features become less separated, whereas the combined method remains accurate over a larger range by using graph structure to stabilize the classification. These trends match Thm.~\ref{thm:combined_recovery}:  each source can fail in isolation, but using both jointly can still achieve perfect recovery.

\paragraph{Conclusion.} We established the first non-asymptotic, distribution-free guarantees characterizing when graph structure helps node classification through a novel unified convex optimization formulation. Using graph structure and node information jointly certifies perfect recovery under strictly weaker conditions than either alone,  making precise for a given graph how node information improves clustering and graph structure improves classification. These results provide rigorous theoretical foundations for the empirical success of graph-based learning.

\newpage
\begin{ack}
This work was partially supported by the Swedish Research Council (VR) under grants 2020-03687
and 2023-05065, by the Wallenberg AI, Autonomous Systems and Software Program (WASP) funded
by the Knut and Alice Wallenberg Foundation, and by the Swedish innovation Agency (Vinnova)
under grant 2022-03063.

The computations were enabled by resources provided by the National Academic Infrastructure for
Supercomputing in Sweden (NAISS), partially funded by the Swedish Research Council through
grant agreement no. 2022-06725.
\end{ack}

% \newpage
\bibliographystyle{plainnat}
\bibliography{refs}

\newpage

\appendix

\newpage
\section{Proofs for the Graph-Only Regime}
\label{appendix:graph_proofs}

Throughout this section we work with the polarized adjacency matrix $\bar\bA$ defined by $\bar A_{uv}=+1$ if $\{u,v\}\in\calE$ and $\bar A_{uv}=-1$ otherwise (with the convention $\bar A_{vv}=+1$). Up to additive constants in $\bA$, the graph-only program~\eqref{eq:LR_approx} is equivalent to
\begin{equation}\label{eq:appendix_primal}
\min_{\bL\in\calB} \ -\inner{\bar\bA}{\bL} + \tau n \|\bL\|_*,
\end{equation}
with $\calB=\{\bL=\bL^\top\succeq 0 : 0\le L_{uv}\le 1, \ \forall u,v\}$. The ideal solution $\bL^\star$ is the block-constant matrix presented in \eqref{eq:optimal_L}.

\subsection{Proof of Theorem~\ref{thm:graph_sufficient}}
\label{appendix:graph_sufficient}

Write $p:=p_{\max}$ and use the standing balanced-cluster assumption $n_i=n/K$. Set $\bar\tau:=(1-2p)/K$. The proof first constructs a dual certificate at $\tau=\bar\tau$ and then extends optimality to every $0\le\tau\le\bar\tau$.

\paragraph{KKT conditions at $\bL^\star$.}
Taking the multiplier for the PSD constraint to be zero, it suffices to find a symmetric matrix $\bY\in\tau n\,\partial\|\bL^\star\|_*$ together with box multipliers $\bU,\boldsymbol{V}\in\mathbb{R}^{n\times n}_{\ge 0}$ satisfying
\begin{equation}\label{eq:appendix_stationarity}
-\bar\bA + \bY - \bU + \boldsymbol{V} = 0,
\end{equation}
and complementary slackness $U_{uv}L^\star_{uv}=0$, $V_{uv}(1-L^\star_{uv})=0$. Since $L^\star_{uv}\in\{0,1\}$, complementary slackness reduces to: on intra-cluster entries (where $L^\star_{uv}=1$, including the diagonal) we have $U_{uv}=0$, hence
\begin{equation}\label{eq:appendix_intra_sign}
Y_{uv}\le \bar A_{uv},\qquad u,v\in\calC_i;
\end{equation}
on inter-cluster entries (where $L^\star_{uv}=0$) we have $V_{uv}=0$, hence
\begin{equation}\label{eq:appendix_inter_sign}
Y_{uv}\ge \bar A_{uv},\qquad u\in\calC_i,\ v\in\calC_j,\ i\neq j.
\end{equation}

Considering the subdifferential of the nuclear norm at the PSD matrix $\bL^\star$, to construct $\bY\in\tau n\,\partial\|\bL^\star\|_*$, it suffices to find a matrix $\bZ$ such that  
\[
\bY=\tau n\bPi+\bZ, \quad \bPi\bZ=\bZ\bPi=\bzero, \quad \|\bZ\|_2\le \tau n, 
\]
where $\bPi$ is the orthogonal projector onto $\operatorname{range}(\bL^\star)$, given by
$
\bPi
\!:=\!
\operatorname{blkdiag}\!\left(
\frac{1}{n_1}\bone\bone^\top,
\ldots,
\frac{1}{n_K}\bone\bone^\top
\right).
$

\paragraph{Centered certificate construction.}
For each pair of clusters $(i,j)$, define the centered block $\bC_{ij}\in\mathbb{R}^{n_i\times n_j}$ as follows. Let $\bM_{ii}\in\{0,1\}^{n_i\times n_i}$ have $(M_{ii})_{uv}=1$ if $u\neq v$ and $\{u,v\}\notin\calE$, and $0$ otherwise (the missing intra-cluster edges, with the diagonal treated as correct). Let $\bM_{ij}\in\{0,1\}^{n_i\times n_j}$ for $i\neq j$ have $(M_{ij})_{uv}=1$ if $\{u,v\}\in\calE$, and $0$ otherwise (the spurious inter-cluster edges). Set
\[
\bC_{ii} \;:=\; \bP_i^\perp\bM_{ii}\bP_i^\perp,
\qquad
\bC_{ij} \;:=\; \bP_i^\perp\bM_{ij}\bP_j^\perp,\quad i\neq j,
\]
where $\bP_i^\perp = \bI_{n_i}-\tfrac{1}{n_i}\mathbf{1}\mathbf{1}^\top$. By construction, each $\bC_{ij}$ has zero row and column sums.

Choose nonnegative scalars $\Delta_{ij}=\Delta_{ji}\ge 0$ and define $\bZ$ blockwise:
\begin{equation}\label{eq:appendix_Z_blocks}
\bZ_{ii} \;:=\; -\Delta_{ii}\bC_{ii},
\qquad
\bZ_{ij} \;:=\; \Delta_{ij}\bC_{ij},\quad i\neq j.
\end{equation}
Since each $\bC_{ij}$ has zero row and column sums, $\bPi\bZ=\bZ\bPi=0$ automatically.

\paragraph{Entrywise sign verification.}
Direct computation of $\bC_{ii}$ and $\bC_{ij}$ entrywise yields, 
\begin{equation*}
Z_{uv} \;=\;
\begin{cases}
\displaystyle
\Delta_{ii}\!\left(\tfrac{\bar n_{u,i}}{n_i}+\tfrac{\bar n_{v,i}}{n_i}-\bar\rho_{i,i}\right)
 & u,v\in\calC_i,\ u=v\ \text{or}\ \{u,v\}\in\calE,\\[1.2ex]
\displaystyle
\Delta_{ii}\!\left(\tfrac{\bar n_{u,i}}{n_i}+\tfrac{\bar n_{v,i}}{n_i}-\bar\rho_{i,i}-1\right)
 & u,v\in\calC_i,\ u\neq v,\ \{u,v\}\notin\calE,\\[1.2ex]
\displaystyle
\Delta_{ij}\!\left(-\tfrac{\bar n_{u,j}}{n_j}-\tfrac{\bar n_{v,i}}{n_i}+\bar\rho_{i,j}\right)
 & u\in\calC_i,\ v\in\calC_j,\ i\neq j,\ \{u,v\}\notin\calE,\\[1.2ex]
\displaystyle
\Delta_{ij}\!\left(1-\tfrac{\bar n_{u,j}}{n_j}-\tfrac{\bar n_{v,i}}{n_i}+\bar\rho_{i,j}\right)
 & u\in\calC_i,\ v\in\calC_j,\ i\neq j,\ \{u,v\}\in\calE.
\end{cases}
\end{equation*}
Using the deviation bound in Section \ref{sec:graph_only} and the definition of $p=p_{\max}$,
\[
\tfrac{\bar n_{u,i}}{n_i}+\tfrac{\bar n_{v,i}}{n_i}-\bar\rho_{i,i}\;\le\; p,
\qquad
\tfrac{\bar n_{u,j}}{n_j}+\tfrac{\bar n_{v,i}}{n_i}-\bar\rho_{i,j}\;\le\; p.
\]
Indeed, the inter-cluster expression is at most $\bar\rho_{i,j}+\delta_{ij}+\delta_{ji}\le\max\{p_{ij},p_{ji}\}\le p$.
Combined with $\bY=\tau n\bPi+\bZ$, the conditions~\eqref{eq:appendix_intra_sign}--\eqref{eq:appendix_inter_sign} are ensured by
\begin{equation}
\Delta_{ii} \;\in\; \Bigl[\tfrac{1+\tau n/n_i}{1-p},\;\tfrac{1-\tau n/n_i}{p}\Bigr],
\qquad
\Delta_{ij} \;\in\; \Bigl[\tfrac{1}{1-p},\;\tfrac{1}{p}\Bigr],\quad i\neq j.
\label{eq:appendix_inter_interval}
\end{equation}
For $p>0$, the intra-cluster interval is non-empty iff $\tau\le \nicefrac{n_i}{n}(1-2p)$, and the inter-cluster interval is non-empty iff $p\le 1/2$. When $p=0$, the upper bounds involving division by $p$ are omitted, and the sign conditions hold provided $\tau\le n_i/n$.

\paragraph{Spectral-norm bound.}
We claim that, with $\Delta_{ii}^\star:=\tfrac{1+\tau n/n_i}{1-p}$ and $\Delta_{ij}^\star:=\tfrac{1}{1-p}$ (the lower endpoints), $\|\bZ\|_2\le\tau n$ holds whenever $\tau\ge p/(1-2p)$.

Indeed, define the symmetric matrix $\bS$ by $\bS_{ii}=-\Delta_{ii}^\star\bM_{ii}$ and $\bS_{ij}=\Delta_{ij}^\star\bM_{ij}$ for $i\ne j$. Then $\bZ=(\bI-\bPi)\bS(\bI-\bPi)$. Every row sum of $\bM_{ij}$ is at most $n_jp$, so, using $n_i=n/K$,
\[
\|\bZ\|_2\le\|\bS\|_2
\le\max_u\sum_v|S_{uv}|
\le\frac{(1+\tau K)(n/K)p+(K-1)(n/K)p}{1-p}
=\frac{np+\tau np}{1-p}.
\]
The last expression is at most $\tau n$ whenever $\tau\ge p/(1-2p)$.

\paragraph{Putting it together.}
Since $p\le p_K^\star<1/2$, we have $Kp\le(1-2p)^2$, hence $\bar\tau\ge p/(1-2p)$. At $\tau=\bar\tau$, both intervals in \eqref{eq:appendix_inter_interval} are non-empty and the spectral-norm bound holds. 
By construction, $\bPi\bZ=\bZ\bPi=\bzero$, and the spectral-norm bound gives $\|\bZ\|_2\le\tau n$. Hence, by the nuclear-norm subdifferential characterization above,
$
\bY=\tau n \bPi+\bZ\in\tau n \,\partial\|\bL^\star\|_*.
$
Defining the box multipliers entrywise from $Y_{uv}-\bar A_{uv}$ via the sign conditions yields $\bU,\boldsymbol{V}\ge 0$ satisfying~\eqref{eq:appendix_stationarity} with complementary slackness. Hence $\bL^\star$ satisfies the KKT conditions of~\eqref{eq:appendix_primal} at $\tau=\bar\tau$, and is a global minimizer by convexity.

\paragraph{Extension to smaller regularization parameters.}
For every feasible $\bL\succeq0$, we have $\|\bL\|_*=\operatorname{tr}(\bL)\le n=\operatorname{tr}(\bL^\star)$. Write
\[
F_\tau(\bL):=-\inner{\bar\bA}{\bL}+\tau n\operatorname{tr}(\bL).
\]
For $0\le\tau\le\bar\tau$,
\begin{equation*}
F_\tau(\bL)-F_\tau(\bL^\star)
=F_{\bar\tau}(\bL)-F_{\bar\tau}(\bL^\star)
 +(\bar\tau-\tau)n\bigl(n-\operatorname{tr}(\bL)\bigr)
\ge0.
\end{equation*}
Thus $\bL^\star$ is a global minimizer throughout~\eqref{eq:graph_sufficient_cond}, including $\tau=0$.

\paragraph{Recovery threshold.}
\label{appendix:graph_threshold}

For $0\le p_{\max}<1/2$, the certificate at the upper endpoint $\bar\tau=(1-2p_{\max})/K$ is valid whenever
\[
\frac{p_{\max}}{1-2p_{\max}}\le\frac{1-2p_{\max}}{K},
\]
i.e., $Kp_{\max}\le(1-2p_{\max})^2$, which expands to $4p_{\max}^2-(K+4)p_{\max}+1\ge0$. The smaller root of the corresponding equality is
\[
p_K^\star=\frac{(K+4)-\sqrt{(K+4)^2-16}}{8}.
\]
On $0\le p_{\max}<1/2$, the inequality holds iff $p_{\max}\le p_K^\star$. The preceding proof then gives recovery for every $0\le\tau\le(1-2p_{\max})/K$. 
\qed

\subsection{Proof of Theorem~\ref{thm:graph_tight_upper}}
\label{appendix:graph_tight_upper}

Fix a class $\calC_i$ and consider the candidate matrix
\(
\bL^{(-i)} := \bL^\star - \mathbf{1}_{\calC_i}\mathbf{1}_{\calC_i}^\top
= \sum_{k\neq i} n_k\,\be_k^\star(\be_k^\star)^\top.
\)
Both $\bL^\star$ and $\bL^{(-i)}$ are symmetric, PSD, and entrywise in $[0,1]$, so $\bL^{(-i)}\in\calB$. Their nuclear norms equal their traces: 
$
\|\bL^\star\|_* = n,
\|\bL^{(-i)}\|_* = n - n_i.
$
For the linear term,
\[
\inner{\bar\bA}{\bL^\star-\bL^{(-i)}}
\;=\; \sum_{u,v\in\calC_i}\bar A_{uv}
\;=\; n_i^2 - 2\bar N_{i,i}
\;=\; n_i^2(1-2\bar\rho_{i,i}),
\]
using $\bar N_{i,i}=n_i^2\bar\rho_{i,i}$. Defining the objective $F(\bL):=-\inner{\bar\bA}{\bL}+\tau n\|\bL\|_*$,
\[
F(\bL^{(-i)})-F(\bL^\star)
\;=\; \inner{\bar\bA}{\bL^\star-\bL^{(-i)}} -\tau n\,n_i
\;=\; n_i\bigl[n_i(1-2\bar\rho_{i,i})-\tau n\bigr].
\]
Take a graph configuration with $\bar\rho_{i,i}=p_{\max}$ and $\delta_{ii}=0$, so $p_{ii}=p_{\max}$. If $\tau> \tfrac{n_i(1-2p_{\max})}{n}$, then $F(\bL^{(-i)})<F(\bL^\star)$, so $\bL^\star$ is not a minimizer. \qed

\subsection{\texorpdfstring
    {Comparison with \citet{korlakai2014graph} in Remark \ref{rem:sbm}}
    {Comparison with Korlakai et al. (2014) in Remark 4.3}}
\label{appendix:remark4.4}

The deviation argument preceding~\eqref{eq:appendix_inter_interval} gives  the  blockwise forms
\begin{equation}\label{eq:delta_blockwise}
\Delta_{ii}\in\left[\tfrac{1+\tau n/n_i}{1-p_{ii}},\;\tfrac{1-\tau n/n_i}{p_{ii}}\right],
\qquad
\Delta_{ij}\in\left[\tfrac{1}{1-p_{ij}},\;\tfrac{1}{p_{ij}}\right],\quad i\neq j.
\end{equation}

The intra-cluster interval is non-empty iff $\tau\le\tfrac{n_i}{n}(1-2p_{ii})$, so the ceiling is set by the intra-cluster rates alone. For the floor, take the lower endpoints $\Delta^\star_{ii}=\tfrac{1+\tau n/n_i}{1-p_{ii}}$ and $\Delta^\star_{ij}=\tfrac{1}{1-p_{ij}}$ and bound $\|\bZ\|_2$ blockwise. 
Projections are
non-expansive, so $\|\bC_{ij}\|_2\le\|\bM_{ij}\|_2$, and since $\bM_{ii}$ and $\bM_{ij}$ are
nonnegative with row sums at most $n_ip_{ii}$ and $n_jp_{ij}$ and column sums at most
$n_ip_{ij}$,
we have
\[
\|\bZ_{ii}\|_2\le\frac{n_ip_{ii}\,(1+\tau n/n_i)}{1-p_{ii}},
\qquad
\|\bZ_{ij}\|_2\le\frac{\sqrt{n_in_j}\,p_{ij}}{1-p_{ij}} .
\]

Writing $N_{ij}:=\|\bZ_{ij}\|_2$ and $\xi_i:=\|\bx_i\|_2$ for $\bx$ partitioned by cluster,
$|\bx^\top\bZ\bx|\le\boldsymbol\xi^\top\bN\boldsymbol\xi$ with
$\|\boldsymbol\xi\|_2=\|\bx\|_2$, so $\|\bZ\|_2\le\|\bN\|_2\le\max_i\sum_jN_{ij}$, the last
step because $\bN$ is symmetric and nonnegative. Thus $\|\bZ\|_2\le\tau n$ holds as soon as,
for every $i$,
\[
\frac{n_i}{n}\frac{p_{ii}}{1-p_{ii}}
+\tau\,\frac{p_{ii}}{1-p_{ii}}
+\sum_{j\neq i}\frac{\sqrt{n_in_j}}{n}\frac{p_{ij}}{1-p_{ij}}\;\le\;\tau ,
\]
and since $p_{ii}<\nicefrac12$ makes $1-\tfrac{p_{ii}}{1-p_{ii}}=\tfrac{1-2p_{ii}}{1-p_{ii}}$
positive, solving for $\tau$ gives the blockwise window
\begin{equation}
\max_{i}\left[
\frac{n_i}{n}\frac{p_{ii}}{1-2p_{ii}}
+\frac{1-p_{ii}}{1-2p_{ii}}\sum_{j\ne i}\frac{\sqrt{n_in_j}}{n}\frac{p_{ij}}{1-p_{ij}}\right]
\;\le\;\tau\;\le\;
\min_{i}\frac{n_i}{n}(1-2p_{ii}).
\label{eq:appendix_blockwise_window}
\end{equation}

The intra-cluster rates set the ceiling; the inter-cluster rates enter only the floor,
through a weighted sum over the $K-1$ off-diagonal blocks of cluster $i$. Bounding
$p_{ij}\le p_{\max}$ with $n_i=n/K$ collapses the floor to
$
% \tfrac{p_{\max}}{1-2p_{\max}}\bigl[\tfrac1K+\tfrac{K-1}{K}\bigr]=
\tfrac{p_{\max}}{1-2p_{\max}}$ and the ceiling to $\tfrac{1-2p_{\max}}{K}$, recovering the
interval~\eqref{eq:graph_sufficient_cond} of Thm.~\ref{thm:graph_sufficient}.

\paragraph{Specialization to an SBM.}
Let $n_i=n/K$ and write $\pi_{\mathrm{in}}:=1-\max_ip_{ii}$ for the within-cluster connection rate and
$q:=\max_{i\neq j}p_{ij}$ for the between-cluster one.
Then~\eqref{eq:appendix_blockwise_window} becomes
\begin{equation}
\frac{(1-\pi_{\mathrm{in}})(1-q)+\pi_{\mathrm{in}}(K-1)q}{K(2\pi_{\mathrm{in}}-1)(1-q)}
\;\le\;\tau\;\le\;
\frac{2\pi_{\mathrm{in}}-1}{K},
\label{eq:appendix_sbm_window}
\end{equation}
in which $\pi_{\mathrm{in}}$ and $q$ appear separately, rather than through
$p_{\max}=\max\{1-\pi_{\mathrm{in}},q\}$. This window is non-empty iff
$(1-\pi_{\mathrm{in}})(1-q)+\pi_{\mathrm{in}}(K-1)q\le(2\pi_{\mathrm{in}}-1)^2(1-q)$; using $(1-\pi_{\mathrm{in}})-(2\pi_{\mathrm{in}}-1)^2=\pi_{\mathrm{in}}(3-4\pi_{\mathrm{in}})$ and dividing by $\pi_{\mathrm{in}}>0$,
this is
\[
q\;\le\;\frac{4\pi_{\mathrm{in}}-3}{K+4\pi_{\mathrm{in}}-4},
\qquad\text{equivalently}\qquad
\pi_{\mathrm{in}}\;\ge\;\frac34+\frac{K-1}{4}\frac{q}{1-q},
\]
an inter-cluster budget of order $1/K$. Setting $q=1-\pi_{\mathrm{in}}$, so that $p_{\max}=1-\pi_{\mathrm{in}}$, this reduces
to $4p_{\max}^2-(K+4)p_{\max}+1\ge0$, recovering $p_K^\star$ of
Thm.~\ref{thm:graph_sufficient}.

\paragraph{Comparison with \citet{korlakai2014graph}.}
They analyze $\min_{\bL\in\calB}\|\bL-\bA\|_1+\tfrac{1}{\lambda}\|\bL\|_*$, which
is~\eqref{eq:LR_approx} with $\lambda=\nicefrac{1}{(\tau n)}$, and certify recovery with high
probability for $\nicefrac{1}{D_{\min}}\le\lambda\le\Lambda_{\rm succ}$, where
$D_i:=n_i\,r\,(2\pi_{\mathrm{in}}-1)$ is the effective density of cluster $\calC_i$, with $r$ the fraction of observed entries and $\pi_{\mathrm{in}}$ the within-cluster connection rate, and $D_{\min}:=\min_iD_i$.
In the parameterization of~\eqref{eq:LR_approx} their window is
$\nicefrac{1}{(\Lambda_{\rm succ}n)}\le\tau\le\nicefrac{D_{\min}}{n}$. 
For balanced clusters and full observation, $r=1$ and $n_i=n/K$, so
$
\nicefrac{D_{\min}}{n}=\nicefrac{2\pi_{\mathrm{in}}-1}{K},
$
which is exactly the ceiling in~\eqref{eq:appendix_sbm_window}, and equals
$\nicefrac{(1-2p_{\max})}{K}$ whenever $p_{\max}=1-\pi_{\mathrm{in}}$. The two upper limits therefore coincide, with the same constant. The floors differ: theirs decays with $n$, whereas ours depends only on the rates, which is the price of a guarantee that holds for every graph  rather than with high probability under a generative model.

\newpage
\section{Proofs for the Node-Only Regime}
\label{appendix:node_proofs}

This appendix collects the proofs of the recovery guarantees for the generalized convex clustering problem~\eqref{eq:node_info_only}. We use the notation introduced in Secs.~\ref{sec:feature_only_clustering} and~\ref{sec:node_only}: for a class~$\calC_i$, $F_i(\btheta)=\frac{1}{n_i}\sum_{v\in\calC_i}f_v(\btheta)$ is the class loss, $\{\btheta_i^\star\}_{i=1}^K$ are the biased centroids defined by~\eqref{eq:characteristic_opt_features}, $\bomega_i$ is the centered fusion subgradient appearing in the cluster-level first-order condition $-\bomega_i\in\partial F_i(\btheta_i^\star)$, and $\rho_i$ (resp.\ $\rho$) is the within-class variability of class~$\calC_i$ (resp.\ its maximum).
We begin with the two remarks specialising the variability and separation parameters, then prove the sufficient and tight conditions in turn.

\subsection{Proof of Remark~\ref{cor:kmeans_rho}}
\label{appendix:kmeans_rho}

For the squared loss $f_v(\btheta)=\tfrac12\|\bx_v-\btheta\|_2^2$, every $f_v$ is differentiable, so
\(
\nabla f_v(\btheta_i^\star)=\btheta_i^\star-\bx_v
\)
is the unique element of $\partial f_v(\btheta_i^\star)$. Substituting into the differentiable form of $\rho_i$ in Sec.~\ref{sec:node_only},
\[
\rho_i
=\sup_{u,v\in\calC_i}
\bigl\|\nabla f_v(\btheta_i^\star)-\nabla f_u(\btheta_i^\star)\bigr\|_2
=\sup_{u,v\in\calC_i}\|(\btheta_i^\star-\bx_v)-(\btheta_i^\star-\bx_u)\|_2
=\sup_{u,v\in\calC_i}\|\bx_u-\bx_v\|_2,
\]
which is exactly the cluster diameter. Taking the maximum over classes yields $\rho=\max_i\rho_i$.\qed

\subsection{Proof of Remark~\ref{cor:strong_convexity_R_separability}}
\label{appendix:strong_convexity_R_separability}

Assume each node loss $f_v$ is $m$-strongly convex on $\operatorname{dom}(f_v)$, and fix any $\btheta\in\Theta_f$. We show that $\btheta$ cannot be $R$-close, in the node-wise Bregman sense, to two distinct classes.

Suppose, toward a contradiction, that there exist distinct classes $\calC_i$ and $\calC_j$ together with nodes $v\in\calC_i$, $u\in\calC_j$ satisfying
\(
D_{f_v}^{\bs_v}(\btheta,\btheta_i^\star)\le R
\)
and
\(
D_{f_u}^{\bs_u}(\btheta,\btheta_j^\star)\le R.
\)
Since $f_v$ and $f_u$ are $m$-strongly convex, the standard lower bound
$D_{f_w}^{\bs_w}(\btheta,\btheta_k^\star)\ge \tfrac{m}{2}\|\btheta-\btheta_k^\star\|_2^2$ gives
$\|\btheta-\btheta_i^\star\|_2,\,\|\btheta-\btheta_j^\star\|_2\le\sqrt{2R/m}$.
The triangle inequality then yields
\[
\|\btheta_i^\star-\btheta_j^\star\|_2
\le \|\btheta-\btheta_i^\star\|_2+\|\btheta-\btheta_j^\star\|_2
\le 2\sqrt{2R/m}.
\]
If $R<m\,d_{\min}^2/8$, this implies $\|\btheta_i^\star-\btheta_j^\star\|_2<d_{\min}$, contradicting the definition of $d_{\min}$. Hence, at most one class can contain an $R$-close node, and the biased centroids are node-wise $R$-separable.\qed

\subsection{Proof of Theorem~\ref{thm:node_only_sufficient}}
\label{appendix:node_only_sufficient}

We prove that, under the condition
\[
K\rho \le \gamma < {m\,d_{\min}^{(0)}},
\]
the node-information-only problem~\eqref{eq:node_info_only} admits a class-wise constant global minimizer with distinct class representatives. Throughout, we use the class loss $F_i$, the biased centroids $\{\btheta_i^\star\}_{i=1}^K$ obtained from the characteristic problem~\eqref{eq:characteristic_opt_features}, and the centered fusion subgradient $\bomega_i$ as defined in Sec.~\ref{sec:node_only}. The proof proceeds in three steps: a first-order condition for the biased centroids (Step~1); the lower bound $K\rho\le\gamma$ guaranteeing within-class fusion (Step~2); and the upper bound on $\gamma$ ruling out class collapse (Step~3).

\paragraph{Step 1: first-order condition for the biased centroids.}
The first-order condition of the characteristic problem~\eqref{eq:characteristic_opt_features} at $\btheta_i^\star$, divided by $n_i$, is
\begin{equation}
\label{eq:appendix_cluster_foc}
\bzero
\in
\partial F_i(\btheta_i^\star)
+
\frac{\gamma}{n}\sum_{j\neq i} n_j\,\bg_{ij},
\qquad
\bg_{ij}\in\partial_{\btheta_i}\|\btheta_i-\btheta_j\|_2
\big|_{(\btheta_i^\star,\btheta_j^\star)},
\end{equation}
where $\bg_{ij}=(\btheta_i^\star-\btheta_j^\star)/\|\btheta_i^\star-\btheta_j^\star\|_2$ when $\btheta_i^\star\neq\btheta_j^\star$. Defining
\begin{equation}
\label{eq:appendix_omega_def}
\bomega_i:=\frac{\gamma}{n}\sum_{j\neq i} n_j\,\bg_{ij},
\end{equation}
\eqref{eq:appendix_cluster_foc} reads
\(
-\bomega_i \in \partial F_i(\btheta_i^\star).
\)

By the definition of $\rho_i$, we can choose node-level subgradients
\(
\bs_v\in\partial f_v(\btheta_i^\star), v\in\calC_i,
\)
such that
\begin{equation}
\label{eq:appendix_subgrad_average}
\frac{1}{n_i}\sum_{v\in\calC_i}\bs_v = -\bomega_i,
\qquad
\sup_{u,v\in\calC_i}\|\bs_v-\bs_u\|_2 \le \rho_i.
\end{equation}
(If the infimum in the definition of $\rho_i$ is not attained, we run the same argument with $\rho_i+\varepsilon$ and let $\varepsilon\downarrow 0$.) Setting $\bdelta_v:=\bs_v+\bomega_i$ gives the centered deviations
\begin{equation}
\label{eq:appendix_delta_sum_zero}
\sum_{v\in\calC_i}\bdelta_v = \bzero,
\qquad
\|\bdelta_v-\bdelta_u\|_2=\|\bs_v-\bs_u\|_2\le\rho_i.
\end{equation}

\paragraph{Step 2: lower bound on $\gamma$ gives within-class fusion.}
We construct admissible SON subgradients certifying that the class-wise constant assignment satisfies the node-level KKT conditions of~\eqref{eq:node_info_only}. For nodes in different classes, set $\bg_{vu}=\bg_{ij}$ for $v\in\calC_i, u\in\calC_j, i\neq j$; for nodes in the same class $\calC_i$, set
\begin{equation}
\label{eq:appendix_internal_g_choice}
\bg_{vu}:=-\frac{n}{\gamma\,n_i}(\bdelta_v-\bdelta_u),
\qquad u,v\in\calC_i,\ u\neq v,
\end{equation}
which is antisymmetric ($\bg_{uv}=-\bg_{vu}$). Using~\eqref{eq:appendix_delta_sum_zero}, the balanced-clusters identity $n_i=n/K$, and the lower bound $K\rho\le\gamma$,
\[
\|\bg_{vu}\|_2
=\frac{n}{\gamma\,n_i}\|\bdelta_v-\bdelta_u\|_2
\le\frac{n\,\rho_i}{\gamma\,n_i}
=\frac{K\rho_i}{\gamma}\le 1,
\]
so~\eqref{eq:appendix_internal_g_choice} is an admissible subgradient of $\|\btheta_v-\btheta_u\|_2$ at $\btheta_v^\star=\btheta_u^\star=\btheta_i^\star$.

For any $v\in\calC_i$, summing~\eqref{eq:appendix_internal_g_choice} over $u\in\calC_i\setminus\{v\}$ and using $\sum_{u\in\calC_i}\bdelta_u=\bzero$ from~\eqref{eq:appendix_delta_sum_zero},
\begin{equation}
\label{eq:appendix_internal_rowsum}
\sum_{u\in\calC_i\setminus\{v\}}\bg_{vu}
=-\frac{n}{\gamma\,n_i}\bigl((n_i-1)\bdelta_v+\bdelta_v\bigr)
=-\frac{n}{\gamma}\,\bdelta_v.
\end{equation}
Splitting the per-node first-order condition into within- and between-class contributions and substituting~\eqref{eq:appendix_internal_rowsum},
\begin{equation}
\label{eq:appendix_node_kkt}
\bs_v+\frac{\gamma}{n}\sum_{u\neq v}\bg_{vu}
=\bs_v
+\frac{\gamma}{n}\sum_{\substack{u\in\calC_i\\u\neq v}}\bg_{vu}
+\frac{\gamma}{n}\sum_{j\neq i}n_j\,\bg_{ij}
=\bs_v-\bdelta_v+\bomega_i
=\bzero,
\end{equation}
using $\bdelta_v=\bs_v+\bomega_i$. Thus $\bzero\in\partial f_v(\btheta_v^\star)+\frac{\gamma}{n}\sum_{u\neq v}\partial_{\btheta_v}\|\btheta_v^\star-\btheta_u^\star\|_2$ for every $v$, which is the first-order optimality condition of~\eqref{eq:node_info_only}; convexity then yields global optimality of the class-wise constant assignment.

\paragraph{Step 3: upper bound on $\gamma$ prevents class collapse.}
Assume $\gamma<m\,d_{\min}^{(0)}$ and suppose, toward a
contradiction, that a maximal group $S\subseteq\{1,\ldots,K\}$
of $q\ge2$ class representatives coincides at a point $\mathbf c$.
Thus $\btheta_i^\star=\mathbf c$ for $i\in S$ and
$\btheta_j^\star\neq\mathbf c$ for $j\notin S$.

For $i\in S$, set
\[
\mathbf a_i:=\btheta_i^{(0)}-\mathbf c,
\qquad
\overline{\mathbf a}:=\frac1q\sum_{i\in S}\mathbf a_i.
\]
Since $\bzero\in\partial F_i(\btheta_i^{(0)})$ and
$-\bomega_i\in\partial F_i(\mathbf c)$,
$m$-strong convexity of $F_i$ gives
\[
m\|\mathbf a_i\|_2^2
\le \langle\bomega_i,\mathbf a_i\rangle.
\]
Because the clusters are balanced, the fusion subgradients satisfy
\[
\bomega_i
=
\frac{\gamma}{K}
\sum_{\substack{j\in S\\j\neq i}}\bg_{ij}
+\mathbf w,
\qquad
\mathbf w:=
\frac{\gamma}{K}\sum_{j\notin S}
\frac{\mathbf c-\btheta_j^\star}
{\|\mathbf c-\btheta_j^\star\|_2},
\qquad
\|\mathbf w\|_2\le\frac{\gamma(K-q)}{K}.
\]
Here the external contribution $\mathbf w$ is common to all
$i\in S$, while $\bg_{ji}=-\bg_{ij}$ and
$\|\bg_{ij}\|_2\le1$ for the internal pairs.

Define the root-mean-square separation within $S$ by
\[
\Delta_S^2:=
\frac{2}{q(q-1)}
\sum_{\substack{i<j\\i,j\in S}}
\|\mathbf a_i-\mathbf a_j\|_2^2.
\]
Since $\mathbf a_i-\mathbf a_j
=\btheta_i^{(0)}-\btheta_j^{(0)}$, we have
$\Delta_S\ge d_{\min}^{(0)}$.
Summing the strong-convexity inequalities and using
antisymmetry gives
\begin{align*}
m\sum_{i\in S}\|\mathbf a_i\|_2^2
&\le
\frac{\gamma}{K}
\sum_{\substack{i<j\\i,j\in S}}
\langle\bg_{ij},\mathbf a_i-\mathbf a_j\rangle
+q\langle\mathbf w,\overline{\mathbf a}\rangle\\
&\le
\frac{\gamma q(q-1)}{2K}\Delta_S
+q\langle\mathbf w,\overline{\mathbf a}\rangle,
\end{align*}
where the last inequality follows from
$\|\bg_{ij}\|_2\le1$ and Cauchy--Schwarz over the
$q(q-1)/2$ pairs.

Using the variance identity
\[
\sum_{i\in S}\|\mathbf a_i\|_2^2
=
\frac{q-1}{2}\Delta_S^2
+q\|\overline{\mathbf a}\|_2^2
\]
and the bound
\[
\langle\mathbf w,\overline{\mathbf a}\rangle
-m\|\overline{\mathbf a}\|_2^2
=
\frac{\|\mathbf w\|_2^2}{4m}
-m\left\|\overline{\mathbf a}
-\frac{\mathbf w}{2m}\right\|_2^2
\le\frac{\|\mathbf w\|_2^2}{4m},
\]
we obtain
\[
\frac{m(q-1)}2\Delta_S^2
\le
\frac{\gamma q(q-1)}{2K}\Delta_S
+\frac{q\|\mathbf w\|_2^2}{4m}.
\]
Writing $G:=\gamma/m$ and substituting the bound on
$\|\mathbf w\|_2$ yields
\begin{equation}
\label{eq:appendix_fused_group_bound}
\Delta_S^2
\le
\frac qK G\Delta_S
+\frac{q}{2(q-1)}
\left(1-\frac qK\right)^2G^2
\le
\frac qK G\Delta_S
+\left(1-\frac qK\right)G^2.
\end{equation}
The second inequality uses $q\ge2$ and $q\le K$.
Rearranging gives
\[
(\Delta_S-G)
\left(\Delta_S+\left(1-\frac qK\right)G\right)
\le0,
\]
so $\Delta_S\le G$. Therefore any collapse would require
\[
d_{\min}^{(0)}
\le\Delta_S
\le\frac{\gamma}{m},
\]
contradicting $\gamma<m\,d_{\min}^{(0)}$.

Thus all class representatives are distinct. Together with
Step~2, this proves that the class-wise constant assignment
is a global minimizer with distinct class representatives.
\qed

\subsection{Proof of Corollary~\ref{cor:node_only_nonempty}}
\label{appendix:node_only_nonempty}

The interval~\eqref{eq:node_sufficient_cond} is non-empty iff its lower endpoint is strictly below the upper one:
\(
K\rho<m\,d_{\min}^{(0)}.
\)
Rearranging and using $\mathrm{SVR}=d_{\min}^{(0)}/\rho$ yields
\(
\mathrm{SVR}>K/m,
\)
which is~\eqref{eq:node_svr}. \qed

\subsection{Proof of Theorem~\ref{thm:node_only_tight}}
\label{appendix:node_only_tight}

We first show that the lower bound $K\rho\le\gamma$ in
Thm.~\ref{thm:node_only_sufficient} is tight in a worst-case sense:
if $K\rho_i>\gamma$ for some class~$\calC_i$, then there exists
a data configuration with within-class variability~$\rho_i$ on
which perfect recovery fails. This argument requires that the
loss family is rich enough to realize an arbitrary centered
subgradient pattern at $\btheta_i^\star$. We then construct
$m$-strongly convex losses with $\rho=0$ showing that the strict
upper bound $\gamma<m\,d_{\min}^{(0)}$ cannot be enlarged uniformly.

\paragraph{Subgradient-richness.}
The loss family $f(\bz;\btheta)$ is \emph{subgradient-rich at
$\btheta_i^\star$} if there is a unit vector $\bu\in\bbR^d$ such that,
for every centered collection
$\{\bdelta_v\}_{v\in\calC_i}\subset\operatorname{span}\{\bu\}$ with
$\sum_{v\in\calC_i}\bdelta_v=\bzero$, there exist data points
$\{\bz_v\}_{v\in\calC_i}$ and subgradients
$\bs_v\in\partial f_v(\btheta_i^\star)$ such that
$\bs_v=-\bomega_i+\bdelta_v$ and
$\langle\bs-\bs_v,\bu\rangle=0$ for all $\bs\in\partial f_v(\btheta_i^\star)$,
for every $v\in\calC_i$, where
$f_v(\btheta):=f(\bz_v;\btheta)$, and $\bomega_i$ is defined
in~\eqref{eq:appendix_omega_def}.

This condition is mild for standard losses: for instance, for
the squared loss
$f(\bx;\btheta)=\tfrac12\|\bx-\btheta\|_2^2$, the gradient is
$\nabla_{\btheta}f(\bx;\btheta_i^\star)=\btheta_i^\star-\bx$,
so every prescribed subgradient $\bs$ is realised by the
choice $\bx=\btheta_i^\star-\bs$.

\paragraph{Modulus attainment.}
An $m$-strongly convex loss family $f(\bz;\btheta)$ \emph{attains its modulus} if, for all
$\bar\btheta,\bc\in\Theta_f$, there exist data $\bz$ such that
$\bzero\in\partial f(\bz;\bar\btheta)$ and $m(\bc-\bar\btheta)\in\partial f(\bz;\bc)$.
For instance, $f(\bx;\btheta)=\tfrac m2\|\btheta-\bx\|_2^2$ attains its modulus with
$\bx=\bar\btheta$.

\paragraph{Adversarial construction.}
Fix a class $\calC_i$ with $n_i\ge2$ and assume
$K\rho_i>\gamma$, equivalently $\rho_i>\gamma n_i/n$
under balanced clusters. Pick a distinguished node
$v_0\in\calC_i$, let $\bu\in\bbR^d$ be the unit vector in the
definition of subgradient-richness, and set
$a:=\tfrac{n_i-1}{n_i}\rho_i$.
Define centered deviations
\[
\bdelta_{v_0}:=a\,\bu,
\qquad
\bdelta_v:=-\tfrac{a}{n_i-1}\bu
\quad\text{for }v\in\calC_i\setminus\{v_0\}.
\]
A direct computation gives
$\sum_{v\in\calC_i}\bdelta_v=\bzero$ and
$\sup_{u,v\in\calC_i}\|\bdelta_v-\bdelta_u\|_2=\rho_i$.
Since $\bdelta_v\in\operatorname{span}\{\bu\}$, subgradient-richness produces data
$\{\bz_v\}_{v\in\calC_i}$ with
$\bs_v=-\bomega_i+\bdelta_v\in\partial f_v(\btheta_i^\star)$
and $\langle\bs-\bs_v,\bu\rangle=0$ for all $\bs\in\partial f_v(\btheta_i^\star)$.
As $\frac1{n_i}\sum_{v\in\calC_i}\bs_v=-\bomega_i$ and
$\frac1{n_i}\sum_{v\in\calC_i}\partial f_v\subseteq\partial F_i$, the biased centroids
still satisfy the first-order condition of~\eqref{eq:characteristic_opt_features}.
Moreover, any admissible collection $\bs_v'\in\partial f_v(\btheta_i^\star)$ with
$\frac1{n_i}\sum_{v\in\calC_i}\bs_v'=-\bomega_i$ satisfies, for $w\neq v_0$,
\[
\|\bs_{v_0}'-\bs_w'\|_2
\ge\langle\bs_{v_0}'-\bs_w',\bu\rangle
=\langle\bdelta_{v_0}-\bdelta_w,\bu\rangle
=\rho_i,
\]
and $\{\bs_v\}$ attains this value. Hence the infimum defining the within-class
variability equals~$\rho_i$.

\paragraph{Failure of the class-wise constant solution.}
Suppose, toward a contradiction, that the class-wise constant assignment $\btheta_v^\star=\btheta_i^\star$ is optimal. Since the SON term is finite on all of $\bbR^{nd}$, the subdifferential of the objective of~\eqref{eq:node_info_only} splits into those of its terms, so optimality requires some $\bs_v'\in\partial f_v(\btheta_i^\star)$ satisfying the node-wise KKT condition~\eqref{eq:appendix_node_kkt}. Writing
$\bdelta_v':=\bs_v'+\bomega_i$, 
this condition at $v\in\calC_i$ forces, after using the fixed between-class subgradients
$\bg_{vu}=(\btheta_i^\star-\btheta_j^\star)/
\|\btheta_i^\star-\btheta_j^\star\|_2$
for $u\in\calC_j$, $j\neq i$,
\begin{equation}
\label{eq:appendix_node_only_tight_kkt}
\bzero
=\bdelta_v'+\frac{\gamma}{n}
\sum_{u\in\calC_i\setminus\{v\}}\bg_{vu},
\qquad
\|\bg_{vu}\|_2\le1,\quad \bg_{uv}=-\bg_{vu}.
\end{equation}
Apply~\eqref{eq:appendix_node_only_tight_kkt} at $v=v_0$ and take the inner product with $\bu$. By subgradient-richness, $\langle\bdelta_{v_0}',\bu\rangle=\langle\bdelta_{v_0},\bu\rangle=a$ for every such choice of $\bs_{v_0}'$, so we obtain
\[
a
=-\frac{\gamma}{n}
\sum_{u\in\calC_i\setminus\{v_0\}}
\langle\bg_{v_0u},\bu\rangle
\le
\frac{\gamma}{n}
\sum_{u\in\calC_i\setminus\{v_0\}}\|\bg_{v_0u}\|_2
\le\frac{\gamma}{n}(n_i-1).
\]
Substituting $a=\tfrac{n_i-1}{n_i}\rho_i$ and dividing by $n_i-1>0$ yields $\rho_i\le\gamma n_i/n$, i.e., $K\rho_i\le\gamma$, contradicting the assumption.
Hence the class-wise constant assignment at the biased centroids cannot be optimal under this data, so perfect recovery fails.

\paragraph{Tightness of the upper bound.}
Let $\btheta_1^{(0)},\ldots,\btheta_K^{(0)}$ be the vertices of a regular simplex of side length $d>0$ in the common convex parameter domain, and set
$
\bc:=\frac1K\sum_{i=1}^K\btheta_i^{(0)}.
$
By modulus attainment, choose data $\bz_i$ such that
\[
\bzero\in\partial f(\bz_i;\btheta_i^{(0)}),
\qquad
\bs_i:=m(\bc-\btheta_i^{(0)})
\in\partial f(\bz_i;\bc).
\]
Set $\bz_v:=\bz_i$ for every $v\in\calC_i$. Strong convexity makes $\btheta_i^{(0)}$ the unique
minimizer of $F_i=f(\bz_i;\cdot)$, so
$d_{\min}^{(0)}=d$.
For any $\gamma\ge md$, consider the collapsed assignment
$\btheta_v=\bc$ for all $v$, and define
\[
\bg_{vu}:=
\begin{cases}
\bzero, & v,u\in\calC_i,\\[3pt]
\displaystyle
\frac{m}{\gamma}
(\btheta_i^{(0)}-\btheta_j^{(0)}),
& v\in\calC_i,\ u\in\calC_j,\ i\ne j.
\end{cases}
\]
These vectors are antisymmetric and have norm at most one, so they are admissible fusion subgradients at the collapsed assignment. For $v\in\calC_i$, balance gives
\[
\bs_i+\frac{\gamma}{n}\sum_{u\ne v}\bg_{vu}
=
m(\bc-\btheta_i^{(0)})
+\frac mK\sum_{j\ne i}
(\btheta_i^{(0)}-\btheta_j^{(0)})
=\bzero.
\]
Thus the node-level KKT conditions hold. By convexity the collapsed assignment is a global minimizer, and strong convexity makes it unique.

The same fusion subgradients give
$\bomega_i=m(\btheta_i^{(0)}-\bc)$.
Consequently, the identical within-class selections $\bs_v=\bs_i=-\bomega_i$ are admissible in the definition of $\rho_i$ and have zero spread. Hence $\rho=0$, and all classes collapse whenever
$\gamma\ge m\,d_{\min}^{(0)}$.

\newpage
\section{Proofs for the Combined Regime}
\label{sec:appendix:perfect}

This appendix establishes the recovery guarantees for the unified problem~\eqref{eq:main_model} that couples the graph and the node-information regimes through a shared atomic representation. 

\subsection{Proof of Theorem~\ref{thm:combined_recovery}}
\label{appendix:combined_recovery}

We follow a dual-certificate approach: we verify that the perfect-recovery solution defined in~\eqref{eq:gen_ideal_atoms} satisfies the KKT conditions of the unified problem~\eqref{eq:main_model} by exhibiting a symmetric dual matrix $\bZ\in\bbR^{n\times n}$ together with SON subgradients $\{\bg_{vu}\}$ that certify global optimality. The proof has six parts: (i) a polarized reformulation of~\eqref{eq:main_model} as a lifted atomic program, together with the cluster- and node-level subgradient calculus; (ii) the atomwise feature/SON gap $h_v$; (iii) the data parameters controlling the recovery condition; (iv) the optimality conditions, together with a global-optimality argument explaining why a certificate satisfying them is sufficient; (v) the construction and sign verification of the certificate $\bZ$; (vi) the PSD reduction and a case-by-case analysis. Throughout we adopt the notation of Secs.~\ref{sec:node_classification_problem}--\ref{sec:combined}: the partition $\calV=\bigcup_{i=1}^K\calC_i$ with $|\calC_i|=n_i$, 
node losses $f_v(\btheta)=f(\bz_v;\btheta)$ convex on a closed convex domain $\Theta_f$, the class loss $F_i(\btheta)=\frac1{n_i}\sum_{v\in\calC_i}f_v(\btheta)$, and the biased centroids $\{\btheta_i^\star\}$ from~\eqref{eq:characteristic_opt_features}.

\subsubsection{Polarized reformulation and Bregman tools}
\label{appendix:combined_setup}

Let
\(
\calA:=\{(\be,\btheta):\|\be\|_2=1,\ \btheta\in\Theta_f\}
\)
denote the atomic set, and for $a=(\be_a,\btheta_a)\in\calA$ write $e_{a,v}$ for the $v$-th coordinate of $\be_a$; 
we index atoms by $a$ rather than by $i$ as in~\eqref{eq:atomic_L}--\eqref{eq:main_model}, to avoid ambiguity with the class indices.
Working with the polarized adjacency $\bar\bA$ ($\bar A_{uv}=+1$ for $\{u,v\}\in\calE$, $\bar A_{uv}=-1$ otherwise), problem~\eqref{eq:main_model} is equivalent, up to additive constants, to the lifted atomic program
\begin{align}
\min_{\substack{\bL,\,\{\btheta_v\}_{v\in\calV},\\ \{\lambda_a\ge 0\}_{a\in\calA}}}\quad
&-\langle\bar\bA,\bL\rangle
+n\mu\sum_{a\in\calA}\lambda_a\sum_{v\in\calV}e_{a,v}^2 f_v(\btheta_a)
+\gamma\mu\sum_{u<v}\|\btheta_u-\btheta_v\|_2,
\label{eq:unified_atomic_problem}\\
\text{s.t.}\quad
&\bL=\sum_{a\in\calA}\lambda_a\be_a\be_a^\top,
\qquad
\btheta_v=\sum_{a\in\calA}\lambda_a e_{a,v}^2\btheta_a,\ \forall v\in\calV,
\nonumber\\
& L_{vv}=1,\ \forall v\in\calV,
\qquad 0\le L_{uv}\le 1,\ \forall u,v\in\calV.
\nonumber
\end{align}

The atom set $\calA$ is fixed; the optimization variables are the nonnegative weights $\{\lambda_a\}$, the induced matrix $\bL$, and the induced node parameters $\{\btheta_v\}$. The maps $\lambda\mapsto\sum_a\lambda_a\be_a\be_a^\top$ and $\lambda\mapsto\sum_a\lambda_a e_{a,v}^2\btheta_a$ are linear, and the lifted feature term is linear in $\lambda$, so the only nonlinear term in the objective is the SON penalty, which is convex in $\{\btheta_v\}$. Thus~\eqref{eq:unified_atomic_problem} is convex.

The perfect-recovery atoms $(\lambda_i^\star,\be_i^\star,\btheta_i^\star)$ from~\eqref{eq:gen_ideal_atoms} induce the partition matrix $\bL^\star$ in~\eqref{eq:optimal_L}, and the lifted feature loss collapses to
\[
\sum_a\lambda_a^\star\sum_{v\in\calV}(e_{a,v}^\star)^2 f_v(\btheta_a^\star)
=\sum_{i=1}^K\sum_{v\in\calC_i}f_v(\btheta_i^\star)
=\sum_{i=1}^K n_i F_i(\btheta_i^\star),
\]
so the class representatives $\{\btheta_i^\star\}_{i=1}^K$ solve the characteristic problem
\begin{equation}\label{eq:gen_characteristic_problem}
\{\btheta_i^\star\}_{i=1}^K
\in\arg\min_{\{\btheta_i\}}
\sum_{i=1}^K n_i F_i(\btheta_i)
+\frac{\gamma}{n}\sum_{i<j}n_in_j\|\btheta_i-\btheta_j\|_2,
\end{equation}
which is exactly~\eqref{eq:characteristic_opt_features}.

\paragraph{Bregman divergence.}
For a proper closed convex $\phi:\bbR^d\to(-\infty,+\infty]$ and any $\bs\in\bbR^d$, the generalized Bregman divergence is
\begin{equation}\label{eq:general_bregman_def}
D_\phi^{\bs}(\btheta,\btheta^\star)
:=\phi(\btheta)-\phi(\btheta^\star)-\langle\bs,\btheta-\btheta^\star\rangle,
\end{equation}
and $\bs\in\partial\phi(\btheta^\star)$ if and only if $D_\phi^{\bs}(\btheta,\btheta^\star)\ge 0$ for every $\btheta\in\operatorname{dom}(\phi)$.

\paragraph{Cluster-level subgradient.}
Dividing the first-order condition of~\eqref{eq:gen_characteristic_problem} by $n_i$,
\begin{equation}\label{eq:gen_cluster_foc}
\bzero
\in
\partial F_i(\btheta_i^\star)
+\frac{\gamma}{n}\sum_{j\neq i}n_j\,
\frac{\btheta_i^\star-\btheta_j^\star}{\|\btheta_i^\star-\btheta_j^\star\|_2}.
\end{equation}
With
\begin{equation}\label{eq:gen_omega_def}
\bomega_i:=\frac{\gamma}{n}\sum_{j\neq i}n_j\,\frac{\btheta_i^\star-\btheta_j^\star}{\|\btheta_i^\star-\btheta_j^\star\|_2},
\end{equation}
this reads
\(-\bomega_i\in\partial F_i(\btheta_i^\star),
\)
which by~\eqref{eq:general_bregman_def} gives
\(D_{F_i}^{-\bomega_i}(\btheta,\btheta_i^\star)\ge 0,
 \forall\btheta\in\operatorname{dom}(F_i).
\)
\paragraph{Node-level SON subgradients.}
Evaluating the SON subgradient at the cluster-constant assignment $\btheta_v^\star=\btheta_{y_v}^\star$, the antisymmetric subgradient is
\begin{equation}\label{eq:gen_subgrad_SON}
\bg_{vu}
=\begin{cases}
\text{any vector with }\|\bg_{vu}\|_2\le 1, & y_v=y_u,\\[4pt]
\dfrac{\btheta_{y_v}^\star-\btheta_{y_u}^\star}{\|\btheta_{y_v}^\star-\btheta_{y_u}^\star\|_2}, & y_v\neq y_u,
\end{cases}
\qquad \bg_{uv}=-\bg_{vu}.
\end{equation}

By the standard subdifferential sum rule, there exist node-level subgradients $\bs_v\in\partial f_v(\btheta_i^\star)$ with
\begin{equation}\label{eq:node_subgrad_average}
-\bomega_i=\frac{1}{n_i}\sum_{v\in\calC_i}\bs_v;
\end{equation}
when the $f_v$ are differentiable, $\bs_v=\nabla f_v(\btheta_i^\star)$ and $-\bomega_i=\nabla F_i(\btheta_i^\star)$.

\subsubsection{Atomwise loss gap}
\label{appendix:combined_hv}

Define the node-level residual subgradient
\begin{equation}\label{eq:node_residual_tau}
\btau_v:=\bs_v+\frac{\gamma}{n}\sum_{u\neq v}\bg_{vu},\qquad v\in\calC_i.
\end{equation}
Averaging~\eqref{eq:node_residual_tau} over $v\in\calC_i$, the within-class $\bg_{vu}$ terms cancel by antisymmetry and the between-class terms reproduce $\bomega_i$, so~\eqref{eq:node_subgrad_average} gives
\begin{equation}\label{eq:tau_average_zero}
\frac{1}{n_i}\sum_{v\in\calC_i}\btau_v=\bzero.
\end{equation}
For $v\in\calC_i$, the node-wise Bregman divergence
% \begin{equation}\label{eq:node_bregman_def}
\(
D_{f_v}^{\bs_v}(\btheta,\btheta_i^\star)
:=f_v(\btheta)-f_v(\btheta_i^\star)-\langle\bs_v,\btheta-\btheta_i^\star\rangle
\)
% \end{equation}
satisfies
\(
D_{f_v}^{\bs_v}(\btheta,\btheta_i^\star)\ge 0,
 \forall \btheta\in\operatorname{dom}(f_v),
\)
on the common effective domain
\(\Theta_f:=\bigcap_{v\in\calV}\operatorname{dom}(f_v).
\)

Define the \emph{atomwise feature/SON gap}
\begin{equation}\label{eq:def_hv_node_bregman}
h_v(\btheta)
:=\langle\btau_v,\btheta-\btheta_i^\star\rangle
+D_{f_v}^{\bs_v}(\btheta,\btheta_i^\star)
=f_v(\btheta)-f_v(\btheta_i^\star)+\Big\langle\frac{\gamma}{n}\sum_{u\neq v}\bg_{vu},\,\btheta-\btheta_i^\star\Big\rangle,
\end{equation}
which vanishes at the centroid:
\(h_v(\btheta_i^\star)=0, \forall v\in\calC_i.
\)
Averaging $h_v$ over $v\in\calC_i$ and using~\eqref{eq:tau_average_zero},
\begin{equation}\label{eq:cluster_node_bregman_identity}
D_{F_i}^{-\bomega_i}(\btheta,\btheta_i^\star)
=\frac{1}{n_i}\sum_{v\in\calC_i}D_{f_v}^{\bs_v}(\btheta,\btheta_i^\star),
\qquad
\frac{1}{n_i}\sum_{v\in\calC_i}h_v(\btheta)
=D_{F_i}^{-\bomega_i}(\btheta,\btheta_i^\star).
\end{equation}

\subsubsection{Data parameters}
\label{appendix:combined_data_params}

\paragraph{Within-class variability.}
The variability $\rho_i$ and $\rho:=\max_i\rho_i$ are exactly as defined in Sec.~\ref{sec:node_only}: $\rho_i$ is the smallest spread $\sup_{u,v\in\calC_i}\|\bs_v-\bs_u\|_2$ achievable by node-level subgradients satisfying~\eqref{eq:node_subgrad_average}, and reduces to $\sup_{u,v\in\calC_i}\|\nabla f_v(\btheta_i^\star)-\nabla f_u(\btheta_i^\star)\|_2$ when each $f_v$ is differentiable. We assume the infimum is attained; otherwise replace $\rho_i$ with $\rho_i+\varepsilon$ and let $\varepsilon\downarrow 0$.

\paragraph{Node-wise $R$-separability.}
For each $v\in\calC_i$, let $\bs_v\in\partial f_v(\btheta_i^\star)$ be the node-level subgradient chosen in~\eqref{eq:node_subgrad_average}. The centroids $\{\btheta_i^\star\}_{i=1}^K$ are \emph{node-wise $R$-separable} if, for every $\btheta\in\Theta_f$, there is at most one class $i_0\in[K]$ such that
\begin{equation}\label{eq:nodewise_separability_bad_cluster}
\min_{v\in\calC_{i_0}}D_{f_v}^{\bs_v}(\btheta,\btheta_{i_0}^\star)\le R;
\end{equation}
equivalently, for every $j\neq i_0$ and every $v\in\calC_j$,
\begin{equation}\label{eq:nodewise_separability_app}
D_{f_v}^{\bs_v}(\btheta,\btheta_j^\star)>R.
\end{equation}

At most one ``bad'' class can contain a node whose feature Bregman gap to its centroid is small; nodes in every other class have gap exceeding $R$.

\paragraph{Graph parameters.}
We use the graph parameters from Sec.~\ref{sec:graph_only}: the misconnection count $\bar n_{v,j}$, its aggregate $\bar N_{i,j}=\sum_{v\in\calC_i}\bar n_{v,j}$, the block rate $\bar\rho_{i,j}=\bar N_{i,j}/(n_i n_j)$, 
% the deviation bound $|\bar n_{v,j}/n_j-\bar\rho_{i,j}|\le\delta_{ij}$, 
the maximal deviation $\delta_{ij}=\max_{v\in\calC_i}|\bar n_{v,j}/n_j-\bar\rho_{i,j}|$,
the misconnection rate $p_{ij}=\bar\rho_{i,j}+2\delta_{ij}$, and $p_{\max}=\max_{i,j}p_{ij}$ as in~\eqref{eq:p_def}.

\subsubsection{Optimality conditions and global-optimality argument}
\label{appendix:combined_optimality}

To certify optimality of the perfect-recovery solution, we construct a symmetric dual matrix $\bZ\in\bbR^{n\times n}$ and SON subgradients $\{\bg_{vu}\}$ satisfying the conditions below. Write $\bH(\btheta):=\operatorname{diag}(h_v(\btheta))_{v\in\calV}$. The required conditions are
\begin{align}
(\bZ-\bar\bA)_{uv} &\le 0, &&\text{if } u,v\in\calC_i,\ u\neq v,
\label{eq:optimality_L1}\\[1pt]
(\bZ-\bar\bA)_{uv} &\ge 0, &&\text{if } u\in\calC_i,\ v\in\calC_j,\ i\neq j,
\label{eq:optimality_L2}\\[1pt]
\be_i^{\star\top}\bigl(n\mu\bH(\btheta_i^\star)-\bZ\bigr)\be_i^\star &= 0, &&\forall i,
\label{eq:optimality_eq_simplified}\\[1pt]
\be^\top\bigl(n\mu\bH(\btheta)-\bZ\bigr)\be &\ge 0, &&\forall(\be,\btheta)\in\calA.
\label{eq:optimality_ineq_simplified}
\end{align}

Conditions~\eqref{eq:optimality_L1}--\eqref{eq:optimality_L2} are the entrywise complementary-slackness conditions for $\bL$ on off-diagonal entries (the diagonals are unconstrained because $L_{vv}=1$ is fixed); equality~\eqref{eq:optimality_eq_simplified} must hold at the $K$ ideal atoms $\{(\be_i^\star,\btheta_i^\star)\}$, while~\eqref{eq:optimality_ineq_simplified} must hold for every atom in $\calA$.

\paragraph{Why the certificate is sufficient.}
Let $(\bL,\{\btheta_v\},\{\lambda_a\})$ be any feasible point of~\eqref{eq:unified_atomic_problem}. Convexity of the SON penalty together with the subgradients $\{\bg_{vu}\}$ from~\eqref{eq:gen_subgrad_SON} gives
\[
\sum_{u<v}\|\btheta_u-\btheta_v\|_2-\sum_{u<v}\|\btheta_u^\star-\btheta_v^\star\|_2
\ge
\sum_{v\in\calV}\Big\langle\sum_{u\neq v}\bg_{vu},\,\btheta_v-\btheta_v^\star\Big\rangle.
\]

Using the lifted representation $\btheta_v=\sum_a\lambda_a e_{a,v}^2\btheta_a$, the cluster-constant ideal $\btheta_v^\star=\btheta_{y_v}^\star$, and the diagonal constraint $L_{vv}=\sum_a\lambda_a e_{a,v}^2=1$, the lifted feature plus SON gap can be written as
\[
n\mu\sum_{a\in\calA}\lambda_a\sum_{v\in\calV}e_{a,v}^2\,h_v(\btheta_a).
\]

Combining this with the $-\langle\bar\bA,\bL\rangle$ term, the full objective gap admits the lower bound
\[
\sum_{a\in\calA}\lambda_a\,\be_a^\top\bigl(n\mu\bH(\btheta_a)-\bZ\bigr)\be_a
+\langle\bZ-\bar\bA,\,\bL-\bL^\star\rangle.
\]
The first term is nonnegative by~\eqref{eq:optimality_ineq_simplified}; the second is nonnegative by the box constraints $0\le L_{uv}\le 1=L_{uv}^\star$ on intra-cluster pairs and $0=L_{uv}^\star\le L_{uv}$ on inter-cluster pairs, combined with the sign conditions~\eqref{eq:optimality_L1}--\eqref{eq:optimality_L2}. Hence the perfect-recovery solution is globally optimal, so it suffices to construct $\bZ$ and $\{\bg_{vu}\}$ satisfying~\eqref{eq:optimality_L1}--\eqref{eq:optimality_ineq_simplified}.

\subsubsection{Constructing and verifying the dual certificate}
\label{appendix:combined_dual_cert}

\paragraph{Step 1: structured block matrix.}
Let $\bS\in\bbR^{n\times n}$ be defined block-wise by
\[
S_{uv}:=
\begin{cases}
0, & u,v\in\calC_i \text{ and } \{u,v\}\in\calE,\\
-a_{ii}, & u,v\in\calC_i \text{ and } \{u,v\}\notin\calE,\\
0, & u\in\calC_i,\ v\in\calC_j,\ i\neq j,\ \{u,v\}\notin\calE,\\
a_{ij}, & u\in\calC_i,\ v\in\calC_j,\ i\neq j,\ \{u,v\}\in\calE,
\end{cases}
\]
with scalars $a_{ii},a_{ij}>0$ to be determined. By design, $\bS$ assigns weight only to misconnected pairs, penalizing missing intra-cluster edges and spurious inter-cluster edges.

\paragraph{Step 2: orthogonal projection.}
Set $\bE:=[\be_1^\star,\ldots,\be_K^\star]$ and $\bP^\perp:=\bI-\bE\bE^\top$. Define
\begin{equation}\label{eq:gen_Zs}
\bZ:=\bP^\perp(\bS-\lambda\bI)\bP^\perp,
\end{equation}
so that $\be_i^{\star\top}\bZ\be_i^\star=0, \forall i$. Combined with $h_v(\btheta_i^\star)\!=\!0$, this gives $\be_i^{\star\top}(n\mu\bH(\btheta_i^\star)-\bZ)\be_i^\star=0$, i.e.\ \eqref{eq:optimality_eq_simplified}.

\paragraph{Step 3: sign-condition verification.}
Enforcing~\eqref{eq:optimality_L1}--\eqref{eq:optimality_L2} entrywise on $\bZ-\bar\bA$ and using the 
% deviation bound 
definition of $\delta_{ij}$ from Section~\ref{sec:graph_only}
% defined in Section~\ref{sec:graph_only}, 
reduces the sign conditions to the entry-wise intervals
\begin{align}
\frac{1+\lambda/n_i}{1-p_{ii}}&\le a_{ii}\le\frac{1-\lambda/n_i}{p_{ii}},\\
\frac{1}{1-p_{ij}}&\le a_{ij}\le\frac{1}{p_{ij}},\qquad i\neq j;
\end{align}
this is the same calculation as in App.~\ref{appendix:graph_sufficient}. The intra-cluster interval is non-empty iff
\begin{equation}\label{eq:gen_lambda_cond}
\lambda\le(1-2p_{\max})\,n_i.
\end{equation}
The inter-cluster interval requires $p_{ij} \le 1/2$. The smallest feasible (worst-case-uniform) choices are
\begin{equation}\label{eq:optimized_a_values_uniform}
a_{ii}(\lambda)=\frac{1+\lambda/n_i}{1-p_{\max}},
\qquad
a_{ij}=\frac{1}{1-p_{\max}},\ i\neq j,
\end{equation}
which we adopt henceforth.

\subsubsection{Positive-semi-definiteness of \texorpdfstring{$n\mu\bH(\btheta)-\bZ$}{nμH(θ)−Z}}
\label{appendix:combined_psd}

It remains to verify~\eqref{eq:optimality_ineq_simplified} for every $(\be,\btheta)\in\calA$. We proceed in five steps: bound the residuals $\btau_v$, convert the bound to a Bregman residual price, localize via $R$-separability, reduce to a block PSD inequality, and solve it.

\paragraph{Step 1: controlling the node-level residuals.}
Fix a cluster $\calC_i$. By the definition of $\rho_i$, choose admissible $\{\bs_v\}_{v\in\calC_i}$ satisfying~\eqref{eq:node_subgrad_average} and $\|\bs_v-\bs_u\|_2\le\rho_i$. Setting $\bdelta_v:=\bs_v+\bomega_i$ gives $\sum_{v\in\calC_i}\bdelta_v=\bzero$, $\|\bdelta_v-\bdelta_u\|_2\le\rho_i$, and $\|\bdelta_v\|_2\le\rho_i$. We choose the within-class SON subgradients to absorb these residuals while preserving $\|\bg_{vu}\|_2\le 1$. With
$
\eta_i:=
\min\{1/\rho_i,\,n/(\gamma n_i)\},  \rho_i>0,
$
and $0$ for $\rho_i=0$,
where $n/(\gamma n_i)=+\infty$ when $\gamma=0$,
\begin{equation}\label{eq:Zvu_def_2}
\bg_{vu}=\begin{cases}
-\eta_i(\bdelta_v-\bdelta_u), & u,v\in\calC_i,\\[4pt]
\dfrac{\btheta_i^\star-\btheta_j^\star}{\|\btheta_i^\star-\btheta_j^\star\|_2}, & v\in\calC_i,\ u\in\calC_j,\ j\neq i,
\end{cases}
\end{equation}
which is antisymmetric and admissible (in the within-class case, $\|\bg_{vu}\|_2=\eta_i\|\bdelta_v-\bdelta_u\|_2\le\eta_i\rho_i\le 1$), as required by~\eqref{eq:gen_subgrad_SON}. Substituting~\eqref{eq:Zvu_def_2} into~\eqref{eq:node_residual_tau} and using $\sum_{u\in\calC_i}\bdelta_u=\bzero$,
\(
\btau_v=(1-\gamma n_i\eta_i/n)\,\bdelta_v,
\)
hence $\|\btau_v\|_2\le(\rho_i-\gamma n_i/n)_+$. Defining
\begin{equation}\label{eq:Delta_i_def_bregman}
r_i:=(\rho_i-\gamma n_i/n)_+,
\qquad r_{\max}:=\max_i r_i,
\end{equation}
we obtain
\(
\|\btau_v\|_2\le r_i,
\forall v\in\calC_i.
\)
Since $\btau_v=(1-\gamma n_i\eta_i/n)\,\bdelta_v$ with $1-\gamma n_i\eta_i/n\in[0,1]$, the residuals inherit the centering and the spread of the $\bdelta_v$:
\begin{equation}\label{eq:tau_pattern}
\sum_{v\in\calC_i}\btau_v=\bzero,
\qquad
\max_{u,v\in\calC_i}\|\btau_v-\btau_u\|_2\le r_i.
\end{equation}

\paragraph{Step 2: centered Bregman residual price.}
For each class $\calC_i$ and $\btheta\in\Theta_f$, define
\begin{equation}\label{eq:centered_price_ingredients}
\underline D_i(\btheta)
:=
\min_{v\in\calC_i}
D_{f_v}^{\bs_v}(\btheta,\btheta_i^\star),
\qquad
\ell_i(\btheta)
:=
\max_{u,v\in\calC_i}
\langle\btau_u-\btau_v,\btheta-\btheta_i^\star\rangle.
\end{equation}
Here $\underline D_i(\btheta)\ge0$ is the smallest node-wise
Bregman gap in the class, and $\ell_i(\btheta)$ is the range of
the centered quantities
$\langle\btau_v,\btheta-\btheta_i^\star\rangle$.
By~\eqref{eq:tau_pattern},
\[
0\le\ell_i(\btheta)
\le r_i\|\btheta-\btheta_i^\star\|_2.
\]
Define the class-level and global residual-price coefficients by
\begin{equation}\label{eq:zeta_centered_def}
\zeta_i
:=
\sup_{\btheta\in\Theta_f}
\frac{\|\btheta-\btheta_i^\star\|_2^2}
{4\,\underline D_i(\btheta)},
\qquad
\zeta:=\max_i\zeta_i,
\end{equation}
with the conventions $0/0\!:=\!0$ and $c/0\!:=\!+\infty$ for $c\!>\!0$.
Thus $\zeta_i\!\in\![0,+\infty]$, and whenever $\zeta_i\!<\!+\infty$,
\begin{equation}\label{eq:centered_price_bound}
\ell_i(\btheta)^2
\le 4\zeta_i r_i^2\,\underline D_i(\btheta),
\qquad \forall\btheta\in\Theta_f.
\end{equation}

In particular, $\ell_i(\btheta)^2\le 4\zeta r_{\max}^2\underline D_i(\btheta)$ when $\zeta<+\infty$. If $r_i=0$, then $\ell_i(\btheta) = 0$, independently of $\zeta_i$.

\paragraph{Step 3: localizing the Bregman gap via $R$-separability.}
Fix $\btheta\in\Theta_f$ and assume $\zeta<+\infty$. Write $\bH_j(\btheta)$ for the block of $\bH(\btheta)$ corresponding to $\calC_j$. For any $\bx\in\bbR^{n_j}$,
\eqref{eq:def_hv_node_bregman} gives
\begin{align*}
\bx^\top\bH_j(\btheta)\bx
=\sum_{v\in\calC_j}h_v(\btheta)x_v^2
=
\sum_{v\in\calC_j}
\langle\btau_v,\btheta-\btheta_j^\star\rangle x_v^2
+
\sum_{v\in\calC_j}
D_{f_v}^{\bs_v}(\btheta,\btheta_j^\star)x_v^2.
\end{align*}
We bound these two contributions separately, using part of the nonnegative Bregman term to control the possibly negative residual term.

First consider the Bregman term. By node-wise $R$-separability \eqref{eq:nodewise_separability_bad_cluster}, at most one class $i_0$ can satisfy $\underline D_{i_0}(\btheta)\!\le\! R$; pick $i_0$ arbitrarily if no such class exists.
Setting $R_{i_0}\!:=\!0$ and $R_j\!:=\!R$ for $j\!\neq\! i_0$, we obtain
\[
D_{f_v}^{\bs_v}(\btheta,\btheta_j^\star)\ge R_j,
\qquad v\in\calC_j.
\]
and, since each gap is also at least $\underline D_j(\btheta)$ by~\eqref{eq:centered_price_ingredients}, 
\[
D_{f_v}^{\bs_v}(\btheta,\btheta_j^\star)
\ge
(1-t_j)R_j+t_j\underline D_j(\btheta),
\qquad\forall \ t_j\in(0,1].
\]

For the residual term, set
$
a_v:=\langle\btau_v,\btheta-\btheta_j^\star\rangle, v\in\calC_j.
$
By~\eqref{eq:tau_pattern} and \eqref{eq:centered_price_ingredients},
\[
\sum_{v\in\calC_j}a_v=0,
\qquad
\max_{v\in\calC_j}a_v-\min_{v\in\calC_j}a_v
=\ell_j(\btheta).
\]
Lemma~\ref{lem:centered_diagonal_bound} therefore gives
$
\left|\sum_{v\in\calC_j}a_vx_v^2\right|
\le
\ell_j(\btheta)\|\bx\|_2\|\bP_j^\perp\bx\|_2,
$
where 
$
\bP_j^\perp:=\bI_{n_j}-\frac{1}{n_j}\bone\bone^\top.
$

Combining the bounds on the two contributions yields
\begin{align*}
\bx^\top\bH_j(\btheta)\bx
\ge
(1-t_j)R_j\|\bx\|_2^2
+t_j\underline D_j(\btheta)\|\bx\|_2^2
-\ell_j(\btheta)\|\bx\|_2\|\bP_j^\perp\bx\|_2.
\end{align*}
Since~\eqref{eq:centered_price_bound} gives
$\ell_j(\btheta)\le
2r_{\max}\sqrt{\zeta\,\underline D_j(\btheta)}$,
the last two terms satisfy
\begin{align*}
&t_j\underline D_j(\btheta)\|\bx\|_2^2
-\ell_j(\btheta)\|\bx\|_2\|\bP_j^\perp\bx\|_2\\
&\quad\ge
\left(
\sqrt{t_j\underline D_j(\btheta)}\,\|\bx\|_2
-r_{\max}\sqrt{\frac{\zeta}{t_j}}\,
 \|\bP_j^\perp\bx\|_2
\right)^2
-\frac{\zeta r_{\max}^2}{t_j}
 \|\bP_j^\perp\bx\|_2^2\\
&\quad\ge
-\frac{\zeta r_{\max}^2}{t_j}
 \|\bP_j^\perp\bx\|_2^2.
\end{align*}

Consequently,
\[
\bH_j(\btheta)
\succeq
(1-t_j)R_j\bI_{n_j}
-\frac{\zeta r_{\max}^2}{t_j}\bP_j^\perp.
\]
% The argument also covers $\underline D_j(\btheta)=0$:
% finiteness of $\zeta$ then forces $\ell_j(\btheta)=0$.

Choose $R_{i_0}=0$ and $t_{i_0}=1$ on the exceptional class, and $R_j=R$ and $t_j=t\in(0,1]$ on the other classes. Then
\begin{equation}
\bH_{i_0}(\btheta)
\succeq-\zeta r_{\max}^2\bP_{i_0}^\perp,
\qquad
\bH_j(\btheta)\succeq
(1-t)R\bI_{n_j}
-\frac{\zeta r_{\max}^2}{t}\bP_j^\perp
\quad(j\neq i_0).
\end{equation}

Equivalently,
\begin{equation}\label{eq:H_lower_nodewise}
\bH(\btheta)\succeq\bQ-\boldsymbol\chi\bP^\perp,
\qquad
\bQ:=\operatorname{diag}(q_v),\quad
\boldsymbol\chi:=\operatorname{diag}(\xi_v),
\end{equation}
where
\[
(q_v,\xi_v):=
\begin{cases}
(0,\zeta r_{\max}^2), & v\in\calC_{i_0},\\[1.5pt]
((1-t)R,\zeta r_{\max}^2/t), & v\notin\calC_{i_0}.
\end{cases}
\]

\paragraph{Step 4: PSD reduction.}
The matrices $\bQ$ and $\boldsymbol\chi$ are constant on each
class and therefore commute with $\bP^\perp$ and
$\bE\bE^\top=\bI-\bP^\perp$.
Combining~\eqref{eq:gen_Zs} and~\eqref{eq:H_lower_nodewise},
\begin{align*}
n\mu\bH(\btheta)-\bZ
&\succeq
n\mu\bQ-n\mu\boldsymbol\chi\bP^\perp
+\bP^\perp(\lambda\bI-\bS)\bP^\perp
\\
&=
n\mu\bQ\bE\bE^\top
+\bP^\perp
\bigl(\lambda\bI-\bS+n\mu\bQ-n\mu\boldsymbol\chi\bigr)
\bP^\perp.
\end{align*}
Since $\bQ$ is nonnegative and constant on each class,
$\bQ\bE\bE^\top\succeq0$. Hence
\eqref{eq:optimality_ineq_simplified} holds whenever
\begin{equation}\label{eq:Gamma_F_def}
\bGamma:=
\lambda\bI-\bS+n\mu\bQ-n\mu\boldsymbol\chi
\succeq0.
\end{equation}

\paragraph{Step 5: block PSD bound with the optimized graph weights.}
Partition $\bGamma$ according to $\calC_{i_0}$ and
$\calC_{\neq i_0}:=\calV\setminus\calC_{i_0}$.
For balanced clusters, write $n_i=n/K$.
Using the optimized choices~\eqref{eq:optimized_a_values_uniform}, write $a_{\rm in}(\lambda):=(1+\lambda/n_i)/(1-p)$ and $a_{\rm out}:=1/(1-p)$ with $p:=p_{\max}$. The row-sum bounds on the diagonal blocks of $\bS$ and the row- and column-sum bounds on its off-diagonal block give
\begin{align*}
\bGamma_{11}
&\succeq
\bigl(\lambda-a_{\rm in}(\lambda)\,p\,n_i-n\mu\zeta r_{\max}^2\bigr)
\bI_{n_i},\\[0pt]
\bGamma_{22}
&\succeq
\left(
\lambda-a_{\rm in}(\lambda)\,p\,n_i
-a_{\rm out}\,p(n-2n_i)
+n\mu\left((1-t)R-\frac{\zeta}{t}r_{\max}^2\right)
\right)\bI_{n-n_i},\\[0pt]
\sigma_{\max}^2(\bGamma_{12})
&\le a_{\rm out}^2\,p^2\,n_i(n-n_i).
\end{align*}

By Lemma~\ref{lemma:block_psd}, $\bGamma\succeq0$ is implied by
\begin{equation}\label{eq:block_psd_condition_optimized}
a_{\rm out}^2\,p^2\,n_i(n-n_i)\le\alpha\beta,
\qquad \alpha\ge0,\quad\beta\ge0,
\end{equation}
where
\begin{align*}
\alpha
&:=\lambda-a_{\rm in}(\lambda)\,p\,n_i-n\mu\zeta r_{\max}^2,\\
\beta
&:=\lambda-a_{\rm in}(\lambda)\,p\,n_i
-a_{\rm out}\,p(n-2n_i)
+n\mu\Big((1-t)R-\frac{\zeta}{t}r_{\max}^2\Big).
\end{align*}

\subsubsection{Admissible regimes}
\label{appendix:combined_admissible}

It remains to verify~\eqref{eq:block_psd_condition_optimized}
under~\eqref{eq:combined_single}. Write
\[
p:=p_{\max},\qquad
a:=1-4p,\qquad
b:=\frac{p}{1-p},\qquad
q:=\zeta r_{\max}^2,\qquad
B:=K\mu q.
\]
We first consider $0<p<1/4$.

\paragraph{Step 6: scalar reduction.}
$\alpha$ and $\beta$ increase with $\lambda$, while the off-diagonal bound is independent of $\lambda$. We take the largest value allowed by~\eqref{eq:gen_lambda_cond},
$\lambda=(1\!-\!2p)n_i$, for which $a_{\rm in}(\lambda)=2$. With
\[
s:=a-B,
\qquad
R_{\rm eff}(t):=(1-t)R-q\left(\frac1t-1\right),
\]
the quantities in~\eqref{eq:block_psd_condition_optimized} become
\[
\alpha=n_i s,\qquad
\beta=n_i\bigl(s-(K-2)b+K\mu R_{\rm eff}(t)\bigr),
\qquad
a_{\rm out}^2p^2n_i(n-n_i)=n_i^2(K-1)b^2.
\]
Since $(K-1)b^2>0$, the block condition requires $s>0$ and is equivalent to
\begin{equation}\label{eq:scalar_condition}
K\mu R_{\rm eff}(t)\ge h(s),
\qquad
h(s):=-s+(K-2)b+\frac{(K-1)b^2}{s}.
\end{equation}
Indeed, this inequality also ensures
$\beta/n_i\ge (K-1)b^2/s>0$.

\paragraph{Step 7: optimizing the split.}
For $q>0$,
\[
\max_{0<t\le1}R_{\rm eff}(t)
=\bigl(\sqrt R-\sqrt q\,\bigr)_+^2.
\]
The maximum is attained at $t=\sqrt{q/R}$ when $R>q$, and at $t=1$ when $R\le q$. If $q=0$, the residual contribution in Step~3 vanishes, so its bound holds directly with the full separation term $R$. Thus, in either case, the block condition is ensured by
\begin{equation}\label{eq:budget_certificate}
B<a,
\qquad
h(a-B)\le
\bigl(\sqrt{K\mu R}-\sqrt B\,\bigr)_+^2.
\end{equation}

\paragraph{Step 8: explicit residual budget.}
For $B<a$, the right-hand side of~\eqref{eq:budget_certificate} is bounded below by
\[
T:=\bigl(\sqrt{K\mu R}-\sqrt a\,\bigr)_+^2.
\]
It therefore suffices to require $h(a-B)\le T$. The function $h$ is strictly decreasing on $(0,\infty)$, since
\[
h'(s)=-1-\frac{(K-1)b^2}{s^2}<0.
\]
Its unique positive solution to $h(s_T)=T$ satisfies
\[
s_T^2+\bigl(T-(K-2)b\bigr)s_T-(K-1)b^2=0,
\]
and hence
\begin{equation}\label{eq:sT_app}
s_T=
\frac{2(K-1)b^2}
{T-(K-2)b+\sqrt{T^2-2(K-2)bT+K^2b^2}}.
\end{equation}
Consequently,
\[
h(a-B)\le T
\quad\Longleftrightarrow\quad
B\le a-s_T=\hat B(R,p).
\]
Because $s_T>0$, this condition also gives $B<a$. Thus~\eqref{eq:combined_single} implies \eqref{eq:budget_certificate}, and therefore \eqref{eq:block_psd_condition_optimized}.

Finally, if $p=0$, take $\lambda=n_i$ and $t=1$. The off-diagonal bound vanishes and
\[
\alpha=\beta=n_i(1-B).
\]
The block condition therefore holds whenever $B\le1=\hat B(R,0)$, including equality.

In all cases, the certificate satisfies \eqref{eq:optimality_L1}--\eqref{eq:optimality_ineq_simplified}. The global-optimality argument of App.~\ref{appendix:combined_optimality} proves that the perfect-recovery solution is a global minimizer of~\eqref{eq:main_model}.
\qed

\subsection{Proof of Cor.~\ref{cor:combined_graph_only}}
\begin{proof}
Let $p_{\max}\in[0,1/4)$ be the graph misconnection rate. Thm.~\ref{thm:combined_recovery} certifies recovery when
\[
K\mu\zeta r_{\max}^2\le\hat B(R,p_{\max}).
\]
For all sufficiently small $\mu$, we have $K\mu R\le1-4p_{\max}$, so $T=0$ and this condition becomes
\[
K\mu\zeta r_{\max}^2
\le
\frac{1-(K+4)p_{\max}+4p_{\max}^2}{1-p_{\max}}.
\]
The numerator has its smaller root at $p_K^\star$. If $p_{\max}<p_K^\star$, the right-hand side is a fixed positive number, whereas the left-hand side tends to zero as $\mu\to0$. Hence the theorem's condition holds for all sufficiently small $\mu$. If $p_{\max}>p_K^\star$, the right-hand side is negative, so the condition fails because its left-hand side is nonnegative.
\end{proof}

\subsection{Proof of Cor.~\ref{cor:combined_node_only}}
\begin{proof}
Write $p:=p_{\max}<1/4$. Since $R>0$, we have
$T\to\infty$ as $\mu\to\infty$, 
and therefore
\[
\hat B(R,p)\longrightarrow1-4p>0.
\]
If $\rho\le\gamma/K$, then $r_{\max}=0$, so the condition of Thm.~\ref{thm:combined_recovery} reduces to $0\le\hat B(R,p)$, which holds for all sufficiently large $\mu$.

If $\rho>\gamma/K$ and $\zeta>0$, then
\[
K\mu\zeta r_{\max}^2
=K\mu\zeta(\rho-\gamma/K)^2
\longrightarrow+\infty,
\]
whereas $\hat B(R,p)\le1-4p$. Hence~\eqref{eq:combined_single} fails for all sufficiently large $\mu$.
\end{proof}

\subsection{Proof of Cor.~\ref{cor:combined_both}}

\begin{proof}
Write $p:=p_{\max}$. The variability assumption gives
$r_{\max}=(\rho-\gamma/K)_+\le\sqrt{B/(K\mu\zeta)}$, 
and hence
$K\mu\zeta\, r_{\max}^2\le B$.
We therefore prove that each of the two cases gives
$B\le\hat B(R,p)$, 
which implies the condition of Thm.~\ref{thm:combined_recovery}.
Suppose first that $p>0$, and set
\[
a:=1-4p,\qquad
b:=\frac{p}{1-p},\qquad
T:=\bigl(\sqrt{K\mu R}-\sqrt a\,\bigr)_+^2.
\]
By the definition of the theorem's budget, $\hat B(R,p)=a-s_T$, where $s_T>0$ is the positive root of
\[
s_T^2+\bigl(T-(K-2)b\bigr)s_T-(K-1)b^2=0.
\]
Equivalently, $h(s_T)=T$ for
\[
h(s):=-s+(K-2)b+\frac{(K-1)b^2}{s}
=\frac{\bigl((K-1)b-s\bigr)(s+b)}{s}.
\]
Since $h'(s)=-1-(K-1)b^2/s^2<0$, the conditions
\(
s>0 
\)
and
\(
h(s)\le T
\)
imply
\(
s\ge s_T.
\)
Thus, with $s:=a-B$, it suffices to establish $s>0$ and $h(s)\le T$.

\textbf{\emph{Case (a).}}
The threshold $p_K^\star(B)$ is the smaller root of
\[
4p^2-(K+4-B)p+1-B=0.
\]
Consequently, $p\le p_K^\star(B)$ gives
\[
0\le4p^2-(K+4-B)p+1-B
=(1-p)\left(a-B-(K-1)\frac{p}{1-p}\right).
\]
Since $1-p>0$, this yields
\(
s=a-B\ge(K-1)b>0.
\)
The factorized expression for $h$ then gives $h(s)\le0$. Since $T\ge0$, we obtain $h(s)\le T$ without imposing a lower bound on $R$. It follows that $s\ge s_T$, and therefore $B\le a-s_T=\hat B(R,p)$.

\textbf{\emph{Case (b).}}
The upper bound $p<(1-B)/4$ gives $s=1-4p-B>0$.
The lower bound $p>p_K^\star(B)$ gives $s<(K-1)b$, so
\[
\Delta:=h(s)
=\frac{\bigl((K-1)b-s\bigr)(s+b)}{s}>0.
\]
The separation assumption is precisely
\(
K\mu R\ge
R_{\min}=\bigl(\sqrt a+\sqrt\Delta\bigr)^2.
\)
Taking square roots and subtracting $\sqrt a$ gives
\(
\sqrt{K\mu R}-\sqrt a\ge\sqrt\Delta>0,
\)
and hence
\[
T=\bigl(\sqrt{K\mu R}-\sqrt a\,\bigr)^2
\ge\Delta=h(s).
\]
Again $s\ge s_T$, which gives
$B\le\hat B(R,p)$.

If $p=0$, case (a) requires $B\le1$, while case (b) is impossible. Using $\hat B(R,0)=1$, we obtain the same budget bound.

In every case,
\[
K\mu\zeta r_{\max}^2\le B\le\hat B(R,p).
\]
Thus~\eqref{eq:combined_single} holds, and
Thm.~\ref{thm:combined_recovery} certifies perfect recovery.
\end{proof}
\subsection{Proof of Theorem~\ref{thm:combined_strongly_convex}}
\label{appendix:combined_strongly_convex}

This is a specialization of Thm.~\ref{thm:combined_recovery} under the assumption that each node loss $f_v$ and each class loss $F_i$ is $m$-strongly convex on its domain. Recall $r_i:=(\rho_i-\gamma/K)_+$, $r_{\max}:=\max_i r_i$, and $\rho:=\max_i\rho_i$. Let $\zeta$ be the residual-price coefficient defined in~\eqref{eq:zeta}, and write $p:=p_{\max}$. Fix an admissible residual budget $K\mu\zeta r_{\max}^2 \le B < 1-4p$. Write $R_{\min}$ for the separation threshold in Thm.~\ref{thm:combined_recovery}.

\paragraph{Step 1: strongly convex losses control the residual price.}
By $m$-strong convexity, 
$
D_{f_v}^{\bs_v}(\btheta,\btheta_i^\star)
\ge\frac{m}{2}\|\btheta-\btheta_i^\star\|_2^2
$
for every $v\in\calC_i$, so
$
\underline D_i(\btheta)
\ge\frac{m}{2}\|\btheta-\btheta_i^\star\|_2^2.
$
From the definition of the residual-price coefficient, we obtain
\begin{equation}\label{eq:zeta_strong_convex_bound}
\zeta_i
=\sup_{\btheta\in\Theta_f}
\frac{\|\btheta-\btheta_i^\star\|_2^2}
{4\underline D_i(\btheta)}
\le\frac1{2m},
\qquad
\zeta\le\frac1{2m}.
\end{equation}

\paragraph{Step 2: lower bound on $\gamma$ from the residual condition.}
By~\eqref{eq:zeta_strong_convex_bound},
$
K\mu\zeta r_{\max}^2
\le\frac{K\mu r_{\max}^2}{2m},
$
so the budget condition $K\mu\zeta r_{\max}^2\le B$ is ensured by
$
r_{\max}\le\sqrt{\nicefrac{2mB}{K\mu}}.
$
Since $r_{\max}=(\rho-\gamma/K)_+$, this is equivalent to
\begin{equation}\label{eq:gamma_lower_strong_convex}
\gamma\ge
K\left(\rho-\sqrt{\frac{2mB}{K\mu}}\right)_+.
\end{equation}

\paragraph{Step 3: upper bound on $\gamma$ from the separation condition.}
Let $\btheta_i^{(0)}\in\arg\min_{\btheta}F_i(\btheta)$ and define
\[
d_{\min}^{(0)}:=\min_{i\neq j}
\|\btheta_i^{(0)}-\btheta_j^{(0)}\|_2,
\qquad
d_{\min}:=\min_{i\neq j}
\|\btheta_i^\star-\btheta_j^\star\|_2.
\]
Recall that $\bzero\in\partial F_i(\btheta_i^{(0)})$ and
$-\bomega_i\in\partial F_i(\btheta_i^\star)$, where, by balance,
\[
\bomega_i=\frac{\gamma}{K}\sum_{j\neq i}\bg_{ij},
\qquad
\bg_{ji}=-\bg_{ij},
\qquad
\|\bg_{ij}\|_2\le1.
\]
Strong monotonicity of $\partial F_i$ gives
$
m\|\btheta_i^{(0)}-\btheta_i^\star\|_2^2
\le\langle\bomega_i,\btheta_i^{(0)}-\btheta_i^\star\rangle,
$
which, by completing the square, yields
\begin{equation}
\label{eq:centroid_displacement_ball_strong_convex}
\left\|\btheta_i^{(0)}-\btheta_i^\star
-\frac{\bomega_i}{2m}\right\|_2
\le\frac{\|\bomega_i\|_2}{2m}.
\end{equation}

Suppose first that $d_{\min}>0$, and choose a pair $i\neq j$ attaining $d_{\min}$. Set
\[
\mathbf e:=\frac{\btheta_i^\star-\btheta_j^\star}{d_{\min}},
\qquad
\mathbf a:=\sum_{k\notin\{i,j\}}\bg_{ik},
\qquad
\mathbf b:=\sum_{k\notin\{i,j\}}\bg_{jk}.
\]
For every $k\notin\{i,j\}$, the closest-pair property implies
\[
\|\bg_{ik}-\bg_{jk}\|_2
\le
\frac{\|\btheta_i^\star-\btheta_j^\star\|_2}
{\min\{\|\btheta_i^\star-\btheta_k^\star\|_2,
\|\btheta_j^\star-\btheta_k^\star\|_2\}}
\le1.
\]
Since
$
\bomega_i=\frac{\gamma}{K}(\mathbf e+\mathbf a),
% \qquad
\bomega_j=\frac{\gamma}{K}(-\mathbf e+\mathbf b),
$
we obtain
\begin{equation}
\label{eq:closest_pair_force_difference}
\|\bomega_i-\bomega_j\|_2
\le\frac{\gamma}{K}\bigl(2+K-2\bigr)=\gamma.
\end{equation}
Moreover, $\|\mathbf a\|_2,\|\mathbf b\|_2\le K-2$ and
$\langle\mathbf e,\mathbf a-\mathbf b\rangle\le K-2$, so
\begin{equation*}
\|\bomega_i\|_2^2+\|\bomega_j\|_2^2
=\frac{\gamma^2}{K^2}
\left(2+2\langle\mathbf e,\mathbf a-\mathbf b\rangle
+\|\mathbf a\|_2^2+\|\mathbf b\|_2^2\right)\\
\le\frac{2\gamma^2}{K^2}(K^2-3K+3).
\end{equation*}
Consequently, Cauchy--Schwarz gives
\begin{equation}
\label{eq:closest_pair_force_norm_sum}
\|\bomega_i\|_2+\|\bomega_j\|_2
\le\frac{2\gamma}{K}\sqrt{K^2-3K+3}.
\end{equation}
Applying~\eqref{eq:centroid_displacement_ball_strong_convex} to the selected pair and using
\eqref{eq:closest_pair_force_difference}--\eqref{eq:closest_pair_force_norm_sum},
\begin{align*}
d_{\min}^{(0)}
&\le\|\btheta_i^{(0)}-\btheta_j^{(0)}\|_2\\
&\le d_{\min}
+\frac{\|\bomega_i-\bomega_j\|_2}{2m}
+\frac{\|\bomega_i\|_2+\|\bomega_j\|_2}{2m}\\
&\le d_{\min}
+\frac{\gamma}{m}
\left(\frac12+\frac{\sqrt{K^2-3K+3}}K\right),
\end{align*}
Hence
\begin{equation}\label{eq:biased_centroid_separation_lower_strong_convex}
d_{\min}\ge d_{\min}^{(0)}-\frac{c_K\gamma}{m}.
\end{equation}
\[
c_K:=
\begin{cases}
1, & p\le p_K^\star(B),\\[2pt]
\displaystyle \frac12+\frac{\sqrt{K^2-3K+3}}{K},
& p>p_K^\star(B).
\end{cases}
\]

If $p\le p_K^\star(B)$, no positive separation is required,  $R_{\min}=0$. The noncollision argument in App.~\ref{appendix:node_only_sufficient} ensures distinct biased centroids whenever $\gamma<m\,d_{\min}^{(0)}$. Strong convexity then gives node-wise $0$-separability, and Cor.~\ref{cor:combined_both}(a) applies. Thus the upper endpoint holds with $c_K=1$ in this case.

By Remark~\ref{cor:strong_convexity_R_separability}, the biased centroids are node-wise $R$-separable for every $R<m d_{\min}^2/8$. For $R_{\min}>0$, the separation condition of Thm.~\ref{thm:combined_recovery} is
$K \mu R\ge R_{\min}.$
A value of $R$ satisfying both requirements therefore exists whenever
$
d_{\min}^{(0)}-\frac{c_K\gamma}{m}
>
\sqrt{\frac {8 R_{\min}}{K \mu m}}.
$
Solving for $\gamma$ gives
\begin{equation}\label{eq:gamma_upper_strong_convex}
\gamma<\frac{m}{c_K}
\left(
d_{\min}^{(0)}
-\sqrt{\frac{8R_{\min}}{K \mu m}}
\right).
\end{equation}

\paragraph{Step 4: admissible window.}
Combining~\eqref{eq:gamma_lower_strong_convex} and~\eqref{eq:gamma_upper_strong_convex} gives
\[
K\left(\rho-\sqrt{\frac{2mB}{K\mu}}\right)_+
\le\gamma
<\frac{m}{c_K}
\left(
d_{\min}^{(0)}
-\sqrt{\frac{8R_{\min}}{K \mu m}}
\right).
\]
Whenever this interval is non-empty for an admissible budget $B$, both the residual-budget and separation conditions of Thm.~\ref{thm:combined_recovery} hold.

If $R_{\min}\le0$, the choice $t=1$ removes the additional separation requirement. In that case, the noncollision argument in App.~\ref{appendix:node_only_sufficient} ensures distinct centroids whenever
$
\gamma<m\,d_{\min}^{(0)}.
$
Combining this with~\eqref{eq:gamma_lower_strong_convex} gives the stated window for $R_{\min}\le0$.
\qed

\subsection{Proof of Corollary~\ref{cor:combined_svr}}
\label{appendix:combined_svr}

Assume $\rho>0$ and fix an admissible budget $B$. The interval~\eqref{eq:comb_gamma} is nonempty if and only if
\[
K\left(\rho-\sqrt{\frac{2mB}{K\mu}}\right)_+
<
\frac{m}{c_K}
\left(
d_{\min}^{(0)}
-\sqrt{\frac{8R_{\min}}{K\mu m}}
\right).
\]
Rearranging and dividing by $\rho$ gives
\[
\frac{d_{\min}^{(0)}}{\rho}
>
\frac{c_KK}{m}
\left(1-\frac1\rho\sqrt{\frac{2mB}{K\mu}}\right)_+
+\frac1\rho\sqrt{\frac{8R_{\min}}{K\mu m}},
\]
which is exactly~\eqref{eq:comb_svr}. The definitions give $\alpha=0$ whenever
$K\mu\le2mB/\rho^2$, and $\beta\to0$ as
$\mu\to\infty$.

To establish the optimized guarantee, suppose $B>0$ and consider two choices of $\mu$.

First, taking
$
K\mu=\frac{2mB}{\rho^2}
$
gives $\alpha=0$ and
\[
\beta
=\frac1\rho
\sqrt{\frac{8R_{\min}\rho^2}{2m^2B}}
=\frac2m\sqrt{\frac{R_{\min}}{B}}.
\]
Hence the interval is nonempty, and recovery is certified, whenever
$
\mathrm{SVR}>
\frac2m\sqrt{\nicefrac{R_{\min}}{B}}.
$

Second, suppose $\mathrm{SVR}>K/m$. Then we can choose $\gamma$ with
$
K\rho<\gamma<m\,d_{\min}^{(0)}.
$
The lower bound gives $r_{\max}=0$, so the residual budget condition holds for every $\mu$.
The upper bound, by the noncollision argument in App.~\ref{appendix:node_only_sufficient}, ensures that the biased centroids are pairwise distinct. With
$
d_{\min}:=\min_{i\neq j}
\|\btheta_i^\star-\btheta_j^\star\|_2>0,
$
strong convexity gives node-wise $R$-separability for any fixed $0<R<m\,d_{\min}^2/8$. These centroids and this choice of $R$ are independent of $\mu$. Since $R_{\min}$ is also independent of $\mu$, the condition $K\mu R\ge R_{\min}$ holds for all sufficiently large $\mu$. Cor.~\ref{cor:combined_both} therefore certifies recovery whenever $\mathrm{SVR}>K/m$.

Combining the two choices, recovery is certified whenever
\[
\mathrm{SVR}>
\min\left\{
\frac{K}{m},
\frac2m\sqrt{\frac{R_{\min}}{B}}
\right\}
=
\frac{K}{m}
\min\left\{
1,\frac2K\sqrt{\frac{R_{\min}}{B}}
\right\},
\]
as claimed.
\qed

\subsection{Residual-price coefficients for different loss families}
\label{appendix:zeta_loss_families}

The combined recovery condition in Theorem~\ref{thm:combined_recovery} depends on the residual-price coefficient in \eqref{eq:zeta}
\[
\zeta \;:=\; \max_i \sup_{\boldsymbol{\theta} \in \Theta_f} \frac{\|\boldsymbol{\theta} - \boldsymbol{\theta}_i^\star\|_2^2}{4 \min_{v \in \mathcal{C}_i} D^{\mathbf{s}_v}_{f_v}(\boldsymbol{\theta}, \boldsymbol{\theta}_i^\star)},
\]
where $\bs_v\in\partial f_v(\btheta_i^\star)$, with the conventions $0/0:=0$ and $c/0:=+\infty$ for $c>0$. In this subsection, we bound $\zeta$ for several families of convex losses.

If $\zeta\le\bar\zeta$ for some $0<\bar\zeta<\infty$, then the residual-budget condition $K\mu\zeta r_{\max}^2\le B$ is ensured by
$
\rho\le\frac{\gamma}{K}
+\sqrt{\frac{B}{K\mu\bar\zeta}},
$
or equivalently,
\[
\gamma\ge
K\left(
\rho-\sqrt{\frac{B}{K\mu\bar\zeta}}
\ \right)_+.
\]
The admissibility and separation conditions of Theorem~\ref{thm:combined_recovery} must also hold.

\paragraph{Uniformly convex losses.}

Assume
$
D_{f_v}^{\bs_v}(\btheta,\btheta_i^\star)
\ge
\varphi_v(\|\btheta-\btheta_i^\star\|_2),
% \qquad 
\forall v\in\calC_i,
$
where each $\varphi_v$ is increasing and convex, $\varphi_v(0)=0$, and $\varphi_v(t)>0$ for $t>0$. Define
$
\varphi(t):=\min_v\varphi_v(t).
$
If
$
\bar\zeta_\varphi
:=
\sup_{t>0}\nicefrac{t^2}{4\varphi(t)}
<\infty,
$
then $\zeta\le\bar\zeta_\varphi$, and the residual budget is ensured by
$
\gamma\ge
K\left(
\rho-\sqrt{\nicefrac{B}{K\mu\bar\zeta_\varphi}}
\right)_+.
$
Thus uniform convexity supplies a finite coefficient whenever the given modulus dominates a positive quadratic function.
In particular, for the quadratic modulus
$
\varphi(t)=\frac{m}{2}t^2,
$
we obtain
\[
\zeta\le\frac1{2m},
\qquad
\gamma\ge
K\left(
\rho-\sqrt{\frac{2mB}{K\mu}}
\right)_+.
\]
This recovers the residual-budget condition in Theorem~\ref{thm:combined_strongly_convex}.

\paragraph{Bounded-domain uniformly convex losses.}

Suppose, in addition, that
\[
0<\ell
:=
\max_i\sup_{\btheta\in\Theta_f}
\|\btheta-\btheta_i^\star\|_2
<\infty.
\]
Only distances up to $\ell$ need to be considered.
Hence, if
$
\bar\zeta_{\varphi,\ell}
:=
\sup_{0<t\le\ell} t^2/(4\varphi(t))
<\infty,
$
then
$
\zeta\le\bar\zeta_{\varphi,\ell},
$
and the residual budget is ensured by
$\gamma\ge K\bigl(\rho-\sqrt{B/(K\mu\bar\zeta_{\varphi,\ell})}\bigr)_+.$

For the polynomial modulus $\varphi(t)=(c/q)\,t^q$ with $c>0$ and
$1<q\le 2$, $\sup_{0<t\le\ell}t^{2-q}=\ell^{2-q}$, so
$\bar\zeta_{\varphi,\ell}=q\,\ell^{2-q}/(4c)$. Consequently,
$\zeta\le q\,\ell^{2-q}/(4c)$ and
$\gamma\ge K\bigl(\rho-\sqrt{4cB/(qK\mu\,\ell^{2-q})}\bigr)_+$. The
case $q=2$ recovers the strongly convex bound with $m=c$.

\paragraph{Ridge and elastic-net regularized losses.}

Let
\[
f_v(\btheta)
=
\ell_v(\btheta)
+\lambda\|\btheta\|_1
+\frac{\kappa}{2}\|\btheta\|_2^2,
\]
where $\ell_v$ is convex, $\lambda\ge0$, and $\kappa>0$.
Since $f_v$ is $\kappa$-strongly convex,
$
D_{f_v}^{\bs_v}(\btheta,\btheta_i^\star)
\ge
\frac{\kappa}{2}\|\btheta-\btheta_i^\star\|_2^2.
$
Therefore,
$
\zeta\le\frac1{2\kappa},
$
and the residual budget is ensured by
\[
\gamma\ge
K\left(
\rho-\sqrt{\frac{2\kappa B}{K\mu}}
\right)_+.
\]
This recovers the same residual-budget dependence as Theorem~\ref{thm:combined_strongly_convex}.

\newpage
\section{Lemmas}
\label{appendix:lemmas}

\begin{lemma}\label{lemma:block_psd}
Let
\[
\mathbf\Gamma=
\begin{bmatrix}
\mathbf\Gamma_{11} & \mathbf\Gamma_{12}\\[3pt]
\mathbf\Gamma_{12}^\top & \mathbf\Gamma_{22}
\end{bmatrix}.
\]
Assume
\[
\mathbf\Gamma_{11}\succeq \alpha\bI,
\qquad
\mathbf\Gamma_{22}\succeq \beta\bI,
\qquad
\alpha,\beta\ge0.
\]
Then \(\mathbf\Gamma\succeq0\) if
\[
\sigma_{\max}^2(\mathbf\Gamma_{12})\le \alpha\beta.
\]
\end{lemma}

\begin{proof}
For any \(\bx=[\bx_1^\top,\bx_2^\top]^\top\),
\begin{align*}
\bx^\top\mathbf\Gamma\bx
&=
\bx_1^\top\mathbf\Gamma_{11}\bx_1
+
2\bx_1^\top\mathbf\Gamma_{12}\bx_2
+
\bx_2^\top\mathbf\Gamma_{22}\bx_2
\\[2.5pt]
&\ge
\alpha\|\bx_1\|^2
+
\beta\|\bx_2\|^2
-
2\sigma_{\max}(\mathbf\Gamma_{12})\|\bx_1\|\|\bx_2\|
\\[2.5pt]
&=
(\sqrt{\alpha}\|\bx_1\|-\sqrt{\beta}\|\bx_2\|)^2
+
2(\sqrt{\alpha\beta}-\sigma_{\max}(\mathbf\Gamma_{12}))
\|\bx_1\|\|\bx_2\|.
\end{align*}
The right-hand side is nonnegative when
\(
\sigma_{\max}^2(\mathbf\Gamma_{12})\le\alpha\beta.
\)
Hence \(\mathbf\Gamma\succeq0\).
\end{proof}

\vspace{2pt}
\begin{lemma}[Spectral bound from sparsity and entry size]
\label{lem:spectral-sparsity-bound}
Let \(\bM\in\mathbb R^{n\times n}\) be symmetric. Suppose each row has at most \(d_{\max}\) nonzero entries and
\(
|M_{ij}|\le a
\)
for all \(i,j\). Then
\(
\|\bM\|_2\le a d_{\max}.
\)
\end{lemma}

\begin{proof}
Since \(\bM\) is symmetric,
\(
\|\bM\|_2\le \sqrt{\|\bM\|_1\|\bM\|_\infty}.
\)
Also,
\(
\|\bM\|_\infty
=
\max_i\sum_j|M_{ij}|
\le
ad_{\max}.
\)
By symmetry, \(\|\bM\|_1=\|\bM\|_\infty\). Hence
\(
\|\bM\|_2\le ad_{\max}.
\)
\end{proof}

\begin{lemma}\label{lem:centered_diagonal_bound}
Let $a_1,\ldots,a_m\in\bbR$ satisfy $\sum_{v=1}^m a_v=0$,
and define
\[
\ell:=\max_v a_v-\min_v a_v,
\qquad
\bP^\perp:=\bI_m-\frac{1}{m}\bone\bone^\top.
\]
Then, for every $\bx\in\bbR^m$,
\[
\left|\sum_{v=1}^m a_vx_v^2\right|
\le
\ell\|\bx\|_2\|\bP^\perp\bx\|_2.
\]
\end{lemma}

\begin{proof}
The claim is immediate for $\bx=\bzero$.
By homogeneity, it suffices to consider $\|\bx\|_2=1$.
Set
\[
\be:=\frac{\bone}{\sqrt m},
\qquad
c:=\frac{\max_v a_v+\min_v a_v}{2},
\qquad
\bT:=\operatorname{diag}(a_v-c).
\]
Then $\|\bT\|_2\le\ell/2$.
Since $\sum_v a_ve_v^2=0$ and
$\|\bx\|_2=\|\be\|_2=1$,
\[
\sum_v a_vx_v^2
=\bx^\top\bT\bx-\be^\top\bT\be
=(\bx-\be)^\top\bT(\bx+\be).
\]
Therefore,
\[
\left|\sum_v a_vx_v^2\right|
\le
\frac{\ell}{2}\|\bx-\be\|_2\|\bx+\be\|_2.
\]
Finally,
\[
\|\bx-\be\|_2\|\bx+\be\|_2
=
\sqrt{(2-2\be^\top\bx)(2+2\be^\top\bx)}
=
2\sqrt{1-(\be^\top\bx)^2}
=
2\|\bP^\perp\bx\|_2.
\]
Combining these inequalities proves the claim.
\end{proof}

\newpage
\section{Extended numerical validations}
\label{appendix:extended_sims}

\begin{figure}
  \centering
    \includegraphics[width=0.48\textwidth]{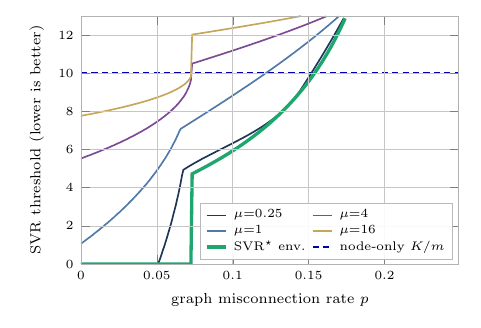}
    \hfill
    \includegraphics[width=0.48\textwidth]{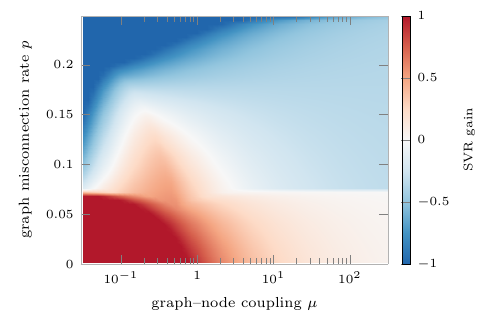}
    \caption{Numerical view of Corollary~\ref{cor:combined_svr}.
    \textbf{Left:} the combined SVR threshold $\mathrm{SVR}^{\star}_{\mathrm{comb}}(p,\mu)$ defined in~\eqref{eq:SVR_comb_app}, plotted as a function of the graph misconnection rate $p$ for $\mu \in \{0.25, 1, 4, 16\}$, together with the envelope
    $\mathrm{SVR}^{\star}(p)$    
    (thick green). The dashed horizontal line at $\mathrm{SVR}^{\star}_{\mathrm{node}} = K/m = 10$ is the node-only threshold. Lower is better: any data with SVR above the curve is recovered. For moderate $\mu$ (e.g.\ $\mu \in [1, 4]$) the combined threshold lies strictly below the node-only line over a wide range of $p$, confirming the strict-improvement claim of Corollary~\ref{cor:combined_svr}. Settings: $K=10$, $m=1$, $\rho=0.5$.
    \textbf{Right:} the relative improvement $1 - \mathrm{SVR}^{\star}_{\mathrm{comb}}/\mathrm{SVR}^{\star}_{\mathrm{node}}$ as a function of $(\mu, p)$. Red indicates regimes where the combined formulation requires a strictly weaker SVR than node-only; blue indicates regimes where the combined bound provides no improvement.
    }
  \label{fig:svr_view}
\end{figure}

\paragraph{Discussion of Fig.~\ref{fig:svr_view}.}
Fig.~\ref{fig:svr_view} provides a numerical interpretation of the strongly-convex specialization in Thm.~\ref{thm:combined_strongly_convex} and Cor.~\ref{cor:combined_svr}.
The minimization over \(B\) is performed numerically on a uniform grid of \(800\) points in the interval \((0,1-4p)\).
We fix \(K=10\), \(m=1\), and \(\rho=0.5\) throughout this experiment.
In the left panel of Fig.~\ref{fig:svr_view}, we sweep the graph misconnection rate \(p\) for four values of the graph--node coupling parameter \(\mu\in\{0.25,1,4,16\}\), and overlay the envelope \(\mathrm{SVR}^{\star}(p):=\min_{\mu,\,B}\,\mathrm{SVR}^{\star}_{\rm comb}(p,\mu,B)\) obtained by optimizing jointly over \(\mu\) and \(B\).
In the right panel, we jointly sweep \((\mu,p)\) and report the relative improvement
\(
1-\frac{\mathrm{SVR}_{\rm comb}^\star}
{\mathrm{SVR}_{\rm node}^\star}.
\)
Recall that the node-information-only regime of Cor.~\ref{cor:node_only_nonempty} requires
\begin{align}
\mathrm{SVR}_{\rm node}^\star
>
\frac{K}{m},
\label{eq:SVR_node_app}
\end{align}
while Cor.~\ref{cor:combined_svr} shows that, in the combined regime,
\begin{align}
    \label{eq:SVR_comb_app}
\mathrm{SVR}_{\rm comb}^\star
>
\alpha\, c_K\, \frac{K}{m}
+
\beta,
\end{align}
where
\(
\alpha
:=
\Bigl(1-\tfrac{1}{\rho}\sqrt{\tfrac{2mB}{K\mu}}\Bigr)_{+}
\le 1
\)
captures the gain from graph structure,
\(
\beta
:=
\tfrac{1}{\rho}\sqrt{\tfrac{8R_{\min}}{K\mu\,m}}
\)
quantifies the price of graph noise, and
\(c_K:=\tfrac12+\tfrac{\sqrt{K^{2}-3K+3}}{K}\) for \(p\ge p_K^{\star}(B)\) 
(with \(c_K=1\) otherwise).
Therefore, whenever
\[
\alpha\, c_K\,\frac{K}{m}+\beta
<
\frac{K}{m},
\]
the combined formulation requires a strictly weaker separation-to-variability ratio than the node-only regime. In case~(a) (\(p\le p_K^{\star}(B)\)) this holds for every \(B>0\) because \(c_K=1\) and \(\alpha<1\); in case~(b) (\(p>p_K^{\star}(B)\)) the factor \(c_K>1\) is a genuine cost of the strong-convexity bound, and strict improvement over the node-only requirement is obtained only after optimizing over \(\mu\) and \(B\), in which case Cor.~\ref{cor:combined_svr} gives the envelope \(\mathrm{SVR}^{\star}(p,B)=\tfrac{K}{m}\min\{1,\tfrac{2}{K}\sqrt{R_{\min}/B}\}\).

The left panel evaluates the combined threshold \(\mathrm{SVR}_{\rm comb}^\star(p,\mu)\) as a function of the graph misconnection rate \(p\), together with the envelope \(\mathrm{SVR}^{\star}(p):=\min_{\mu,B}\mathrm{SVR}_{\rm comb}^\star(p,\mu,B)\) (thick green). The dashed horizontal line corresponds to the node-only threshold from Cor.~\ref{cor:node_only_nonempty}. For moderate values of \(\mu\), the combined threshold lies strictly below the node-only baseline over a substantial range of graph noise levels, and the envelope stays below it for all \(p\lesssim 0.154\), confirming the improvement predicted by Cor.~\ref{cor:combined_svr}.
As \(p\) increases, the graph-noise price \(\beta\) becomes dominant and the benefit of graph structure gradually disappears. Likewise, when \(\mu\) becomes too small or too large, the formulation approaches the graph-only and node-only limits of Cor.~\ref{cor:combined_graph_only} and Cor.~\ref{cor:combined_node_only}, reducing the gain obtained from combining the two sources.

The right panel visualizes the relative improvement
\(
1-
\frac{\mathrm{SVR}_{\rm comb}^\star}
{\mathrm{SVR}_{\rm node}^\star},
\)
over the \((\mu,p)\)-plane. Red regions indicate regimes where the combined formulation provably relaxes the node-only requirement, while blue regions correspond to parameter choices where the graph information is either too noisy or too weakly coupled to provide a meaningful benefit. The sharp transition near \(p\approx p_K^{\star}\approx 0.073\) is the case-(a)/(b) boundary of Cor.~\ref{cor:combined_both}, while the admissibility condition
\(
p_{\max}<\frac{1-B}{4}
\)
from Cor.~\ref{cor:combined_both} caps the gain region at the top, beyond which graph corruption dominates the recovery guarantee.

Overall, Fig.~\ref{fig:svr_view} numerically illustrates the main message of Thms.~\ref{thm:combined_recovery} and \ref{thm:combined_strongly_convex}: graph structure can provably reduce the amount of feature separation required for perfect recovery.

\begin{figure}
  \centering
  \includegraphics[width=0.48\textwidth]{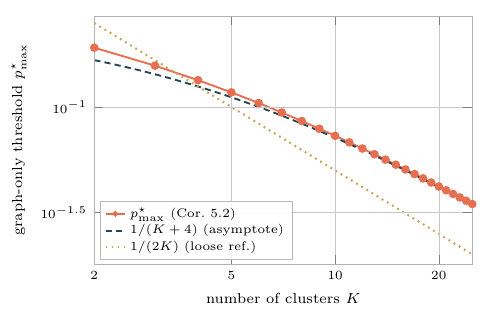}
  \hfill
  \includegraphics[width=0.48\textwidth]{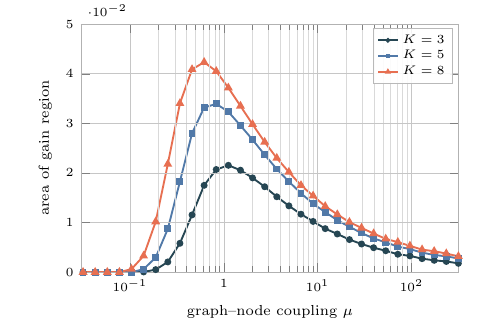}
  \caption{Threshold scaling and gain volume.
    \textbf{Left:} the graph-only threshold $p^{\star}_{\max}(K) = ((K+4) - \sqrt{(K+4)^2 - 16})/8$ of Corollary~\ref{cor:combined_graph_only} (red, with markers), evaluated for $K \in \{2,\dots,25\}$, against the asymptotic approximation $1/(K+4)$ (gray) predicted by the same corollary and the looser SBM-style reference $1/(2K)$ (olive). At $K=50$ the closed form and the asymptote agree to four digits;  for $K\ge 4$ the closed form lies above $1/(2K)$.
    \textbf{Right:} area of the gain region $\mathcal{G}_{\mu,K}$ in the $(p,\rho)$ plane as a function of $\mu$, for $K \in \{3, 5, 8\}$. Settings: $\gamma=2$, $m=1$, $d_{\min}^{(0)}=4.7$. The gain volume decreases toward zero at both ends; $\mu \to 0$ (the formulation reduces to graph-only) and $\mu \to \infty$ (it reduces to node-only). It also attains an interior maximum near $\mu^{\star}\approx 0.6\!-\!1.1$. The gain volume grows with $K$: the harder the problem (more classes, smaller per-class budget), the more the combined formulation buys you.}
  \label{fig:gain_volume}
\end{figure}

\paragraph{Discussion of Fig.~\ref{fig:gain_volume}.}
Fig.~\ref{fig:gain_volume} studies the scaling behavior of the recovery thresholds and the size of the gain region predicted by the theory.

The left panel evaluates the graph-only threshold \(p_K^\star(K)\) from Cor.~\ref{cor:combined_graph_only} as a function of the number of clusters \(K\). The exact deterministic threshold closely follows the asymptotic scaling
\[
p_K^\star(K)\sim \frac{1}{K+4},
\]
derived in Cor.~\ref{cor:combined_graph_only},  and it lies above the simpler reference scaling \(1/(2K)\) for every \(K\ge 4\). This confirms that the admissible graph noise shrinks approximately as \(O(1/K)\), matching the qualitative behavior of classical SBM recovery thresholds while remaining fully deterministic and instance-specific. In particular, the figure illustrates the increasing difficulty of graph-only recovery as the number of classes grows.

The right panel evaluates the area of the gain region \(\mathcal G_{\mu,K}\) in the \((p_{\max},\rho)\)-plane, namely the set of parameters where the combined formulation succeeds while neither the graph-only nor the node-only regime alone is guaranteed to recover perfectly. This region is predicted by Thm.~\ref{thm:combined_recovery} and Cor.~\ref{cor:combined_svr}. The curves exhibit a non-monotone dependence on the coupling parameter \(\mu\): the gain vanishes both as \(\mu\to0\), where the formulation reduces to the graph-only regime, and as \(\mu\to\infty\), where it approaches the node-only regime. Between these extremes, there exists an intermediate coupling regime where the synergy between graph structure and node information is maximized.

Moreover, the gain region grows with the number of clusters \(K\). This is consistent with the scaling of Cor.~\ref{cor:combined_graph_only} and Cor.~\ref{cor:node_only_nonempty}: as \(K\) increases, both the graph-only and node-only recovery conditions become more restrictive, creating a larger parameter regime where neither source alone is sufficient, but their combination still achieves perfect recovery. Fig.~\ref{fig:gain_volume} therefore illustrates the main qualitative conclusion of the paper: the benefit of combining graph structure and node information becomes increasingly pronounced in harder classification problems.

\newpage
\section{An atomic conditional-gradient solver}
\label{sec:algorithm}

The unified problem~\eqref{eq:main_model} is convex in the weights $\{\lambda_i\}$ of the atoms, and the set of atoms $\calA$ is infinite. We solve it with an atomic conditional-gradient scheme in the spirit of CoGEnT~\citep{rao2015forward}: at every outer iteration we (i)~add a single new atom (forward step), (ii)~refine the active set by alternating minimization (enhancement step), and (iii)~prune redundant atoms (backward step). 

\subsection{Reduced parameterization and reduced objective}
\label{sec:alg_setup}

We maintain a finite active set of $k$ atoms $\{(\be_i,\btheta_i)\}_{i=1}^k$, with $(\be_i/\|\be_i\|_2,\btheta_i)\in\calA$, where each $\be_i\in\bbR^n$ has nonnegative entries $e_{i,v}\ge 0$ that absorb the atomic coefficients $\lambda_i$ (so that $\|\be_i\|_2^2=\lambda_i$) via the normalization $L_{vv}=\sum_i e_{i,v}^2=1$. Define the \emph{node--atom weights}
\begin{equation}\label{eq:alg-weights}
\omega_{v,i}\;:=\;e_{i,v}^{\,2},
\qquad
\sum_{i=1}^{k}\omega_{v,i}=1\quad\forall v\in\calV,
\end{equation}
so the weights $\boldsymbol{\Omega}\in\bbR_+^{n\times k}$ form a row-stochastic matrix; each row $\boldsymbol{\Omega}_{v,:}$ lies in the simplex $\Delta_k:=\{\bomega\in\bbR^k_{\ge 0}:\sum_i\omega_i=1\}$. Under~\eqref{eq:alg-weights}, the lifted variables of~\eqref{eq:unified_atomic_problem} take the compact form
\begin{equation}\label{eq:alg-induced}
\bL(\boldsymbol{\Omega})
=\sum_{i=1}^k\sqrt{\boldsymbol{\Omega}_{:,i}}\sqrt{\boldsymbol{\Omega}_{:,i}}^{\!\top},
\qquad
\btheta_v(\boldsymbol{\Omega},\{\btheta_i\})
=\sum_{i=1}^k\omega_{v,i}\,\btheta_i,\quad \forall v\in\calV,
\end{equation}
where $\sqrt{\boldsymbol{\Omega}_{:,i}}$ denotes the entrywise square-root of the $i$-th column. The diagonal constraint $L_{vv}=1$ is exactly the row-sum constraint in~\eqref{eq:alg-weights}; the box constraint $0\le L_{uv}\le 1$ then follows from the Cauchy--Schwarz inequality together with $e_{i,v}\ge 0$. The polarized program~\eqref{eq:unified_atomic_problem} therefore becomes
{\small
\begin{equation}\label{eq:alg-objective}
\Phi_k(\boldsymbol{\Omega},\{\btheta_i\})
=
\underbrace{-\langle\bar\bA,\bL(\boldsymbol{\Omega})\rangle}_{\text{graph term}}
+
n\mu\!\!\underbrace{\sum_{v\in\calV}\sum_{i=1}^k\omega_{v,i}\,f_v(\bz_v;\btheta_i)}_{\text{node loss (linear in $\boldsymbol{\Omega}$)}}
+
\gamma\mu\!\!\underbrace{\sum_{u<v}\|\btheta_u(\boldsymbol{\Omega},\{\btheta_i\})-\btheta_v(\boldsymbol{\Omega},\{\btheta_i\})\|_2}_{\text{fusion (SON) term}},
\end{equation}}with optimization variables $\boldsymbol{\Omega}\in(\Delta_k)^n$ and $\{\btheta_i\}_{i=1}^k\subset\Theta_f$. The lifted node loss in~\eqref{eq:alg-objective} is \emph{linear} in $\boldsymbol{\Omega}$: each $f_v$ is evaluated at every active atom $\btheta_i$ and weighted by $\omega_{v,i}$, rather than at the per-node parameter $\btheta_v=\sum_i\omega_{v,i}\btheta_i$.

\subsection{Gradient calculus}
\label{sec:alg_grads}

We collect the gradients of $\Phi_k$ used by every step of the algorithm. Throughout, for $v\in\calV$ we let $\nabla f_v\in\partial f_v(\btheta_v)$ denote any (sub)gradient at the current per-node parameter $\btheta_v=\sum_i\omega_{v,i}\btheta_i$, and we write $\bg_{vu}\in\partial_{\btheta_v}\|\btheta_v-\btheta_u\|_2$ for the SON subgradient (with $\bg_{uv}=-\bg_{vu}$ and $\|\bg_{vu}\|_2\le 1$).

\paragraph{Gradient w.r.t.\ row $\boldsymbol{\Omega}_{v,:}$.}
Differentiating~\eqref{eq:alg-objective} entrywise and applying the chain rule $\partial\omega_{v,i}=2e_{i,v}\,\partial e_{i,v}$ to the graph term,
\begin{align}\label{eq:alg-grad-omega}
\frac{\partial \Phi_k}{\partial \omega_{v,i}}
&\;=\;
\underbrace{-\frac{(\bar\bA\,\be_i)_v}{\max(e_{i,v},\,\varepsilon)}}_{\text{graph}}
\;+\;
\underbrace{n\mu\,f_v(\bz_v;\btheta_i)}_{\text{node loss}}
\;+\;
\underbrace{\gamma\mu\Big\langle\sum_{u\neq v}\bg_{vu},\,\btheta_i\Big\rangle}_{\text{fusion}},
\end{align}
where $\varepsilon>0$ is a tiny safeguard used only when $e_{i,v}=0$. The node-loss contribution is \emph{linear} in $\omega_{v,i}$, so its row-derivative collapses to the value $f_v(\bz_v;\btheta_i)$; the fusion contribution arises through $\btheta_v=\sum_i\omega_{v,i}\btheta_i$.

\paragraph{Gradient w.r.t.\ atom parameter $\btheta_i$.}
Since $\Phi_k$ depends on $\btheta_i$ both directly through the node-loss term and indirectly through $\btheta_v(\boldsymbol{\Omega},\{\btheta_i\})$ in the fusion term,
\begin{equation}\label{eq:alg-grad-theta}
\nabla_{\btheta_i}\Phi_k
\;=\;
n\mu\sum_{v\in\calV}\omega_{v,i}\,\nabla f_v(\bz_v;\btheta_i)
\;+\;
\gamma\mu\sum_{u<v}\bg_{vu}\,\bigl(\omega_{v,i}-\omega_{u,i}\bigr).
\end{equation}

The first term is a $\boldsymbol{\Omega}$-weighted average of node-level gradients, with $\nabla f_v$ evaluated at 
% the atom 
$\btheta_i$. The second term is the fusion contribution carried through the linear dependence $\btheta_v(\boldsymbol{\Omega},\{\btheta_i\})=\sum_i\omega_{v,i}\btheta_i$.

\paragraph{Gradient w.r.t.\ per-node parameter $\btheta_v$.}
For the forward step (Sec.~\ref{sec:alg_forward}), it is convenient to also have the residual gradient at the per-node parameter
\begin{equation}\label{eq:alg-Q}
\bq_v\bigl(\boldsymbol{\Omega},\{\btheta_i\}\bigr)
\!\coloneqq\!
\nabla_{\btheta_v}\!\Bigl[n\mu\sum_{u\in\calV}f_u(\bz_u;\btheta_u)
\!+\!
\gamma\mu\sum_{u<w}\|\btheta_u-\btheta_w\|_2\Bigr]
%- old: \nabla_{\btheta_v}\!\bigl[n\mu\,\text{node}+\gamma\mu\,\text{fusion}\bigr]
\!=\!
n\mu\,\nabla f_v(\bz_v;\btheta_v)
\!+\!
\gamma\mu\!\!\sum_{u\neq v}\bg_{vu}.
\end{equation}

By the cluster-level first-order condition, $\bq_v$ vanishes over each cluster at the perfect-recovery solution, so $\|\bq_v\|_2$ is a clean ``residual signal'' identifying which node still needs an atom.

\subsection{Inner step (enhancement): alternating minimization on the active set}
\label{sec:alg_inner}

With the active set fixed, $\Phi_k$ admits the closed-form gradients above. We decrease it by block coordinate descent: every row of $\boldsymbol{\Omega}$ is updated by a Frank--Wolfe (FW) step on $\Delta_k$, and every atom $\btheta_i$ is updated by an FW step on $\Theta_f$. As in the enhancement step of CoGEnT, any method that keeps every row of $\boldsymbol{\Omega}$ in $\Delta_k$ and does not increase $\Phi_k$ can replace these FW updates.

\paragraph{Row LMO (\boldmath{$\boldsymbol{\Omega}_{v,:}$}-update).}
The linearization of $\Phi_k$  in \eqref{eq:alg-objective} decouples over rows of $\boldsymbol{\Omega}$. The FW linearization at row $v$ reads
\begin{equation}\label{eq:alg-row-lmo}
\widetilde{\boldsymbol{\Omega}}_{v,:}
\;\in\;
\arg\min_{\bomega\in\Delta_k}\;
\bigl\langle \nabla_{\boldsymbol{\Omega}_{v,:}}\Phi_k,\,\bomega\bigr\rangle
\;=\;
\arg\min_{\bomega\in\Delta_k}\;
\sum_{i=1}^k\frac{\partial\Phi_k}{\partial\omega_{v,i}}\,\omega_i,
\end{equation}
which is a linear program over the simplex; its optimum is a vertex,
\begin{equation}\label{eq:alg-row-lmo-solution}
i_v^\star\;=\;\arg\min_{i\in\{1,\ldots,k\}}\;\frac{\partial\Phi_k}{\partial\omega_{v,i}},
\qquad
\widetilde{\omega}_{v,i}\;=\;\delta_{i\,i_v^\star},
\end{equation}
with $\partial\Phi_k/\partial\omega_{v,i}$ given by~\eqref{eq:alg-grad-omega}. In words: each node $v$ is provisionally assigned to the atom that locally minimizes the sum of (i)~the graph residual $-(\bar\bA\be_i)_v/e_{i,v}$, (ii)~the lifted node loss $n\mu f_v(\btheta_i)$, and (iii)~the fusion alignment $\gamma\mu\langle\sum_{u\neq v}\bg_{vu},\btheta_i\rangle$. 

With FW step-size 
$\alpha_t\in[0,1]$, chosen for instance by a line search on $\Phi_k$, update
\begin{equation}\label{eq:alg-row-fw}
\omega_{v,i}^{(t+1)}\;=\;(1-\alpha_t)\,\omega_{v,i}^{(t)}+\alpha_t\,\widetilde\omega_{v,i},
\quad i=1,\ldots,k.
\end{equation}
Mapping back to embedding coordinates, we have
$ e_{i,v}\!=\!\sqrt{\omega_{v,i}}, \, \be_i\!=\!( e_{i,v})_{v\in\calV}.$
This preserves $e_{i,v}\ge 0$ and $\sum_i e_{i,v}^{\,2}=1$ for every $v$.

\paragraph{Atom LMO (\boldmath{$\btheta_i$}-update).}
For each active atom $i$, the FW linearization gives
\begin{equation}\label{eq:alg-theta-lmo}
\widetilde\btheta_i\;\in\;
\arg\min_{\btheta\in\Theta_f}\;
\bigl\langle\nabla_{\btheta_i}\Phi_k,\,\btheta\bigr\rangle,
\end{equation}
with $\nabla_{\btheta_i}\Phi_k$ given by~\eqref{eq:alg-grad-theta}. This is the standard support-function LMO for the convex set $\Theta_f$; it admits closed forms in all the cases of interest (Euclidean ball, $\ell_1$-ball, simplex, spectral box, Cartesian products). The FW update is
\begin{equation}\label{eq:alg-theta-fw}
\btheta_i^{(t+1)}\;=\;(1-\alpha_t)\,\btheta_i^{(t)}\;+\;\alpha_t\,\widetilde\btheta_i,\qquad i=1,\ldots,k,
\end{equation}%
which preserves $\btheta_i^{(t+1)}\in\Theta_f$ by convexity.

\subsection{Forward step: adding one atom}
\label{sec:alg_forward}
 
The inner step refines $\boldsymbol{\Omega}$ and $\{\btheta_i\}$ on the current active set but cannot introduce \emph{new} atoms. We add one new atom per outer iteration through a 
CoGEnT forward step for the reduced objective~\eqref{eq:alg-objective}: we compute the rate at which the objective changes when a candidate atom is added, select the candidate with the most negative rate by an LMO, and add it with a line-search step.
 
\paragraph{Rate of a candidate atom.}
Let $(\boldsymbol{\Omega}^{(t)},\{\btheta_i^{(t)}\})$ be the current iterate, with corresponding per-node parameters $\btheta_v^{(t)}=\sum_i\omega_{v,i}^{(t)}\btheta_i^{(t)}$, graph matrix $\bL^{(t)}=\bL(\boldsymbol{\Omega}^{(t)})$, and lifted node losses $\ell_v^{(t)}:=\sum_i\omega_{v,i}^{(t)}f_v(\bz_v;\btheta_i^{(t)})$. Consider augmenting the active set with a candidate atom $(\be,\btheta)$, with $\be\in\bbS^{n-1}$, $\be\ge\mathbf{0}$ and $\btheta\in\Theta_f$, of mass $\rho>0$, 
in which node $v$ gives the mass $\rho\,e_v^2$ to the new atom and scales its other weights by $1-\rho\,e_v^2$:
\begin{equation}\label{eq:alg-forward-path}
\boldsymbol{\Omega}(\rho)_{v,i}=\bigl(1-\rho\,e_v^{2}\bigr)\,\omega_{v,i}^{(t)}\ \ (i\le k),
\qquad
\boldsymbol{\Omega}(\rho)_{v,k+1}=\rho\,e_v^{2},
\qquad
0\le\rho\le 1/\max_{v}e_v^{2},
\end{equation}
so that every row stays in $\Delta_{k+1}$, and atom $\btheta_{k+1}=\btheta$. The first-order change in the objective is
\begin{gather}
\frac{d}{d\rho}\bigl[\Phi_{k+1}\bigl(\boldsymbol{\Omega}(\rho),\{\btheta_i^{(t)}\}\cup\{\btheta\}\bigr)\bigr]_{\rho=0^+}
\;=\;
\nonumber \\
-\,\be^\top\bar\bA\be
\;+\;\sum_{v\in\calV}e_v^{\,2}\,d_v(\btheta)+\gamma\mu\sum_{\substack{u\neq v\\\btheta_u=\btheta_v}}
|e_u^2-e_v^2|\,\|\btheta-\btheta_v\|_2.,
\label{eq:alg-fw-gain}\\
d_v(\btheta)\;:=\;n\mu\bigl[f_v(\bz_v;\btheta)-\ell_v^{(t)}\bigr]+\gamma\mu\Bigl\langle\sum_{\substack{u\neq v\\\btheta_u\neq\btheta_v}}\bg_{vu},\,\btheta-\btheta_v^{(t)}\Bigr\rangle+\bigl(\bar\bA\bL^{(t)}\bigr)_{vv}.
\label{eq:alg-dv}
\end{gather}

\paragraph{Solving the oracle.}
Let
$\mathcal P_t:=\{(u,v):u\neq v,\,
\btheta_u^{(t)}=\btheta_v^{(t)}\}$.
We approximately minimize the directional derivative \eqref{eq:alg-fw-gain} by alternating minimization over $\be$ and $\btheta$. Starting from a feasible pair, we update the embedding for fixed $\btheta^{(j)}$ as
\begin{equation}
\be^{(j+1)}\in\arg\min_{\be\ge 0,\|\be\|_2=1}
-\be^\top\bar{\bA}\be+\sum_{v\in V}e_v^2d_v(\btheta^{(j)})
+\gamma\mu\sum_{(u,v)\in\mathcal P_t}|e_u^2-e_v^2|\|\btheta^{(j)}-\btheta_v^{(t)}\|_2.
\label{eq:alg-oracle-e}
\end{equation}
For fixed $\be^{(j+1)}$, we update the atom parameter as
\begin{equation}
\btheta^{(j+1)}\in\arg\min_{\btheta\in\Theta_f}
\sum_{v\in V}(e_v^{(j+1)})^2d_v(\btheta)
+\gamma\mu\sum_{(u,v)\in\mathcal P_t}
\big|(e_u^{(j+1)})^2-(e_v^{(j+1)})^2\big|
\|\btheta-\btheta_v^{(t)}\|_2.
\label{eq:alg-oracle-theta}
\end{equation}
The embedding subproblem is generally nonconvex, while the parameter subproblem is convex. Both can be solved numerically using iterative methods. We accept only updates that do not increase the directional derivative.

\paragraph{Forward update.}
Let $(\be^{\rm new},\btheta^{\rm new})$ denote the resulting atom. If its directional derivative \eqref{eq:alg-fw-gain} is negative, we append it to the active set using
\begin{equation}
\omega_{v,i}^{(t+\frac12)}
=
\begin{cases}
(1-\rho_t(e_v^{\rm new})^2)\omega_{v,i}^{(t)},
& i\le k,\\
\rho_t(e_v^{\rm new})^2,
& i=k+1,
\end{cases}
\label{eq:alg-forward-update}
\end{equation}
where
\[
\rho_t\in
\arg\min_{0\le\rho\le 1/\max_v(e_v^{\rm new})^2}
\Phi_{k+1}\big(
\boldsymbol{\Omega}(\rho),
\{\btheta_i^{(t)}\}_{i=1}^k
\cup\{\btheta^{\rm new}\}
\big).
\]
We then set $\btheta_{k+1}=\btheta^{\rm new}$, increment $k$, and rerun the inner refinement step. If the directional derivative is nonnegative, no atom
is added.

\subsection{Backward step: atom truncation}
\label{sec:alg_backward}

After the inner refinement, several atoms may carry negligible mass $\sum_v\omega_{v,i}\!\approx\!0$ or duplicate the geometry of another atom. We prune them with the standard CoGEnT truncation rule, one atom at a time. For each currently active atom $j$, evaluate the candidate objective $\Phi_{k-1}^{(-j)}$ obtained by deleting atom $j$, re-projecting each row of $\boldsymbol{\Omega}$ onto the reduced simplex $\Delta_{k-1}$, and re-running the inner refinement on the remaining atoms. 
The atom $j$ with the smallest $\Phi_{k-1}^{(-j)}$ is pruned if
\begin{equation}\label{eq:alg-trunc-rule}
\Phi_{k-1}^{(-j)}\;-\;\Phi_k\;\le\;\eta\,\bigl(\Phi_k^{\rm prev}-\Phi_k\bigr),
\qquad \eta\in(0,1),
\end{equation}
where $\Phi_k^{\rm prev}$ is the objective at the start of the outer iteration and $\Phi_k$ the objective after the forward and inner steps. If an atom is pruned, the test is repeated on the remaining atoms, with $\Phi_k^{\rm prev}$ and $\Phi_k$ kept fixed, until it fails. The rule guarantees that truncation gives back at most a fraction $\eta$ of the gain produced by the forward  and inner steps. 

\begin{algorithm}[t]
\caption{Atomic conditional gradient solver for the unified problem~\eqref{eq:main_model}}
\label{alg:atomic-cg}
\begin{algorithmic}[1]
\setlength{\itemsep}{1.2pt}   
\Require  $\bar\bA$, node losses $\{f_v\}_{v\in\calV}$, parameters $\gamma,\mu,\eta$, max outer iterations $T$, inner cycles $T_{\rm inner}$
\State Initialize $k\!\gets\!1$, 
$\be_1\gets\mathbf{1}_n$ (i.e., $\lambda_1=n$), $\btheta_1\!\gets\!\arg\min_{\btheta\in\Theta_f}\sum_v f_v(\bz_v;\btheta)$
\State Set $\boldsymbol{\Omega}\!\gets\!\mathbf{1}_n$ (single column of all ones), $\Phi^{\rm prev}\!\gets\!$\,
$\Phi_1(\boldsymbol{\Omega},\{\btheta_1\})$
\For{$t = 1,\dots,T$}
    \Statex \hspace{1.2em}\textit{// (a) Forward step: add one atom (Sec.~\ref{sec:alg_forward})}
   \State Compute $d_v(\cdot)$ via~\eqref{eq:alg-dv} for all $v\in\calV$ and the tied pairs $\mathcal P_t$
   \State From a feasible pair $(\be^{(0)},\btheta^{(0)})$, alternate~\eqref{eq:alg-oracle-e} and~\eqref{eq:alg-oracle-theta}, keeping only updates that decrease~\eqref{eq:alg-fw-gain}; denote the result $(\be^{\rm new},\btheta^{\rm new})$
   \If{the rate~\eqref{eq:alg-fw-gain} at $(\be^{\rm new},\btheta^{\rm new})$ is negative}
      \State Choose $\rho_t$ by line search on $\Phi_{k+1}$ along the path~\eqref{eq:alg-forward-path}
      \State Update $\boldsymbol{\Omega}$ via~\eqref{eq:alg-forward-update}; set $\btheta_{k+1}\gets\btheta^{\rm new}$, $k\gets k+1$
   \EndIf
   \Statex \hspace{1.2em}\textit{// (b) Inner step (enhancement)}: alternating minimization on the active set
   \For{$T_{\rm inner}$ cycles}
      \For{each $v\in\calV$}
         \State Compute $\partial\Phi_k/\partial\omega_{v,i}$ via~\eqref{eq:alg-grad-omega}
         \State Solve row LMO~\eqref{eq:alg-row-lmo} and update row $\boldsymbol{\Omega}_{v,:}$ via~\eqref{eq:alg-row-fw}
      \EndFor
      \For{each $i=1,\ldots,k$}
         \State Compute $\nabla_{\btheta_i}\Phi_k$ via~\eqref{eq:alg-grad-theta}
         \State Solve atom LMO~\eqref{eq:alg-theta-lmo} and update $\btheta_i$ via~\eqref{eq:alg-theta-fw}
      \EndFor
   \EndFor
   \Statex \hspace{1.2em}\textit{// (c) Backward step: truncate redundant atoms (Sec.~\ref{sec:alg_backward})}
   \State Prune atoms by rule~\eqref{eq:alg-trunc-rule}; let $k$ be the new active size
   \State Set $\Phi^{\rm prev}\!\gets\!\Phi_k(\boldsymbol{\Omega},\{\btheta_i\})$
\EndFor
\State \Return $(\boldsymbol{\Omega},\{\btheta_i\}_{i=1}^k)$ and the partition $\widehat{\calC}_v=\arg\max_i\omega_{v,i}$
\end{algorithmic}
\end{algorithm}

\vfill

% \newpage
% \input{checklist.tex}

\end{document}